\documentclass[11pt]{article}

\usepackage[final]{acl}

 \usepackage{microtype}

\usepackage[english,bidi=default]{babel} 
\babelfont{rm}{TeXGyreTermesX} 
\babelprovide[import]{hindi}
\babelfont[*devanagari]{rm}{Lohit Devanagari}
\babelprovide[import]{arabic}
\babelfont[*arabic]{rm}{Noto Sans Arabic}

\usepackage{lipsum}
\usepackage{booktabs}
\usepackage{multirow}
\usepackage{caption}
\usepackage{amssymb}
\usepackage{subcaption}
\usepackage{graphicx}
\usepackage{makecell}
\usepackage{kotex} 
\usepackage{tcolorbox} 
\tcbuselibrary{skins, breakable} 
\usepackage{geometry} 
\usepackage{listings}
\usepackage{tabularx}
\usepackage{xcolor}
\usepackage{float}
\usepackage{stfloats}
\usepackage{placeins}
\usepackage{afterpage}
\usepackage{relsize}
\usepackage{adjustbox}
\usepackage{amsmath}
\usepackage{enumitem}
\usepackage{minitoc}
\usepackage{booktabs,multirow,longtable}
\setlist[itemize]{leftmargin=10pt, labelindent=0pt}
\usepackage[table]{xcolor}
\definecolor{lightred}{rgb}{1.0, 0.8, 0.8}
\definecolor{palegray}{gray}{0.98}
\definecolor{lightgray}{gray}{0.90}
\definecolor{lightgraytext}{gray}{0.70}
\definecolor{googleblue}{HTML}{4285F4}
\definecolor{googlered}{HTML}{EA4335}
\definecolor{googleyellow}{HTML}{FBBC05}
\definecolor{googlegreen}{HTML}{34A853}
\newcommand{\googb}[1]{\textcolor{googleblue}{#1}}
\newcommand{\googr}[1]{\textcolor{googlered}{#1}}
\newcommand{\googy}[1]{\textcolor{googleyellow}{#1}}
\newcommand{\googg}[1]{\textcolor{googlegreen}{#1}}
\newcommand{\gtrans}[1]{\textcolor{lightgraytext}{\textit{#1}}}
\usepackage{tikz}
\newcommand{\yes}{%
    \tikz[baseline=-0.5ex]{
        \node[fill=green!60!black, rounded corners=2pt, inner sep=1.5pt, anchor=base] 
        {\color{white}\tiny\bfseries\checkmark};
    }%
}
\newcommand{\noBox}{%
    \tikz[baseline=-0.5ex]{
        \node[fill=red, rounded corners=2pt, inner sep=2pt, anchor=base] 
        {\color{white}\tiny\bfseries\sffamily X};
    }%
}
\newcommand{\ko}[1]{\scalebox{0.9}[0.9]{#1}}
\newcolumntype{L}{>{\raggedright\arraybackslash}X}
\newcommand{\narrow}[1]{\scalebox{1.0}[1]{#1}}
\newtcolorbox{promptbox}[1][]{
  colback=white, 
  colframe=black, 
  coltitle=white,          
  title={#1},              
  fonttitle=\relscale{0.98}\bfseries,
  boxrule=0.8pt,           
  fontupper=\relscale{0.95},
}
\newtcolorbox{promptbox_kor}[1][]{
  colback=white, 
  colframe=black,
  coltitle=white,          
  title={#1},              
  fonttitle=\relscale{0.98}\bfseries,
  boxrule=0.8pt,           
  fontupper=\relscale{0.95},
}

\title{K-MetBench: A Multi-Dimensional Benchmark for Fine-Grained Evaluation of Expert Reasoning, Locality, and Multimodality in Meteorology}

\author{
Soyeon Kim \\
KAIST, 
INEEJI \\
Seongnam, Korea\\
\texttt{soyeon.k@kaist.ac.kr}
\And
Cheongwoong Kang \and Myeongjin Lee \\
KAIST \\
Seongnam, Korea\\
\texttt{cw.kang; lmjk311@kaist.ac.kr}
\AND
Eun-Chul Chang \and Jaedeok Lee \\
Kongju National University \\
Gongju, Korea\\
\texttt{echang@kongju.ac.kr, ruio1084@gmail.com}
\And
Jaesik Choi\thanks{Corresponding author.} \\
KAIST, INEEJI \\
Seongnam, Korea\\
\texttt{jaesik.choi@kaist.ac.kr}
}

\begin{document}
\maketitle
\begin{abstract}
The development of practical (multimodal) large language model assistants for Korean weather forecasters is hindered by the absence of a multidimensional, expert-level evaluation framework grounded in authoritative sources. To address this, we introduce K-MetBench, a diagnostic benchmark grounded in national qualification exams. It exposes critical gaps across four dimensions: expert visual reasoning of charts, logical validity via expert-verified rationales, Korean-specific geo-cultural comprehension, and fine-grained domain analysis. Our evaluation of 55 models reveals a profound \emph{modality gap} in interpreting specialized diagrams and a \emph{reasoning gap} where models hallucinate logic despite correct predictions. Crucially, Korean models outperform significantly larger global models in local contexts, demonstrating that parameter scaling alone cannot resolve cultural dependencies. K-MetBench serves as a roadmap for developing reliable, culturally aware expert AI agents. 
The dataset is available at \url{https://huggingface.co/datasets/soyeonbot/K-MetBench}.
\end{abstract}


\section{Introduction}

Large language models (LLMs) and multimodal large language models (MLLMs) have shown growing promise in scientific domains~\cite{taylor2022galactica,team2023gemini,openai2025gpt52}, achieving performance matching passing thresholds on professional certification exams~\cite{singhal2023large,katz2024gpt}. As these models are increasingly positioned as assistants for domain-specific tasks, there is a growing need for evaluation frameworks that go beyond surface-level correctness and more precisely characterize domain-relevant competencies~\cite{liang2022holistic}. However, existing benchmarks for vertical domains often summarize performance using a single aggregate score, making it difficult to understand why a model succeeds or fails in practice. In complex applied fields such as meteorology, this coarse evaluation obscures \textbf{critical limitations} for real-world deployment. We identify four recurring limitations in current evaluations of meteorological reasoning.

\begin{figure}[t!]
    \centering
    \includegraphics[width=\linewidth]{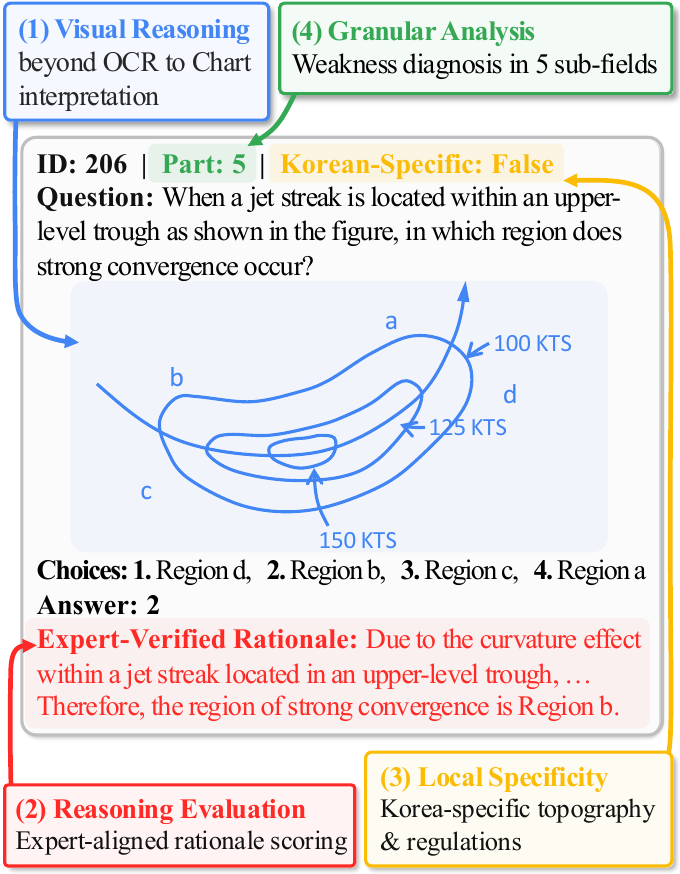}
    \caption{\textbf{An example of the K-MetBench dataset (translated into English).} K-MetBench provides evaluation across four critical dimensions: (1) multimodal understanding, (2) expert-level reasoning, (3) geo-cultural context sensitivity, and (4) fine-grained domain knowledge across five meteorological sub-fields.}
    \label{fig:main}
\end{figure}

First, \textbf{the modality gap}. Meteorological analysis, inherently multimodal, requires the synthesis of numerical data, textual descriptions, and specialized visual charts (e.g., weather maps, skew-T log-P diagrams). However, most scientific benchmarks remain predominantly text-based and provide limited assessment of a model’s ability to interpret domain-specific charts and spatial patterns. As a result, visual understanding capabilities central to operational forecasting remain under-evaluated.

Second, \textbf{the reasoning gap}.
Conventional benchmarks primarily rely on answer accuracy, without explicitly evaluating the validity or structure of the underlying reasoning. In high-stakes domains like weather forecasting, a correct prediction reached through shallow heuristics or incomplete logic may still lead to brittle or unreliable behavior~\cite{turpin2023language}. Without access to expert-aligned rationales, it is difficult to distinguish genuine understanding from shortcut learning.

Third, \textbf{the geo-cultural gap}.
Many existing datasets emphasize global or universal physical principles while abstracting away local geographic and institutional context. In meteorology, however, local topography, climatological conventions, and region-specific regulations play a substantial role in interpretation and decision-making. Models trained and evaluated solely on decontextualized data may therefore fail to generalize reliably to region-specific applications.

Fourth, \textbf{the granularity gap}.
Aggregate performance scores often mask uneven competence across sub-domains. A model may perform well on factual recall or chart interpretation while struggling with quantitative reasoning or applied dynamics. Without fine-grained analysis, such disparities remain difficult to diagnose.

To address these limitations, we introduce \textbf{K-MetBench}, a Korean meteorological benchmark designed for multi-dimensional evaluation of LLMs and MLLMs. Rather than treating meteorological expertise as a monolithic capability, K-MetBench decomposes evaluation along four complementary axes: \textbf{(1) multimodal understanding} of meteorological charts and symbols, \textbf{(2) reasoning quality} assessed using expert-verified rationales, \textbf{(3) sensitivity to geo-cultural and regional context}, and \textbf{(4) fine-grained coverage} across five officially defined meteorological sub-domains. Through this structured design, K-MetBench is intended as a diagnostic tool that helps reveal which aspects of meteorological reasoning remain challenging for current models, and why.

\begin{table*}[ht]
\centering
\setlength{\tabcolsep}{1pt} 
\caption{\textbf{Comparison with existing benchmarks.} K-MetBench distinguishes itself by covering four key axes: \googb{visual understanding}, \googr{rationale reliability}, \googy{geo-cultural alignment}, and \googg{fine-grained diagnosis in sub-domains}.}
\resizebox{\textwidth}{!}{%
\begin{tabular}{lccccccc}
\toprule
\textbf{Dataset} & \textbf{Lang.} & \textbf{Domain} & \textbf{Test Size (Source)} & \textbf{\googb{Modality}} & \textbf{\googr{Reasoning}} & \textbf{\googy{Geo-Cultural}} & \textbf{\googg{Granularity}} \\ 
\midrule
KMMLU~\cite{son2025kmmlu} & Kor & General & 35k (License Exam) & Text & $\times$ & Korea & $\times$ (45 Subjects) \\ 
KMMLU-Redux~\cite{hong2025kmmlu} & Kor & General & 2.6k (License Exam) & Text & $\times$ & Korea & $\times$ (14 Subjects) \\
ClimaQA~\cite{manivannan2024climaqa} & Eng & Climate & 566 (Autogenerate) & Text & $\times$ & Global & $\times$ (3 Tasks) \\
ClimateIQA~\cite{chen2025climateiqa} & Eng & Climate & 152k (Template) & Image+Text & $\times$ & Global & $\times$ (4 Tasks) \\
WeatherQA~\cite{ma2024weatherqa} & Eng & Weather Forecast & 600 (Template) & Image+Text & $\times$ & United States & $\times$ (2 Tasks) \\ 
\midrule
\rowcolor{lightgray}
\textbf{K-MetBench (Ours)} & \textbf{Kor} & \textbf{Meteorology} & \textbf{1.7k (License Exam)} & \textbf{Image+Text} & \textbf{Expert-Verified} & \textbf{Korea} & \textbf{5 Sub-domains} \\ 
\bottomrule
\end{tabular}%
}
\vspace{1ex}
\begin{minipage}{\linewidth}
\footnotesize
\textbf{Note for K-MetBench.} 
\textit{\googb{Modality}:} Includes multimodal questions evaluating interpretation of professional weather charts. 
\textit{\googr{Reasoning}:} Provides rationale verified by domain experts. 
\textit{\googy{Geo-Cultural}:} Includes questions requiring knowledge of local geography and regulations that are specific to Korea (e.g., the Korea Meteorological Administration (KMA) protocols).
\textit{\googg{Granularity}:} Supports fine-grained diagnosis across the five sub-domains officially defined in Korea Engineer Meteorology certification exam.
\end{minipage}
\label{tab:benchmarks}
\end{table*}

\begin{table}[ht]
\centering
\caption{\textbf{Detailed statistics of K-MetBench.} The dataset is structured into four key dimensions to enable structured evaluation: Modality, Reasoning, Geo-cultural, and Granularity.}
\resizebox{\linewidth}{!}{
\setlength{\tabcolsep}{2pt}
\renewcommand{\arraystretch}{1.2} 
\begin{tabular}{llr}
\toprule
\textbf{Diagnostic Axis} & \textbf{Statistic} & \textbf{Value} \\
\midrule
\textbf{1. Overview} & Total Questions & \textbf{1,774} \\
\hline
\textbf{\googb{2. Modality}} & Image+Text Questions & \textbf{82 (4.62\%)} \\
\textit{(Visual Understanding)} & \textit{(Charts, Diagrams)} & \\
\hline
\textbf{\googr{3. Reasoning}} & Avg. Rationale Length & \textbf{93.72 tokens} \\
\textit{(Expert Rationales)} & Text-Only Reasoning & \textbf{121 (6.82\%)} \\
 & Multimodal Reasoning & \textbf{20 (1.13\%)} \\
\hline
\textbf{\googy{4. Geo-Cultural}} & Korean-Specific & \textbf{73 (4.11\%)} \\
\textit{(Local Knowledge)} & Questions 
\\
\hline
\textbf{\googg{5. Granularity}} & Part 1: Forecast Theory & 373 (21.03\%) \\
\textit{(5 Subject Areas)} & Part 2: Observation & 332 (18.71\%) \\
 & Part 3: Atmos. Dynamics & 359 (20.24\%) \\
 & Part 4: Climatology & 376 (21.20\%) \\
 & Part 5: Atmos. Physics & 334 (18.83\%) \\
\bottomrule
\end{tabular}
}
\vspace{1ex}
\begin{minipage}{\linewidth}
\footnotesize
\textit{Note:} The number of tokens is calculated using the \texttt{gemini-2.5-flash} tokenizer.
(Atmos.: Atmospheric)
\end{minipage}
\label{tab:statistics}
\end{table}

\begin{table*}[ht]
\centering
\caption{\textbf{Distribution of K-MetBench across five sub-domains.} The number of questions for each subject area is reported, with the number of reasoning questions featuring expert-verified rationales in parentheses (Reas. stands for Reasoning).}
\label{tab:orthogonal_dist}
\resizebox{\textwidth}{!}{
\setlength{\tabcolsep}{4pt}
\begin{tabular}{cl c cc c}
\toprule
\multirow{2}{*}{\textbf{Part}} & \multirow{2}{*}{\textbf{\googg{Subject Area}}} & \textbf{Overall Volume} & \multicolumn{2}{c}{\textbf{\googb{Modality}}} & \textbf{\googy{Geo-Cultural}} \\
\cmidrule(lr){3-3}\cmidrule(lr){4-5}\cmidrule(lr){6-6}
 & & \textbf{Total (\googr{Reas.})} & \textbf{Text (\googr{Reas.})} & \textbf{Image + Text (\googr{Reas.})} & \textbf{Korean (\googr{Reas.})} \\
\midrule
1 & Weather Analysis \& Forecast Theory
    & 373 (28) & 364 (24) & 9 (4) & 6 (0) \\
2 & Meteorological Observation Methods       
    & 332 (28) & 318 (24) & 14 (4) & 0 (0) \\
3 & Atmospheric Dynamics        
    & 359 (29) & 340 (25) & 19 (4) & 0 (0) \\
4 & Climatology           
    & 376 (28) & 363 (24) & 13 (4) & 50 (7) \\
5 & Atmospheric Physics        
    & 334 (28) & 307 (24) & 27 (4) & 17 (0) \\
\midrule
\textbf{Sum} & \textbf{Total Coverage} & \textbf{1,774 (141)} & \textbf{1,692 (121)} & \textbf{82 (20)} & \textbf{73 (7)} \\
\bottomrule
\end{tabular}%
}
\label{tab:coverage}
\end{table*}

\section{Related Work}

Existing benchmarks for meteorological and climate reasoning reflect diverse assumptions about knowledge sources, modalities, and evaluation objectives. Rather than treating them as competitors, we situate them along complementary axes that highlight different aspects of domain expertise.

ClimaQA~\cite{manivannan2024climaqa} evaluates climate question answering using textbooks as the knowledge source. By grounding questions in established instructional materials, it emphasizes conceptual understanding and theoretical reasoning characteristic of graduate-level climate science. While this approach provides scientific rigor, it remains purely text-based and does not assess visual interpretation or operational reasoning grounded in real-world artifacts.
ClimateIQA~\cite{chen2025climateiqa} constructs instruction-style QA data from numerical weather prediction (NWP) heatmaps and associated geospatial metadata. This enables evaluation of visual pattern recognition and structured data interpretation.
WeatherQA~\cite{ma2024weatherqa} further targets operational forecasting scenarios by combining multiple meteorological images with expert-written mesoscale discussions. These datasets advance multimodal evaluation, but their emphasis remains on task-level performance rather than fine-grained diagnosis across sub-fields or distinct reasoning failures.

In the Korean-language evaluation landscape, KMMLU~\cite{son2025kmmlu} is derived from official national examinations, measuring expert-level linguistic competence across a wide range of professions. Since it is based on official Korean exams, KMMLU captures linguistic and cultural aspects of the Korean language. KMMLU-Redux~\cite{hong2025kmmlu} is a reconstructed version of KMMLU that removes erroneous, ambiguous, or contaminated items to improve reliability. While these benchmarks offer high reliability and clear passing criteria, they are primarily text-based and treat meteorological knowledge as a small subset within a broader evaluation suite, limiting their ability to analyze domain-specific competencies in depth.


\section{K-MetBench Construction}

K-MetBench is designed to complement the existing benchmarks by explicitly separating and jointly examining four dimensions that are often conflated in prior benchmarks. Rather than introducing new task formats, K-MetBench focuses on providing diagnostic visibility into how and where current models succeed or fail when approaching expert-level meteorological reasoning.

\subsection{Data Collection and Processing}
K-MetBench is constructed from raw data drawn from the National Meteorological Engineer certification examinations, covering 25 exam sessions between March 16, 2003 and March 5, 2022. The initial pool comprised 2,500 multiple-choice questions. Because these examinations are generated from a shared question bank, substantial overlap exists across years. To construct a balanced and non-redundant benchmark, we applied a multi-stage filtering and augmentation pipeline.

For deduplication~\citep{lee2021deduplicating}, we first applied \texttt{difflib.SequenceMatcher} with a similarity threshold of 0.6, removing exact duplicates as well as items with trivially permuted answer options. Importantly, questions with inverted logic (e.g., `highest' vs. `lowest', `saturated' vs. `unsaturated') were manually reviewed and retained, as they probe distinct reasoning behaviors despite surface similarity. This process yielded a refined set of 1,774 questions.

To reduce memorization and contamination effects, we applied two transformations. First, we randomized answer option orders for all questions. Second, we paraphrased question stems using \texttt{Gemini-2.5-Pro}, with strict constraints to preserve technical terminology and domain-specific meaning. The system prompt used for paraphrasing is provided in Appendix~\ref{appx:prompt_paraphrasing}. To maintain quality, a human researcher reviewed and refined 14.88\% (264/1,774) of the paraphrased items.

For multimodal questions, both text and visual elements were extracted from the original examination PDFs. Three researchers reviewed and cross-checked all extracted images to correct parsing artifacts such as missing axis labels, distorted symbols, or incomplete annotations. As a design choice to separate perceptual challenges from reasoning difficulty, mathematical formulas embedded as images were transcribed into LaTeX code to prevent OCR bottlenecks, while meteorological charts and diagrams were preserved in their original format.

\subsection{Subset 1: Multimodal Diagnosis}
The multimodal subset of K-MetBench consists of 82 questions (4.62\% of the dataset) that require interpretation of meteorological visuals. Unlike general-purpose multimodal benchmarks that focus on object recognition or scene description, this subset targets domain-specific charts and symbolic representations.
The included materials span surface weather maps, upper-level charts (e.g., 200 and 500 hPa), and thermodynamic diagrams such as Skew-T Log-P plots and emagrams derived from radiosonde measurements. Solving these questions requires extracting structured information, including pressure gradients, wind vectors, and thermodynamic indices—from dense visual fields that cannot be resolved through OCR alone.
Consequently, this subset assesses the ability of MLLMs to integrate textual meteorological knowledge with the interpretation of domain-specific visual cues. Representative examples are provided in Appendix Table~\ref{tab:dataset_example}.

\subsection{Subset 2: Reasoning-Aware Evaluation}
To evaluate reasoning quality beyond final answer correctness, K-MetBench includes a reasoning-aware subset consisting of 141 questions paired with expert-verified rationales. These rationales serve as reference explanations for assessing the validity, coherence, and depth of model-generated reasoning.
Rationale construction followed a two-stage process. First, \texttt{GPT-5} was used to generate initial reasoning drafts, guided by prompts that emphasized logical flow, factual consistency, clarity, and completeness. Second, two meteorology professors reviewed these drafts, correcting factual errors, refining physical explanations, and resolving ambiguities.
We employ an LLM-as-a-Judge framework \cite{zheng2023judging} to score model-generated rationales against the expert-verified rationales as reference standard. The system prompts used for reasoning generation and evaluation are detailed in Appendix~\ref{appx:prompt_rationale_generation} and~\ref{appx:questionnaire_rationale_verification}.
To validate the reliability of this framework in a specialized domain, we conduct a meta-evaluation \cite{li2024llms} comparing LLM judgments with human expert scores. The experimental and survey protocols are provided in Appendix~\ref{appx:exp_setup_meta_eval} and~\ref{appx:survey_expert_scoring}.

\subsection{Subset 3: Geo-Cultural Sensitivity}
Meteorological reasoning is strongly influenced by local geography, climate patterns, and institutional conventions. To capture this dependency, we annotate a \textit{Korean-Specific} subset comprising 73 questions that involve implicit, speaker-centric, or high-context expressions specific to the Korean Peninsula.
Candidate items were identified using prompt-enhanced LLMs (\texttt{GPT-4.1} and \texttt{Gemini-2.5-Pro}) designed to detect references to localized phenomena, such as regional topography (e.g., the Yeongdong region) or regulations issued by the Korea Meteorological Administration (KMA). These candidates were subsequently reviewed and validated by two researchers to ensure relevance and correctness.
Rather than testing translation ability, this subset probes whether models can appropriately ground meteorological knowledge in region-specific context. As such, it provides a controlled setting for analyzing geo-cultural alignment in domain-specific reasoning.

\subsection{Subset 4: Domain Specificity}
To enable fine-grained analysis of meteorological expertise, K-MetBench is organized into five official subject areas defined in the Korean Meteorological Engineer certification exam. These include: Part~1 (Weather Analysis and Forecast Theory), Part~2 (Meteorological Observation Methods), Part~3 (Atmospheric Dynamics), Part~4 (Climatology), and Part~5 (Atmospheric Physics).
Each subject area targets a distinct aspect of professional competence, ranging from chart interpretation and numerical weather prediction principles to instrumentation, large-scale atmospheric motion, climate systems, and thermodynamic calculations. This structure allows model performance to be examined at a level of granularity that is not visible from aggregate scores alone.
By aligning evaluation with established subject boundaries, this design facilitates diagnosis of domain-specific strengths and weaknesses, for example, distinguishing models that perform well on descriptive climatology but struggle with quantitative dynamics or thermodynamics.

\section{Experiments}

\subsection{Experimental Setup}

\paragraph{Evaluated Models.} 
To ensure a comprehensive benchmark, we evaluated a diverse array of models categorized by scale, training language, and modality support. The selection includes proprietary state-of-the-art models renowned for superior reasoning capabilities, such as \texttt{GPT-5.2} (evaluated with and without reasoning modules enabled)~\cite{openai2025gpt52} and \texttt{Gemini-3-Pro-Preview}~\cite{team2023gemini}. We also incorporated open-source models ranging from 0.6B to 235B parameters, exemplified by \texttt{InternVL3.5}~\cite{wang2025internvl3} and \texttt{Qwen3-VL}~\cite{yang2025qwen3}, alongside large-scale foundation models such as \texttt{gpt-oss-120b}~\cite{agarwal2025gpt}, \texttt{command-a-reasoning-08-2025}~\cite{cohere2025command}, and \texttt{Llama-3.2-90B-Vision-Instruct}~\cite{meta2024llama32}. To investigate the impact of geo-cultural knowledge, we specifically included Korean-centric models, including \texttt{EXAONE-4.0}~\cite{research2025exaone}, \texttt{A.X-4.0}~\cite{skt2025adotx4}, \texttt{VARCO-Vision-2.0}~\cite{cha2025varco}, and \texttt{HyperCLOVA X}~\cite{yoo2024hyperclova}. Finally, strictly text-based baselines were established by evaluating non-multimodal models solely on the textual components of questions to quantify text dependency.


\paragraph{Geo-Cultural Disambiguation Protocol.}
To establish a fair evaluation protocol for global models, we designed four experimental configurations that cross-reference question formulation with prompting conditions. This setup ensures that models are assessed on their meteorological competence rather than their ability to decode localized linguistic ambiguities. For question formulation, we compared an \textit{Implicit} condition, using original speaker-centric terms like `Our country,' against an \textit{Explicit} condition, which replaces these with proper nouns (e.g., `South Korea') to isolate and evaluate pure domain knowledge. Regarding prompting conditions, beyond a \textit{Standard} prompt that injects an expert persona, we introduced an \textit{Advanced} prompt providing explicit disambiguation (e.g., `` `Our country' refers to South Korea"). This advanced protocol serves as a specialized support layer, mitigating performance degradation caused by implicit geo-cultural references and enabling global models to compete on an equal footing.

\paragraph{Comparison with Existing Benchmarks.}
We evaluated models using the official test sets of all datasets, employing the Chain-of-Thought (CoT)~\cite{wei2022chain} protocol for WeatherQA. Task orthogonality was analyzed using Kendall's Tau-b rank correlation coefficient. To align the distance-based Haversine metric of ClimaIQA (where lower is better) with standard accuracy metrics, we inverted the sign of ClimaIQA scores prior to calculating correlations.

\paragraph{Meta-Evaluation Setup: Validating LLM-as-a-Judge.}
Given the specialized nature of meteorology, validating the reliability of commercial LLMs as judges is crucial. We conducted a meta-evaluation comparing human expert judgments with LLM judgments. We selected ten representative questions varying in difficulty and type, and collected reasoning outputs from ten open-source LLMs. Two human experts provided gold standard scores, while \texttt{Gemini-2.5-Pro} served as the AI evaluator. Both parties utilized identical expert-verified references and a scoring rubric across four axes: Factuality, Logicality, Depth, and Clarity. We calculated Kendall's Tau-b ($\tau_b$) correlation between human and AI scores, confirming the alignment of the automated judge ($\tau_b > 0.8$). We also computed Krippendorff's $\alpha$ (interval) and Intraclass Correlation Coefficient (ICC, 2-way mixed, absolute) to assess inter-rater reliability, which indicated acceptable agreement ($\alpha > 0.7$). To investigate whether incorporating human expert rationales improves the alignment between the LLM evaluator and human judgment, we compared the correlations of their scores under conditions with and without rationale availability. 

\paragraph{Implementation Details.}
To ensure a fair comparison, we utilized \textit{Standard} prompts across all models. We applied a zero-shot setting to all text, multimodal, and reasoning questions to evaluate intrinsic capabilities. We computed accuracy by extracting final answers via regular expressions. To rigorously assess instruction-following capabilities, we counted any output that violated the required format as a failure case. We employed the vLLM library~\cite{kwon2023efficient} with its default configurations, except for \texttt{A.X-4.0-VL-Light} and \texttt{Llama-3.2-90B-Vision-Instruct}, which were run using Hugging Face Transformers. The random seed was fixed at 42, and sampling temperatures were set to 0.1 by default, while a temperature of 1.0 was employed for reasoning models. 
All prompts and questions were provided in the original Korean to strictly evaluate localized comprehension without translation artifacts.

\section{Results}
\label{sec:result}
Beyond simple leaderboards, we dissect the performance of models across four dimensions to reveal their true capabilities and limitations.

\begin{table*}[ht]
  \centering
  \caption{\textbf{K-MetBench performance scores across diverse models.} Models are sorted by accuracy. Accuracy score ranges from 0 to 100, while the reasoning score (\textbf{\googr{Reas.}}) ranges from 4 to 20. The highest scores in each column are shown in \textbf{bold} for proprietary and open-source models, respectively. (Acc.: Accuracy, \textsf{K}: Korean model, \textsf{V}: Vision language model, \textsf{R}: Reasoning model.)}
  \label{tab:kmetbench_scores}
  \tiny
  \setlength{\aboverulesep}{1pt}
  \setlength{\belowrulesep}{1pt}
  \setlength{\tabcolsep}{2pt}
  \renewcommand{\arraystretch}{1.05}
  \begin{adjustbox}{width=\textwidth,totalheight=\textheight,keepaspectratio}
  \begin{tabular}{@{}l p{0.22\textwidth} c c c rrrrrrrrrr@{}}
  \toprule
  \multirow{2}{*}{\textbf{Type}} & \multirow{2}{*}{\textbf{Model}} & \multicolumn{3}{c}{\textbf{Type}} & \multicolumn{1}{c}{\textbf{Acc.}} & \multicolumn{1}{c}{\textbf{\googr{Reas.}}} & \multicolumn{1}{c}{\textbf{\googy{Geo-Cult.}}} & \multicolumn{2}{c}{\textbf{\googb{Modality}}} & \multicolumn{5}{c}{\textbf{\googg{Granularity (P1--P5)}}} \\
  \cmidrule(lr){3-5}\cmidrule(lr){6-6}\cmidrule(lr){7-7}\cmidrule(lr){8-8}\cmidrule(lr){9-10}\cmidrule(lr){11-15}
   &  & \textsf{K} & \textsf{V} & \textsf{R} &  &  & Korean & Text & Multi & P1 & P2 & P3 & P4 & P5 \\
   
  \midrule
  & \includegraphics[height=0.8em]{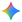} \texttt{gemini-3-pro-preview} (Thinking) &   &  \textsf{V} & \textsf{R} & \textbf{93.7} & \textbf{18.01} & \textbf{90.4} & \textbf{94.6} & \textbf{75.6} & \textbf{92.5} & \textbf{97.9} & \textbf{94.2} & \textbf{92.8} & \textbf{91.6} \\
  \rowcolor{palegray}
  \cellcolor{white} & \includegraphics[height=0.8em]{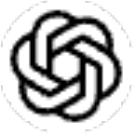} \texttt{gpt-5.2} (Thinking) &   &  \textsf{V} & \textsf{R} & 87.8 & 17.33 & 80.8 & 90.6 & 29.3 & 86.3 & 93.4 & 88.0 & 86.2 & 85.3 \\
  \multirow{-3}{*}{\textbf{Proprietary}}
  & \includegraphics[height=0.8em]{icon/openai.pdf} \texttt{gpt-5.2} &  & \textsf{V} &  & 77.6 & 17.39 & 75.3 & 79.0 & 50.0 & 77.2 & 81.3 & 71.9 & 81.4 & 76.3 \\
  \midrule
    & \multicolumn{14}{c}{\textbf{Multilingual Models}}\\
  \cmidrule(lr){2-15}
   & \includegraphics[height=0.5em]{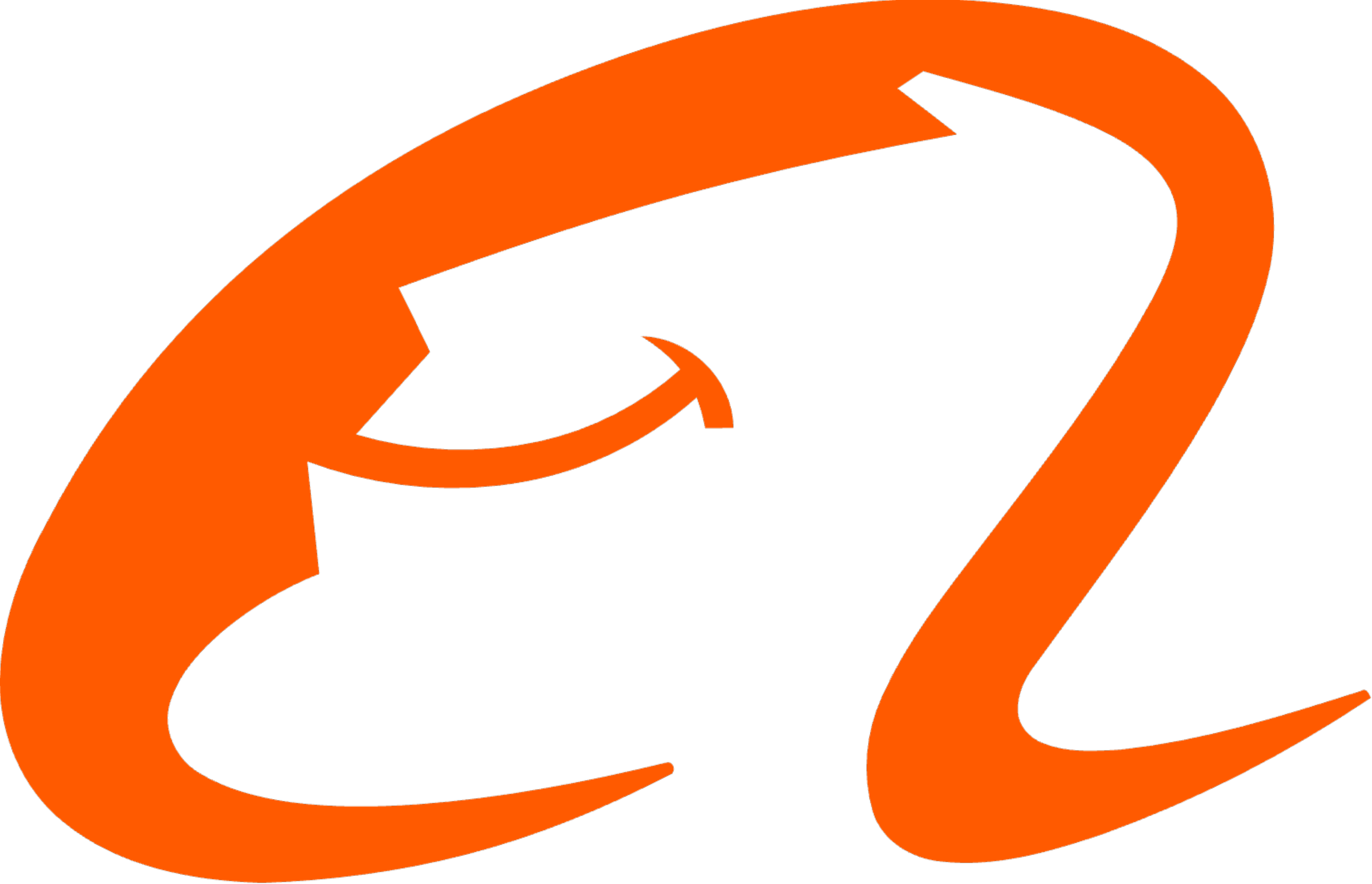} \texttt{Qwen3-VL-235B-A22B-Thinking} &   &  \textsf{V} & \textsf{R} & \textbf{84.4} & \textbf{17.22} & 72.6 & \textbf{86.2} & 48.8 & \textbf{81.5} & \textbf{88.6} & \textbf{87.2} & \textbf{83.2} & \textbf{82.0} \\
   \rowcolor{palegray}
   \cellcolor{white} & \includegraphics[height=0.5em]{icon/qwen.pdf} \texttt{Qwen3-VL-32B-Thinking} &   &  \textsf{V} & \textsf{R} & 78.6 & 16.19 & 60.3 & 79.9 & \textbf{51.2} & 74.3 & 85.2 & 78.8 & 78.7 & 76.3 \\
   & \includegraphics[height=0.8em]{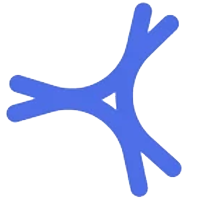} \texttt{command-a-reasoning-08-2025} &   &  & \textsf{R} & 77.8 & 14.12 & 74.6 & 77.8 & - & 73.4 & 85.2 & 73.8 & 78.8 & 78.5 \\
   \rowcolor{palegray}
   \cellcolor{white} & \includegraphics[height=0.8em]{icon/openai.pdf} \texttt{gpt-oss-120b} &   &  & \textsf{R} & 77.3 & 16.12 & 62.0 & 77.3 & - & 72.5 & 85.8 & 76.5 & 77.4 & 74.9 \\
   & \includegraphics[height=0.5em]{icon/qwen.pdf} \texttt{Qwen3-30B-A3B-Thinking-2507} &   &  & \textsf{R} & 76.7 & 15.76 & 67.6 & 76.7 & - & 75.5 & 82.1 & 75.6 & 74.9 & 75.9 \\
   \rowcolor{palegray}
   \cellcolor{white} & \includegraphics[height=0.8em]{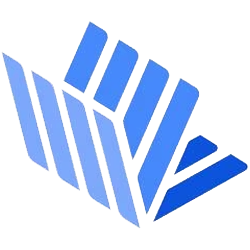} \texttt{InternVL3.5-38B-Instruct} &   &  \textsf{V} & & 57.3 & 11.38 & 47.9 & 58.1 & 40.2 & 56.0 & 64.8 & 48.7 & 61.4 & 55.7 \\
   & \includegraphics[height=0.8em]{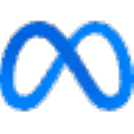} \texttt{Llama-3.2-90B-Vision-Instruct} &  & \textsf{V} &  & 56.9 & 9.72 & 52.1 & 58.2 & 30.5 & 57.1 & 59.3 & 52.4 & 62.2 & 53.3 \\
   \rowcolor{palegray}
   \cellcolor{white} & \includegraphics[height=0.8em]{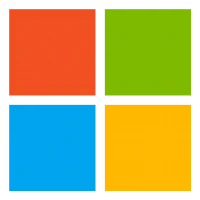} \texttt{Phi-4} &  &  &  & 51.5 & 11.75 & 40.8 & 51.5 & - & 52.5 & 53.8 & 50.0 & 55.1 & 45.3 \\
   
  \cmidrule(lr){2-15}
    & \multicolumn{14}{c}{\textbf{Korean Models}}\\
  \cmidrule(lr){2-15}
   & \includegraphics[height=0.8em]{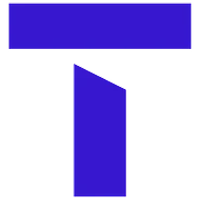} \texttt{A.X-4.0} & \textsf{K} &  &  & 76.1 & 15.46 & \textbf{78.9} & 76.1 & - & 76.6 & 77.7 & 68.2 & 81.3 & 76.5 \\
   \rowcolor{palegray}
   \cellcolor{white} & \includegraphics[height=0.8em]{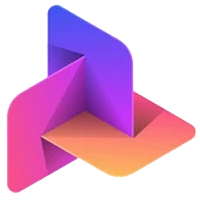} \texttt{EXAONE-4.0-32B} & \textsf{K} &  & \textsf{R} & 59.9 & 13.57 & 59.2 & 59.9 & - & 58.2 & 64.8 & 52.4 & 63.1 & 61.2 \\
   & \includegraphics[height=0.8em]{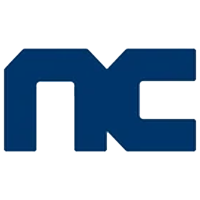} \texttt{VARCO-Vision-2.0-14B} & \textsf{K} & \textsf{V} &  & 58.7 & 11.24 & 57.5 & 59.5 & 42.7 & 59.0 & 62.3 & 54.3 & 61.7 & 56.0 \\
   \rowcolor{palegray}
   \cellcolor{white} & \includegraphics[height=0.8em]{icon/skt.png} \texttt{A.X-4.0-Light} & \textsf{K} &  &  & 55.7 & 11.45 & 60.6 & 55.7 & - & 55.8 & 54.4 & 50.9 & 61.4 & 55.7 \\
   & \includegraphics[height=0.8em]{icon/skt.png} \texttt{A.X-4.0-VL-Light} & \textsf{K} & \textsf{V} &  & 52.5 & 9.76 & 54.8 & 53.0 & 42.7 & 51.5 & 50.6 & 50.1 & 58.0 & 52.1 \\
   
   \rowcolor{palegray}
   \cellcolor{white} \multirow{-17}{*}{\textbf{Open-source}} 
   & \includegraphics[height=0.8em]{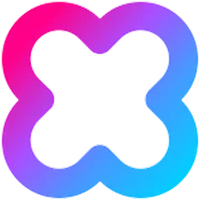} \texttt{HyperCLOVAX-SEED-Think-14B} & \textsf{K} &  & \textsf{R} & 50.8 & 11.29 & 52.1 & 50.8 & - & 51.6 & 53.8 & 41.8 & 55.6 & 51.1 \\
   
  \bottomrule
  \end{tabular}
  \end{adjustbox}
\end{table*}

\begin{figure}[ht]
    \centering
    \includegraphics[width=\linewidth]{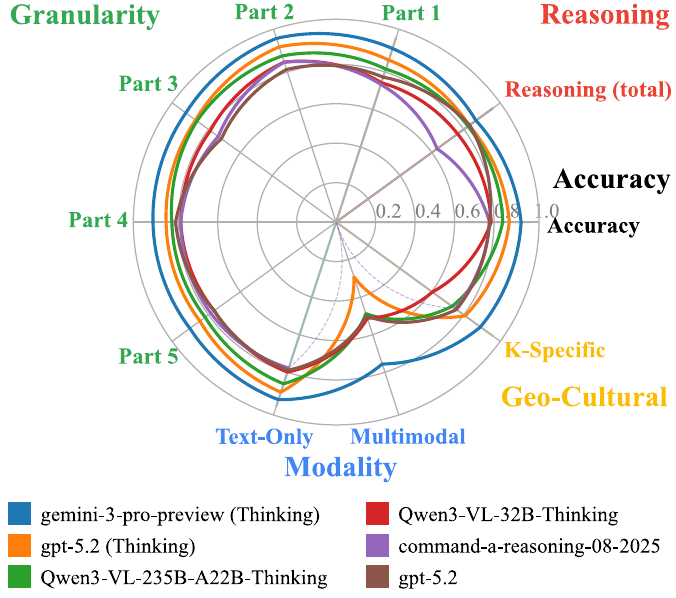}
    \caption{\textbf{Holistic performance analysis of top-6 models across five dimensions.} The radar chart visualizes model capabilities in Accuracy, \googr{Reasoning}, \googy{Geo-Cultural alignment (K-Specific)}, \googb{Modality (Text-only vs. Multimodal)}, and \googg{Granularity (Subject Parts 1–5)}. While models show balanced performance across theoretical subjects, a sharp decline is observed in the \textit{Multimodal} axis, highlighting the modality gap.}
    \label{fig:radar}
\end{figure}

\paragraph{The Modality Gap: Text-Only vs. Multimodal.}
Figure~\ref{fig:radar} reveals a distinct \textit{dented} shape along the \textit{Multimodal} axis, confirming that visual reasoning is the primary bottleneck for current MLLMs. Specifically, models exhibited a sharp accuracy decline (avg. $-$18.55\%) on multimodal questions compared to text-only ones. This deficit is most pronounced in professional tasks involving Skew-T Log-P diagrams and surface weather maps, where models failed to extract key data despite their general vision capabilities.

\paragraph{The Reasoning Gap: Knowledge vs. Reasoning.}
Table~\ref{tab:kmetbench_scores} and Figure~\ref{fig:radar} highlight a distinct gap between answer accuracy and reasoning quality. Although Kendall's $\tau_b$ (0.78) indicates a general correlation (Appendix Figure~\ref{fig:orthogonal_overall}), qualitative analysis reveals that models frequently provide correct answers accompanied by insufficient rationales, including the use of improper or hallucinated terminology (Appendix Table~\ref{tab:reasoning_case_study}). Additionally, while models achieve high accuracy on simple retrieval tasks, performance significantly degrades on calculation and multi-step reasoning tasks, even when CoT prompting—explicitly guiding the model to use a \texttt{<scratchpad>}~\citep{nye2021scratchpads} —is employed.

\paragraph{The Geo-Cultural Gap.} Table~\ref{tab:kmetbench_scores} reveals that large multilingual models struggle with the \textit{Korean-Specific} subset (e.g., Changma, topography) despite their scale. The Korean-centric \texttt{A.X-4.0} (72B) scored 78.9, outperforming the larger \texttt{Qwen3-VL-235B-Thinking} (72.6). This confirms that parameter scaling does not automatically grant proficiency in local domains.


\paragraph{Granular Domain Analysis.}
Finally, decomposing performance across the five official subject areas reveals fine-grained disparities masked by aggregated scores. As shown in Table~\ref{tab:kmetbench_scores}, models generally exhibit robust performance in Part 2 (\textit{Meteorological Observation}), which focuses on instrumentation and factual knowledge (e.g., \texttt{Gemini-3-Pro} reaching 97.9). However, significant performance drops are observed in calculation-intensive and abstract domains like Part 3 (\textit{Atmospheric Dynamics}) and Part 5 (\textit{Atmospheric Physics}). A striking example is the Korean model \texttt{A.X-4.0}, which achieves its highest accuracy in Part 4 (\textit{Climatology}) (81.3)—likely benefiting from training on local meteorological laws—but struggles disproportionately in Part 3 (68.2), where understanding synoptic motions is required. This granular diagnosis identifies specific domain weaknesses: while models may possess sufficient regulatory knowledge (Part 4), they require targeted fine-tuning to enhance quantitative reasoning in thermodynamics and dynamics (Part 3, 5).

\begin{figure}[!t]
    \centering
    \includegraphics[width=\linewidth]{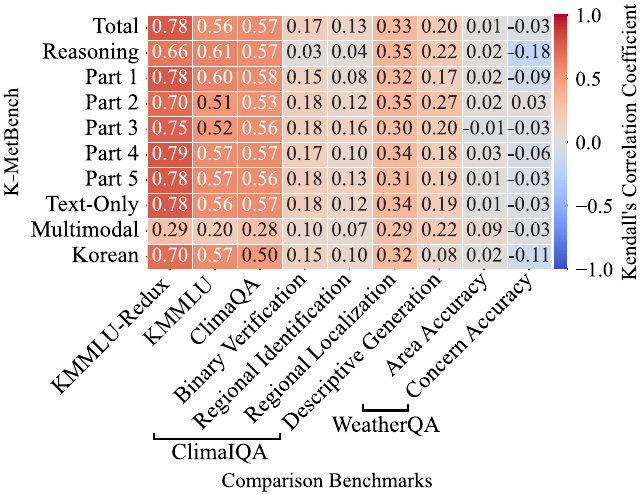}
    \caption{\textbf{Correlation analysis with existing benchmarks.} The heatmap visualizes Kendall's $\tau_b$ correlation coefficients between K-MetBench metrics and existing benchmarks.}
    \label{fig:orthogonality}
\end{figure}

\paragraph{Orthogonality between Existing Baselines.}
As shown in Figure~\ref{fig:orthogonality}, we analyzed Kendall’s $\tau_b$ correlations to assess the independence of K-MetBench. While the \textit{Text-Only} subset correlates strongly with general Korean benchmarks (KMMLU-Redux, $\tau_b=0.78$), we observe a distinct decoupling in complex capabilities. Notably, the correlation weakens for the \textit{Reasoning} subset ($\tau_b=0.66$) and drops sharply for the \textit{Multimodal} subset ($\tau_b=0.29$). Furthermore, correlations with external weather baselines (e.g., ClimaQA, ClimaIQA, and WeatherQA) remain consistently low across both reasoning and multimodal dimensions (avg. $\tau_b < 0.14$). This quantitative gap demonstrates that K-MetBench evaluates specialized domain logic and visual interpretation skills that are orthogonal to general linguistic proficiency and existing meteorological tasks.

\begin{figure}[ht]
    \centering
    \includegraphics[width=\linewidth]{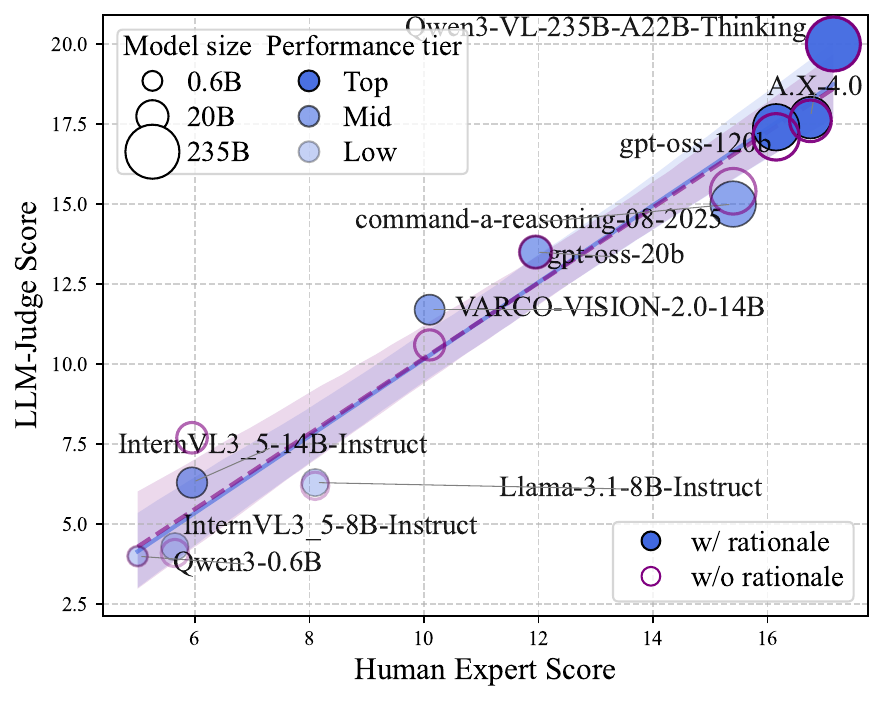}
    \caption{\textbf{Scatter plot comparing human expert vs. LLM-judge scores.} The \textit{w/ rationale} condition ($\tau_b=0.99$) shows slightly higher precision and lower variance than the \textit{w/o rationale} condition ($\tau_b=0.96$), while both maintain a strong correlation.}
    \label{fig:llm_judge_scatter}
\end{figure}

\paragraph{Meta-Evaluation: Human-LLM Agreement.}
We validated our reasoning evaluation framework by measuring inter-rater agreement on 100 sampled responses (Table~\ref{tab:inter_rater_agreement}). All axes surpassed the reliability threshold ($\alpha > 0.7$), with \textit{Reasoning Total} achieving a robust $\alpha$ of 0.838. Additionally, Figure~\ref{fig:llm_judge_scatter} illustrates a strong correlation between human and LLM scores. The \textit{w/ rationale} setting yielded a Kendall's $\tau_b$ of 0.99 with low variance, slightly outperforming the \textit{w/o rationale} setting ($\tau_b=0.96$).

\begin{table}[ht]
\centering
\caption{\textbf{Inter-rater agreement analysis.} The agreement between the average scores of two human experts and the LLM evaluator. We report Krippendorff's $\alpha$ (interval) and Intraclass Correlation Coefficient (ICC, two-way mixed, absolute agreement).}
\label{tab:inter_rater_agreement}
\resizebox{\linewidth}{!}{
\setlength{\tabcolsep}{4pt}
\begin{tabular}{l c c c}
\toprule
\textbf{Evaluation Axis} & \textbf{Krippendorff's $\alpha$} & \textbf{ICC} & \textbf{$N$} \\
\midrule
\textbf{Factuality}     & 0.827 & 0.829 & 100 \\
\textbf{Logicality}     & 0.827 & 0.830 & 100 \\
\textbf{Depth}          & 0.742 & 0.747 & 100 \\
\textbf{Clarity}        & 0.825 & 0.827 & 100 \\
\hline
\textbf{Reasoning Total}& 0.838 & 0.841 & 100 \\
\bottomrule
\end{tabular}
}
\end{table}
\section{Discussion}

\subsection{The Challenge of Visual Reasoning in Specialized Domains}
\label{sec:discussion_modality}
The observed modality gap in Table~\ref{tab:kmetbench_scores} and Figure~\ref{fig:radar} underscores a fundamental limitation: current MLLMs lack the \textit{domain-specific visual literacy} necessary for forecasting. Although proficient in general recognition, models struggle to ground specialized visual patterns—such as isobars, fronts, and wind barbs—in physical principles. This indicates that training on general image-text pairs is insufficient for mastering the fine-grained visual reasoning required in specialized scientific domains.

\subsection{Geo-Cultural Alignment in Meteorology} \label{sec:discussion_korean} 
Meteorology requires applying universal laws to localized contexts. The observed performance gap indicates a critical lack of \textit{geo-cultural alignment} in global models. 
Despite linguistic fluency, multilingual models frequently hallucinate on specific Korean geographic and terminological nuances. Consequently, effective deployment in vertical domains demands more than mere scaling; it requires rigorous alignment with local topographic and legal contexts to bridge the gap between general capability and expert-level application.

\subsection{Superficial Reasoning vs. Causal Deduction}
\label{sec:discussion_reasoning}
The observation that models output correct answers with shallow or erroneous explanations points to \textit{shortcut learning}~\cite{geirhos2020shortcut}—a reliance on surface-level associations rather than genuine understanding. Furthermore, the inability to reach expert-level performance on formula-based problems (e.g., calculating geostrophic wind speed) highlights a critical deficiency in applying physical laws. Addressing this requires shifting from general instruction tuning to training on high-quality reasoning trace data grounded in rigorous physical principles.

\subsection{Reliability of Automated Evaluation in Specialized Domains}
Our results confirm that \texttt{Gemini-2.5-Pro} is a reliable proxy for human experts in meteorology. The high agreement in \textit{Factuality} and \textit{Logicality} in Table~\ref{tab:inter_rater_agreement} demonstrates objective evaluation of logic and evidence. While \textit{Depth} showed slightly more subjectivity, the overall consistency supports the framework's robustness. Furthermore, the tight correlation observed in the scatter plot in Figure~\ref{fig:llm_judge_scatter} indicates that expert rationales effectively minimize variance. However, the model's high intrinsic knowledge ensures reliable grading even in their absence. These findings demonstrate that, when guided by high-quality rubrics, modern LLMs are cost-effective and reliable judges even in fine-grained domains like meteorology. This validates adopting the LLM-as-a-Judge framework for the reasoning evaluation in this study.

\section{Conclusion}

We present K-MetBench, a multi-dimensional benchmark for fine-grained evaluation of large language models in meteorological reasoning. By decomposing performance across modality, reasoning quality, geo-cultural context, and domain-specific sub-fields, K-MetBench provides diagnostic insights that are not observable from aggregate accuracy alone.
Our evaluation reveals persistent challenges in interpreting domain-specific visual artifacts, producing coherent expert-level rationales, and grounding meteorological knowledge in local context. In addition, analysis across official subject areas exposes uneven performance that is obscured by holistic scores.
Overall, K-MetBench is intended as a diagnostic complement to existing benchmarks, helping identify where current models succeed and where targeted improvements are needed for reliable deployment in specialized scientific domains.


\section*{Limitations} 
While K-MetBench serves as a rigorous diagnostic tool for meteorological AI, we acknowledge several limitations. 
First, regarding modality, the benchmark focuses on static visual reasoning (e.g., snapshot weather charts). While interpreting these charts is fundamental to forecasting, the current dataset does not evaluate the temporal reasoning required to interpret atmospheric evolution, such as sequential radar imagery or satellite loops.
Second, the dataset is geo-specifically rooted in the Korean context. Although this design effectively evaluates geo-cultural alignment—a key contribution of our work—it inherently limits direct generalizability to other climatic regions without adaptation. 
Finally, we utilized the official examination passing criteria (60\%) as a proxy for human competency. While this provides a validated baseline for qualification, a fine-grained human expert ceiling (e.g., the upper-bound score of top-tier meteorologists) was not explicitly measured in this study. Future work will focus on establishing this upper bound to quantify the `super-human' gap precisely.

\section*{Ethical Considerations} 
We adhered to copyright laws and ethical guidelines in constructing K-MetBench. The dataset is derived from National Meteorological Engineer examinations administered from March 16, 2003 to March 5, 2022; among 43 sessions in this period, we used only the 25 that were officially released to the public. We also obtained explicit permission from the Human Resources Development Service of Korea (HRDK) to use these materials for research and to release the refined dataset in an open repository. In addition, the dataset was reviewed to ensure that it contains no personally identifiable information or harmful content.

For human annotation, we involved two domain experts from collaborating institutions in the same funded project: one university professor and one research professor. The same experts conducted both reference-rationale verification and scoring of model-generated reasoning, and these activities were compensated separately on a per-item basis in accordance with our institution’s internal standards for expert advisory and review work. We consider this compensation appropriate given the experts’ seniority, domain expertise, and expected time commitment.

\section*{Licensing and Legal Compliance} 
The K-MetBench dataset is derived from public examination materials managed by the HRDK. We conducted a rigorous legal review to ensure compliance with the \textit{Official Information Disclosure Act} and relevant copyright laws (Copyright Act Art. 24-2, 25) in Korea . We confirmed that the questions are not classified as restricted information. To support the research community, the curated dataset is released via an open repository under the CC BY-NC-ND license, permitting non-commercial research use while preserving the integrity of the original artifacts.


\section*{Acknowledgments} 
We express our gratitude to the Human Resources Development Service of Korea (HRDK) for allowing the use of National Technical Qualification Examination data for research purposes. We would like to thank Seongsu Bae and the anonymous reviewers for their valuable comments. 

This research was supported by the High-Performance Computing Support Project, funded by the Ministry of Science and ICT (MSIT) and the National IT Industry Promotion Agency (NIPA) under grant No. RQT-25-070278 (providing 40 H100 GPUs). This work was also supported by the Institute for Information \& Communications Technology Planning \& Evaluation (IITP) grant funded by the Korea government (MSIT) (No. RS-2019-II190075, Artificial Intelligence Graduate School Program (KAIST); and No. RS-2022-II220984, Development of Artificial Intelligence Technology for Personalized Plug-and-Play Explanation and Verification of Explanation), and by the Korea Meteorological Administration (KMA) and National Institute of Meteorological Sciences (NIMS) under grant No. KMA2021-00123 (Developing Intelligent Assistant Technology and Its Application for Weather Forecasting Process).

\section*{Data and Code Availability}
The dataset is hosted on HuggingFace at 
\url{https://huggingface.co/datasets/soyeonbot/K-MetBench}. 
The evaluation toolkit is available at 
\url{https://github.com/kmetbench/kmetbench-release}.
The K-MetBench leaderboard is publicly available at 
\url{https://kmetbench.github.io/}.

\bibliography{custom}

@inproceedings{hong2025kmmlu,
    title = "From {KMMLU}-Redux to Pro: A Professional {K}orean Benchmark Suite for {LLM} Evaluation",
    author = "Hong, Seokhee  and
      Kim, Sunkyoung  and
      Son, Guijin  and
      Kim, Soyeon  and
      Hong, Yeonjung  and
      Lee, Jinsik",
    editor = "Christodoulopoulos, Christos  and
      Chakraborty, Tanmoy  and
      Rose, Carolyn  and
      Peng, Violet",
    booktitle = "Findings of the Association for Computational Linguistics: EMNLP 2025",
    month = nov,
    year = "2025",
    address = "Suzhou, China",
    publisher = "Association for Computational Linguistics",
    url = "https://aclanthology.org/2025.findings-emnlp.1038/",
    doi = "10.18653/v1/2025.findings-emnlp.1038",
    pages = "19067--19096",
    ISBN = "979-8-89176-335-7",
}

@inproceedings{son2025kmmlu,
    title = "{KMMLU}: Measuring Massive Multitask Language Understanding in {K}orean",
    author = "Son, Guijin  and
      Lee, Hanwool  and
      Kim, Sungdong  and
      Kim, Seungone  and
      Muennighoff, Niklas  and
      Choi, Taekyoon  and
      Park, Cheonbok  and
      Yoo, Kang Min  and
      Biderman, Stella",
    editor = "Chiruzzo, Luis  and
      Ritter, Alan  and
      Wang, Lu",
    booktitle = "Proceedings of the 2025 Conference of the Nations of the Americas Chapter of the Association for Computational Linguistics: Human Language Technologies (Volume 1: Long Papers)",
    month = apr,
    year = "2025",
    address = "Albuquerque, New Mexico",
    publisher = "Association for Computational Linguistics",
    url = "https://aclanthology.org/2025.naacl-long.206/",
    doi = "10.18653/v1/2025.naacl-long.206",
    pages = "4076--4104",
    ISBN = "979-8-89176-189-6",
}

@article{manivannan2024climaqa,
  title={ClimaQA: An Automated Evaluation Framework for Climate Question Answering Models},
  author={Manivannan, Veeramakali Vignesh and Jafari, Yasaman and Eranky, Srikar and Ho, Spencer and Yu, Rose and Watson-Parris, Duncan and Ma, Yian and Bergen, Leon and Berg-Kirkpatrick, Taylor},
  journal={arXiv preprint arXiv:2410.16701},
  year={2024}
}

@inproceedings{chen2025climateiqa,
  title={ClimateIQA: A New Dataset and Benchmark to Advance Vision-Language Models in Meteorology Anomalies Analysis},
  author={Chen, Jian and Zhou, Peilin and Hua, Yining and Chong, Dading and Cao, Meng and Li, Yaowei and Chen, Wei and Zhu, Bing and Liang, Junwei and Yuan, Zixuan},
  booktitle={Proceedings of the 31st ACM SIGKDD Conference on Knowledge Discovery and Data Mining V. 2},
  pages={5322--5333},
  year={2025}
}

@article{ma2024weatherqa,
  title={Weatherqa: Can multimodal language models reason about severe weather?},
  author={Ma, Chengqian and Hua, Zhanxiang and Anderson-Frey, Alexandra and Iyer, Vikram and Liu, Xin and Qin, Lianhui},
  journal={arXiv preprint arXiv:2406.11217},
  year={2024}
}

@article{team2023gemini,
  title={Gemini: a family of highly capable multimodal models},
  author={Team, Gemini and Anil, Rohan and Borgeaud, Sebastian and Alayrac, Jean-Baptiste and Yu, Jiahui and Soricut, Radu and Schalkwyk, Johan and Dai, Andrew M and Hauth, Anja and Millican, Katie and others},
  journal={arXiv preprint arXiv:2312.11805},
  year={2023}
}

@misc{openai2025gpt52,
  title={Update to GPT-5 System Card: GPT-5.2},
  author={{OpenAI}},
  year={2025},
  url={{https://cdn.openai.com/pdf/3a4153c8-c748-4b71-8e31-aecbde944f8d/oai_5_2_system-card.pdf}},
  note={Accessed: 2026-01-04}
}

@article{yoo2024hyperclova,
  title={Hyperclova x technical report},
  author={Yoo, Kang Min and Han, Jaegeun and In, Sookyo and Jeon, Heewon and Jeong, Jisu and Kang, Jaewook and Kim, Hyunwook and Kim, Kyung-Min and Kim, Munhyong and Kim, Sungju and others},
  journal={arXiv preprint arXiv:2404.01954},
  year={2024}
}

@article{yang2025qwen3,
  title={Qwen3 technical report},
  author={Yang, An and Li, Anfeng and Yang, Baosong and Zhang, Beichen and Hui, Binyuan and Zheng, Bo and Yu, Bowen and Gao, Chang and Huang, Chengen and Lv, Chenxu and others},
  journal={arXiv preprint arXiv:2505.09388},
  year={2025}
}

@article{qwen2024qwen2,
  title={Qwen2. 5 technical report},
  author={Qwen, A Yang and Yang, Baosong and Zhang, B and Hui, B and Zheng, B and Yu, B and Li, Chengpeng and Liu, D and Huang, F and Wei, H and others},
  journal={arXiv preprint},
  year={2024}
}

@article{bai2025qwen2,
  title={Qwen2. 5-vl technical report},
  author={Bai, Shuai and Chen, Keqin and Liu, Xuejing and Wang, Jialin and Ge, Wenbin and Song, Sibo and Dang, Kai and Wang, Peng and Wang, Shijie and Tang, Jun and others},
  journal={arXiv preprint arXiv:2502.13923},
  year={2025}
}

@article{agarwal2025gpt,
  title={gpt-oss-120b \& gpt-oss-20b model card},
  author={Agarwal, Sandhini and Ahmad, Lama and Ai, Jason and Altman, Sam and Applebaum, Andy and Arbus, Edwin and Arora, Rahul K and Bai, Yu and Baker, Bowen and Bao, Haiming and others},
  journal={arXiv preprint arXiv:2508.10925},
  year={2025}
}

@article{cohere2025command,
  title={Command a: An enterprise-ready large language model},
  author={Cohere, Team and Ahmadian, Arash and Ahmed, Marwan and Alammar, Jay and Alizadeh, Milad and Alnumay, Yazeed and Althammer, Sophia and Arkhangorodsky, Arkady and Aryabumi, Viraat and Aumiller, Dennis and others},
  journal={arXiv preprint arXiv:2504.00698},
  year={2025}
}

@article{cha2025varco,
  title={Varco-vision-2.0 technical report},
  author={Cha, Young-rok and Ju, Jeongho and Park, SunYoung and Lee, Jong-Hyeon and Yu, Younghyun and Kim, Youngjune},
  journal={arXiv preprint arXiv:2509.10105},
  year={2025}
}

@misc{skt2025adotx4,
  title={{A.X 4.0}},
  author={SKT AI Model Lab},
  year={2025},
  url={{https://huggingface.co/skt/A.X-4.0}},
}

@article{research2025exaone,
  title={EXAONE 4.0: Unified Large Language Models Integrating Non-reasoning and Reasoning Modes},
  author={Research, LG and Bae, Kyunghoon and Choi, Eunbi and Choi, Kibong and Choi, Stanley Jungkyu and Choi, Yemuk and Han, Kyubeen and Hong, Seokhee and Hwang, Junwon and Hwang, Taewan and others},
  journal={arXiv preprint arXiv:2507.11407},
  year={2025}
}

@article{wang2025internvl3,
  title={Internvl3. 5: Advancing open-source multimodal models in versatility, reasoning, and efficiency},
  author={Wang, Weiyun and Gao, Zhangwei and Gu, Lixin and Pu, Hengjun and Cui, Long and Wei, Xingguang and Liu, Zhaoyang and Jing, Linglin and Ye, Shenglong and Shao, Jie and others},
  journal={arXiv preprint arXiv:2508.18265},
  year={2025}
}

@misc{meta2024llama32,
  title = {Llama 3.2 Model Card},
  author = {{Meta}},
  year = {2024},
  url = {{https://github.com/meta-llama/llama-models/blob/main/models/llama3_2/MODEL_CARD.md}},
  note = {Accessed: 2024-01-04}
}

@inproceedings{kwon2023efficient,
  title={Efficient memory management for large language model serving with pagedattention},
  author={Kwon, Woosuk and Li, Zhuohan and Zhuang, Siyuan and Sheng, Ying and Zheng, Lianmin and Yu, Cody Hao and Gonzalez, Joseph and Zhang, Hao and Stoica, Ion},
  booktitle={Proceedings of the 29th symposium on operating systems principles},
  pages={611--626},
  year={2023}
}

@article{taylor2022galactica,
  title={Galactica: A large language model for science},
  author={Taylor, Ross and Kardas, Marcin and Cucurull, Guillem and Scialom, Thomas and Hartshorn, Anthony and Saravia, Elvis and Poulton, Andrew and Kerkez, Viktor and Stojnic, Robert},
  journal={arXiv preprint arXiv:2211.09085},
  year={2022}
}

@article{katz2024gpt,
  title={Gpt-4 passes the bar exam},
  author={Katz, Daniel Martin and Bommarito, Michael James and Gao, Shang and Arredondo, Pablo},
  journal={Philosophical Transactions of the Royal Society A},
  volume={382},
  number={2270},
  pages={20230254},
  year={2024},
  publisher={The Royal Society}
}

@article{singhal2023large,
  title={Large language models encode clinical knowledge},
  author={Singhal, Karan and Azizi, Shekoofeh and Tu, Tao and Mahdavi, S Sara and Wei, Jason and Chung, Hyung Won and Scales, Nathan and Tanwani, Ajay and Cole-Lewis, Heather and Pfohl, Stephen and others},
  journal={Nature},
  volume={620},
  number={7972},
  pages={172--180},
  year={2023},
  publisher={Nature Publishing Group}
}

@article{liang2022holistic,
  title={Holistic evaluation of language models},
  author={Liang, Percy and Bommasani, Rishi and Lee, Tony and Tsipras, Dimitris and Soylu, Dilara and Yasunaga, Michihiro and Zhang, Yian and Narayanan, Deepak and Wu, Yuhuai and Kumar, Ananya and others},
  journal={arXiv preprint arXiv:2211.09110},
  year={2022}
}

@article{turpin2023language,
  title={Language models don't always say what they think: Unfaithful explanations in chain-of-thought prompting},
  author={Turpin, Miles and Michael, Julian and Perez, Ethan and Bowman, Samuel},
  journal={Advances in Neural Information Processing Systems},
  volume={36},
  pages={74952--74965},
  year={2023}
}

@article{zheng2023judging,
  title={Judging llm-as-a-judge with mt-bench and chatbot arena},
  author={Zheng, Lianmin and Chiang, Wei-Lin and Sheng, Ying and Zhuang, Siyuan and Wu, Zhanghao and Zhuang, Yonghao and Lin, Zi and Li, Zhuohan and Li, Dacheng and Xing, Eric and others},
  journal={Advances in neural information processing systems},
  volume={36},
  pages={46595--46623},
  year={2023}
}

@article{li2024llms,
  title={Llms-as-judges: a comprehensive survey on llm-based evaluation methods},
  author={Li, Haitao and Dong, Qian and Chen, Junjie and Su, Huixue and Zhou, Yujia and Ai, Qingyao and Ye, Ziyi and Liu, Yiqun},
  journal={arXiv preprint arXiv:2412.05579},
  year={2024}
}

@misc{hrdk2024meteorological,
  author       = {{Human Resources Development Service of Korea}},
  title        = {Examination Standards for Meteorological Engineer (2023.1.1--2026.12.31)},
  year         = {2024},
  howpublished = {\url{https://www.q-net.or.kr/pageLink.do?link=cst/cstReport}},
  note         = {Accessed: 2026-01-06. Available at Q-Net}
}

@article{wei2022chain,
  title={Chain-of-thought prompting elicits reasoning in large language models},
  author={Wei, Jason and Wang, Xuezhi and Schuurmans, Dale and Bosma, Maarten and Xia, Fei and Chi, Ed and Le, Quoc V and Zhou, Denny and others},
  journal={Advances in neural information processing systems},
  volume={35},
  pages={24824--24837},
  year={2022}
}

@article{geirhos2020shortcut,
  title={Shortcut learning in deep neural networks},
  author={Geirhos, Robert and Jacobsen, J{\"o}rn-Henrik and Michaelis, Claudio and Zemel, Richard and Brendel, Wieland and Bethge, Matthias and Wichmann, Felix A},
  journal={Nature Machine Intelligence},
  volume={2},
  number={11},
  pages={665--673},
  year={2020},
  publisher={Nature Publishing Group UK London}
}

@article{nye2021scratchpads,
  title={Show Your Work: Scratchpads for Intermediate Computation with Language Models},
  author={Nye, Maxwell and Andreassen, Anders Johan and Gur-Ari, Guy and Michalewski, Henryk and Austin, Jacob and Bieber, David and Dohan, David and Lewkowycz, Aitor and Bosma, Maarten and Luan, David and Sutton, Charles and Odena, Augustus},
  journal={arXiv preprint arXiv:2112.00114},
  year={2021},
  url={https://arxiv.org/abs/2112.00114}
}

@inproceedings{huang2026dropping,
    title={Dropping Just a Handful of Preferences Can Change Top Large Language Model Rankings},
    author={Jenny Y. Huang and Yunyi Shen and Dennis Wei and Tamara Broderick},
    booktitle={The Fourteenth International Conference on Learning Representations},
    year={2026},
    url={https://openreview.net/forum?id=jNiEMDsRgc}
}

@inproceedings{lee2021deduplicating,
  author    = {Katherine Lee and Daphne Ippolito and A. Nystrom and Chiyuan Zhang and D. Eck and Chris Callison-Burch and Nicholas Carlini},
  year      = {2021},
  title     = {Deduplicating Training Data Makes Language Models Better},
  booktitle = {Annual Meeting of the Association for Computational Linguistics},
  doi       = {10.18653/v1/2022.acl-long.577},
}
\clearpage

\appendix
\doparttoc 
\faketableofcontents 
\onecolumn 
\renewcommand{\partname}{}
\renewcommand{\thepart}{}
\part{Appendix} 
\parttoc
\twocolumn
\begin{table*}[ht]
    \centering
    \tiny
    \renewcommand{\arraystretch}{1.2} 
    \caption{\textbf{Representative examples of K-MetBench tasks.} The examples are organized by modality: \textbf{\googb{Text-only}} (Top) and \textbf{\googb{Multimodal}} (Bottom). We showcase three task types within each modality: \textbf{Standard} (fundamental knowledge), \textbf{\googy{K-Specific}} (geo-cultural context), and \textbf{\googr{Reasoning}} (complex deduction). \textbf{\googg{Part}} denotes the corresponding subject from the five official fields. \gtrans{Gray text} indicates English translations.}
    \label{tab:dataset_example}

    \begin{tabularx}{\linewidth}{l L L L}
        \toprule
        \textbf{\googb{Modality}} & \textbf{Standard MCQA} & \textbf{\googy{K-Specific MCQA}} & \textbf{\googr{Reasoning MCQA}} \\
        \midrule
        
        \textbf{\googb{Text-Only}} & 
        \textbf{ID:} 1535, \googg{\textbf{Part:} 5} \par\medskip
        
        \textbf{질문:} 상층 일기도의 활용에 대해 올바르게 설명한 것은? \par
        \gtrans{Question: Which of the following is a correct description regarding the utilization of upper-level weather charts?} \par\medskip
        
        \textbf{1.} 500 hPa 일기도의 한랭기압골에서 등온선의 진폭이 등고선의 진폭보다 클 경우에는 그 기압골의 후방에 약한 상승기류가 있고, 전방에 약한 하강기류가 있다. \par
        \textbf{2.} 300 hPa 면에서는 온도가 지형, 복사의 영향을 받으므로 전선분석이 용이하다. \par
        \textbf{3.} 300 hPa 제트기류 출구의 좌측에 하강기류, 우측에 상승기류가 있으며, 입구에서는 좌측에 상승기류, 우측에 하강기류가 있다. \par
        \textbf{4.} 500 hPa 기류가 지상 한랭전선에 수직으로 불면 이 전선은 활성으로서 악천이 나타난다. \par
        \gtrans{1. In a cold trough on a 500 hPa chart, if the amplitude of the isotherms is larger than the amplitude of the contours (geopotential height), there is a weak updraft behind the trough and a weak downdraft ahead of it.} \par
        \gtrans{2. On the 300 hPa surface, temperature is affected by topography and radiation, making frontal analysis easy.} \par
        \gtrans{3. At the exit of a 300 hPa jet stream, there is a downdraft on the left and an updraft on the right; at the entrance, there is an updraft on the left and a downdraft on the right.} \par
        \gtrans{4. If the 500 hPa airflow blows perpendicular to a surface cold front, the front becomes active and severe weather occurs.} \par\medskip
        
        \textbf{정답:} 1 \par
        \gtrans{Ground Truth: 1}
        
        &
        \textbf{ID:} 65, \googg{\textbf{Part: 5}} \par\medskip
        
        \textbf{질문:} 다음은 \textbf{\googy{한국 지역}}에 영향을 주는 고기압의 특성을 설명한 것이다. 내용이 옳지 않은 것은? \par
        \gtrans{Question: The following describes the characteristics of high-pressure systems affecting the \textbf{\googy{Korean region}}. Which statement is incorrect?} \par\medskip
        
        \textbf{1.} 시베리아 고기압은 겨울철의 춥고 건조한 날씨를 만든다. \par
        \textbf{2.} 오호츠크해 고기압은 동해안 지방의 고온현상을 일으킨다. \par
        \textbf{3.} 북태평양 고기압은 고온다습하며, 여름철의 무더운 날씨를 만든다. \par
        \textbf{4.} 이동성 고기압의 영향을 받으면 봄에는 따뜻한 날씨, 가을에는 맑은 날씨가 된다. \par
        \gtrans{1. The Siberian High creates cold and dry weather during the winter.} \par
        \gtrans{2. The Okhotsk Sea High causes high-temperature phenomena in the east coastal regions.} \par
        \gtrans{3. The North Pacific High is hot and humid, creating sweltering weather during the summer.} \par
        \gtrans{4. Under the influence of migratory highs, the weather becomes warm in spring and clear in autumn.} \par\medskip
        
        \textbf{정답:} 2 \par
        \gtrans{Ground Truth: 2}
        
        &
        \textbf{ID:} 18, \googg{\textbf{Part: 2}} \par\medskip
        
        \textbf{질문:} 비열의 차원을 올바르게 나타낸 것은 무엇입니까? \par
        \gtrans{Question: What is the correct dimensional representation of specific heat?} \par\medskip
        
        \textbf{1.} [\$L\textasciicircum 2T\textasciicircum 2\textbackslash theta\textasciicircum\{-1\}\$] \par
        \textbf{2.} [\$L\textasciicircum 2T\textasciicircum\{-2\}\textbackslash theta\textasciicircum\{-1\}\$] \par
        \textbf{3.} [\$ML\textasciicircum\{-1\}T\textasciicircum\{-2\}\$] \par
        \textbf{4.} [\$ML\textasciicircum 2T\textasciicircum\{-2\}\$] \par\medskip
        
        \textbf{\googr{전문가 검증 참조 자료: 비열은 단위 질량당 단위 온도 상승에 필요한 에너지로서 차원은 (에너지)/(질량·온도) = 
        \$(ML\textasciicircum 2T\textasciicircum\{-2\})/(M\textbackslash theta) = L\textasciicircum 2T\textasciicircum\{-2\}\textbackslash theta\textasciicircum\{-1\}\$ 
        이므로 2번이 맞고, 4번은 에너지 자체의 차원, 3번은 압력의 차원, 1번은 시간 지수가 부호가 반대라 틀립니다.}} \par
        \gtrans{Expert-Verified Rationale: Specific heat is the energy required to raise the temperature of a unit mass by one unit. Its dimension is (Energy)/(Mass · Temperature) = 
        \$(ML\textasciicircum 2T\textasciicircum\{-2\})/(M\textbackslash theta) = L\textasciicircum 2T\textasciicircum\{-2\}\textbackslash theta\textasciicircum\{-1\}\$. 
        Therefore, option 2 is correct. Option 4 represents the dimension of energy itself, option 3 represents the dimension of pressure, and option 1 is incorrect because the sign of the time exponent is reversed.} \par\medskip
        
        \textbf{정답:} 2 \par
        \gtrans{Ground Truth: 2}
        \\
        \midrule
        \textbf{\googb{Multimodal}} & 
        \textbf{ID:} 460, \googg{\textbf{Part: 3}} \par\medskip
        
        \textbf{질문:} 북반구에서 나타나는 지균풍(\$\textbackslash vec\{V\_\{g\}\}\$), 실제풍(\$\textbackslash vec\{V\}\$), 수평가속도(\$(\textbackslash frac\{d\textbackslash vec\{v\}\}\{dt\})\_\{H\}\$) 사이의 관계를 올바르게 표현한 그림은 어느 것인가? \par
        \gtrans{Question: Which figure correctly represents the relationship between the geostrophic wind (\$\textbackslash vec\{V\_\{g\}\}\$), the actual wind (\$\textbackslash vec\{V\}\$), and the horizontal acceleration (\$(\textbackslash frac\{d\textbackslash vec\{v\}\}\{dt\})\_\{H\}\$) in the Northern Hemisphere?} \par\medskip
        
        \textbf{1.} \includegraphics[height=0.8in, keepaspectratio]{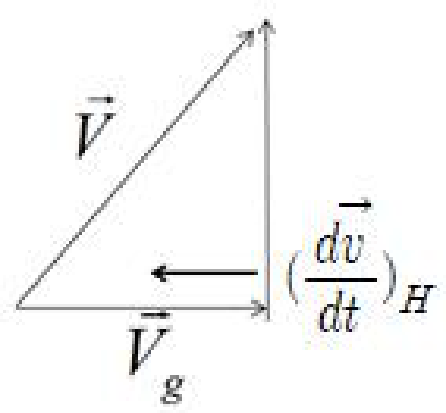} \par
        \textbf{2.} \includegraphics[height=0.8in, keepaspectratio]{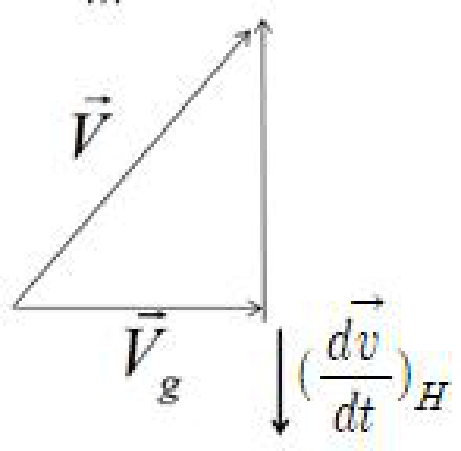} \par
        \textbf{3.} \includegraphics[height=0.8in, keepaspectratio]{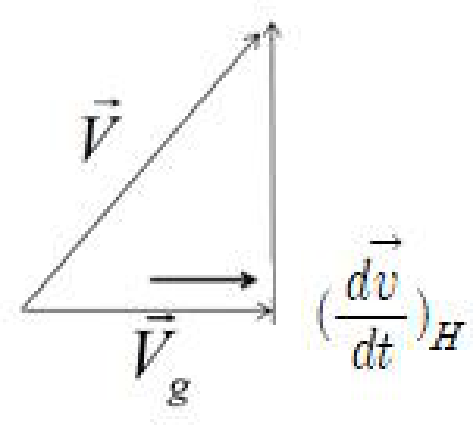} \par
        \textbf{4.} \includegraphics[height=0.8in, keepaspectratio]{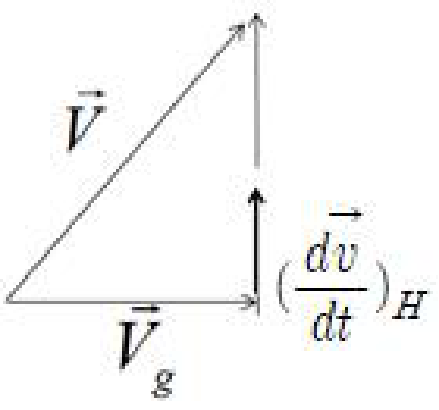} \par\medskip
        
        \textbf{정답:} 3 \par
        \gtrans{Ground Truth: 3}
        
        &
        \textbf{ID:} 687, \googg{\textbf{Part: 4}} \par\medskip
        
        \textbf{질문:} 제시된 그림은 \googy{\textbf{한국}}의 어떤 지점의 연평균 물수지를 보여준다. 이 그림에서 D 부분이 의미하는 것은 무엇인가? \par
        \gtrans{Question: The presented figure shows the annual average water balance of a certain location in \googy{\textbf{Korea}}. What does section D in this figure represent?} \par\medskip
        
        \includegraphics[width=0.9\linewidth]{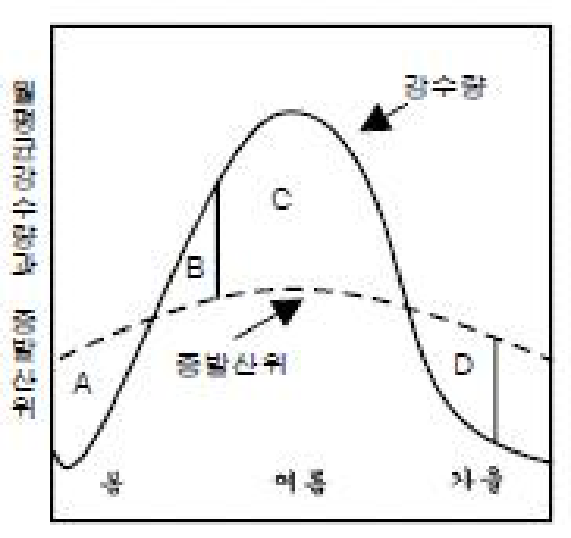} \par\medskip
        
        1. 토양수분의 과잉 \par
        2. 토양수분의 보충 \par
        3. 토양수분의 이용 \par
        4. 토양수분의 결핍 \par\
        \gtrans{1. Soil moisture surplus} \par
        \gtrans{2. Soil moisture recharge} \par
        \gtrans{3. Soil moisture utilization} \par
        \gtrans{4. Soil moisture deficit} \par\medskip
        
        \textbf{정답:} 3 \par
        \gtrans{Ground Truth: 3}
        
        &
        \textbf{ID:} 460, \googg{\textbf{Part: 3}} \par\medskip
        
        \textbf{질문:} 다음 그림이 보여주는 역전층의 종류로 옳은 것은? \par
        \gtrans{Question: Which of the following is the correct type of inversion layer shown in the figure below?} \par\medskip
        
        \includegraphics[width=0.9\linewidth]{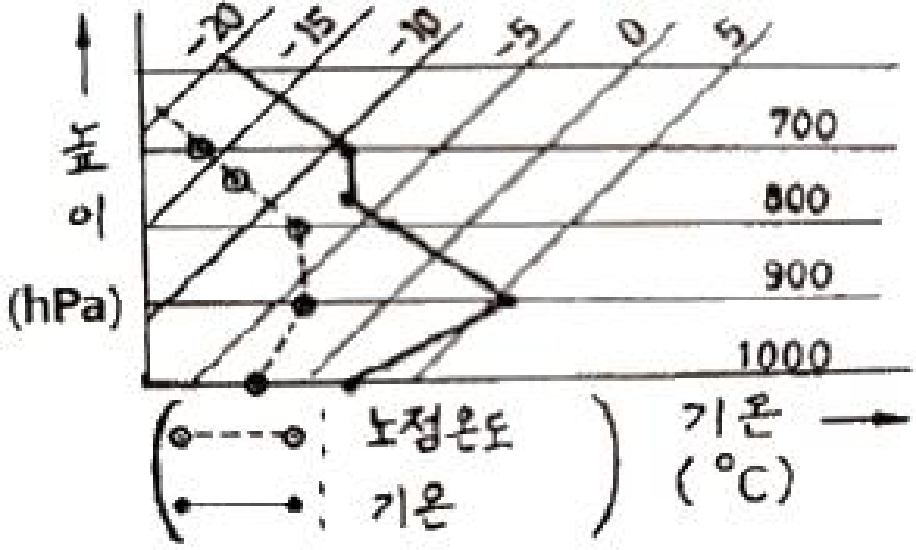} \par\medskip
        
        1. 복사역전 \par
        2. 난류역전 \par
        3. 전선역전 \par
        4. 침강역전 \par
        \gtrans{1. Radiation inversion} \par
        \gtrans{2. Turbulence inversion} \par
        \gtrans{3. Frontal inversion} \par
        \gtrans{4. Subsidence inversion} \par\medskip
        
        \textbf{\googr{전문가 검증 참조 자료: 그림처럼 지표에서 바로 시작하는 얕은 역전층이 위로 갈수록 약화되는 형태는 야간 지표 복사냉각으로 생기는 복사역전의 전형이며, 침강역전은 고기압 하 하강류로 상층에 분리되어 나타나고 전선역전은 전선면을 따라 경사져 있으며 난류역전은 주간 혼합층 꼭대기에 형성되어 지표에서 시작하지 않으므로 그림과 다르다.}} \par
        \gtrans{Expert-Verified Rationale: As shown in the figure, a shallow inversion layer starting directly from the surface and weakening with height is typical of a radiation inversion caused by nocturnal surface radiative cooling. In contrast, a subsidence inversion appears separated in the upper atmosphere due to downdrafts under high pressure, a frontal inversion is tilted along a frontal surface, and a turbulence inversion forms at the top of the daytime mixed layer rather than starting at the surface; thus, they differ from the figure.} \par\medskip
        
        \textbf{정답:} 1 \par
        \gtrans{Ground Truth: 1} 
        \\
        \bottomrule
    \end{tabularx}
\end{table*}
\begin{table*}[ht]
    \centering
    \caption{\textbf{Comparison of \texttt{InternVL3.5-8B-Instruct} and \texttt{Qwen3-VL-235B-A22B-Thinking} on K-MetBench reasoning question.} (\protect\noBox: Incorrect, \protect\yes: Correct, \colorbox{lightred}{shaded text}: Highlighted errors)}
    \label{tab:reasoning_case_study}
    \tiny
    \renewcommand{\arraystretch}{1.2} 
    \begin{tabularx}{\linewidth}{p{3.5cm} L p{4.5cm}}
        \toprule
        \textbf{Reasoning MCQA} & 
        \textbf{Reasoning Evaluation of \texttt{InternVL3.5-8B-Instruct}} & 
        \textbf{Reasoning Evaluation of \texttt{Qwen3-VL-235B-A22B-Thinking}} \\
        
        \midrule
        \textbf{ID:} 105, \googg{\textbf{Part:} 3} \par\medskip
        
        \textbf{질문:} 지균풍 관계식에서 기압경도와 공기밀도가 일정하다면, 풍속은 어떻게 결정되는가? \par
        \gtrans{Question: In the geostrophic wind formula, if the pressure gradient and air density are constant, how is the wind speed determined?} \par\medskip
        
        \textbf{1.} 1. 고도가 증가함에 따라 증가한다. \par
        \textbf{2.} 2. 위도의 증가에 따라 감소한다. \par
        \textbf{3.} 3. 기압의 증가에 따라 증가한다. \par
        \textbf{4.} 4. 공기의 점성이 크면 증가한다. \par
        \gtrans{1. It increases as altitude increases.} \par
        \gtrans{2. It decreases as latitude increases.} \par
        \gtrans{3. It increases as pressure increases.} \par
        \gtrans{4. It increases if air viscosity is high.} \par\medskip

        \googr{\textbf{전문가 검증 참조 자료:} 지균풍은 Vg = |∇p|/(ρ f)이므로 기압경도와 밀도가 일정하면 코리올리매개변수 f=2Ω sinφ만이 변수가 되어 위도가 증가할수록 f가 커져 풍속은 감소하며, 절대 기압의 크기(보기 3)나 고도(보기 1), 점성(보기 4)은 이 관계식에 직접 등장하지 않는다.} \par
        \gtrans{Expert-Verified Rationale: Since the geostrophic wind is defined as Vg = |∇p|/(ρ f), if the pressure gradient and density are constant, the Coriolis parameter f=2Ω sinφ becomes the only variable. As latitude increases, f increases, causing the wind speed to decrease. The magnitude of absolute pressure (Option 3), altitude (Option 1), and viscosity (Option 4) do not appear directly in this equation.} \par\medskip
        
        \textbf{정답:} 2 \par
        \gtrans{Ground Truth: 2}
        
        &
        
        \textbf{Response from \texttt{InternVL3.5-8B-Instruct}} \par\medskip
        
        \textbf{생성된\_추론\_근거:} \colorbox{lightred}{지연풍} 관계식에서 풍속이 기압 경로의 변화와 직접적으로 연관된다는 것을 이해할 수 있습니다. \colorbox{lightred}{기압 경도}가 증가하면 풍속도 증가하는 경향이 있으며, 이는 풍속이 기압의 증가에 따라 증가한다는 3번 옵션이 정답이라는 근거를 제공합니다. \par
        \gtrans{Expert-Verified Rationale: From the \colorbox{lightred}{geostrophic wind relationship}, we can understand that wind speed is directly associated with changes in the pressure path. Since wind speed tends to increase when the \colorbox{lightred}{pressure gradient} increases, this provides grounds for Option 3—that wind speed increases as pressure increases—being the correct answer.} \par\medskip
        
        \textbf{정답:} \colorbox{lightred}{3} \noBox \par
        \gtrans{Predicted Answer: 3} \par \medskip

        \noindent\tikz \draw[dashed, line width=0.5pt, dash pattern=on 3pt off 2pt] (0,0) -- (\linewidth,0); \par\medskip
        
        \textbf{Reasoning Evaluation from Human Experts:}\par \medskip
        
        \textbf{사람 전문가 평균 점수:} 사실적 정확성 (1), 논리적 완결성 (1), 추론의 깊이 (1), 표현의 명확성 (1) \par
        \gtrans{\textbf{Average of Human Expert Score:} factuality (1), logicality (1), depth (1), clarity (1)} \par
        \textbf{평가 의견 1:} 오답 및 사용된 용어 부적절: \colorbox{lightred}{기압경로}, 지연풍 등 \par
        \textbf{평가 의견 2:} 문제에 맞지 않는 추론근거 및 답도 틀림. \par
        \gtrans{\textbf{Evaluation Comment 1:} Incorrect answer and inappropriate use of terminology: \colorbox{lightred}{`pressure path'}, `delayed wind', etc.} \par
        \gtrans{\textbf{Evaluation Comment 2:} The reasoning is irrelevant to the question, and the answer is incorrect.} \par \medskip

        \noindent\tikz \draw[dashed, line width=0.5pt, dash pattern=on 3pt off 2pt] (0,0) -- (\linewidth,0); \par\medskip
        
        \textbf{Reasoning Evaluation from LLM-Evaluator (\texttt{Gemini-2.5-Pro}):}\par \medskip
        
        \textbf{사실\_오류\_목록:} [\par
        지균풍(geostrophic wind)을 `지연풍'이라는 존재하지 않는 기상학 용어로 잘못 지칭하였습니다., \par
        문제에서 `기압경도가 일정하다'는 핵심 가정을 무시하고, `기압경도가 증가하면 풍속이 증가한다'는 관계를 추론의 근거로 사용하였습니다.\par
        `기압경도(pressure gradient)'와 `기압(pressure)'이라는 물리적으로 다른 개념을 혼동하여, 기압경도와 풍속의 관계를 근거로 기압과 풍속의 관계를 설명하는 오류를 범했습니다.\par
        ]\par
        \textbf{평가\_점수:} 정확성: 1, 논리적\_완결성: 1, 추론의\_깊이: 1, 표현의\_명확성: 1\par
        \textbf{평가\_사유:} 수험생 AI의 답변은 모든 평가 축에서 최하점을 기록했습니다. `정확성' 측면에서, `지균풍'을 `지연풍'이라는 환각 용어로 지칭했으며, `기압경도'와 `기압'을 혼동하는 치명적인 개념 오류를 보였습니다. 또한 문제의 핵심 전제인 `기압경도 일정'을 무시하여 사실 관계를 왜곡했습니다. `논리적 완결성' 측면에서, 문제의 조건과 무관한 일반론을 근거로 제시하고, `기압경도'와 `기압'을 혼동하여 결론을 도출했기 때문에 근거가 결론을 전혀 뒷받침하지 못합니다. `추론의 깊이' 측면에서, 지균풍의 핵심 원리인 코리올리 힘과 위도와의 관계에 대한 이해 없이 피상적인 사실을 나열하는 데 그쳤습니다. `표현의 명확성' 측면에서도, 존재하지 않는 용어와 모호한 표현(`기압 경로의 변화')을 사용하여 의미 전달에 심각한 결함이 있습니다.\par
        \gtrans{\textbf{List\_of\_Factual\_Errors:} [}\par
        \gtrans{It incorrectly referred to `geostrophic wind' as `delayed wind', a non-existent meteorological term.,} \par
        \gtrans{It disregarded the problem's key assumption that the `pressure gradient is constant' and used the relationship `wind speed increases as pressure gradient increases' as the basis for reasoning.,}\par
        \gtrans{It confused the distinct physical concepts of `pressure gradient' and `pressure', erroneously explaining the relationship between pressure and wind speed based on the relationship between pressure gradient and wind speed.,}\par
        \gtrans{]}
        \gtrans{\textbf{Evaluation\_Scores:} Accuracy: 1, Logical\_Completeness: 1, Depth\_of\_Reasoning: 1, Clarity\_of\_Expression: 1} \par
        \gtrans{\textbf{Evaluation\_Reasoning:} The candidate AI's response recorded the lowest scores across all evaluation axes. In terms of `Accuracy', it referred to `geostrophic wind' using the hallucinated term `delayed wind' and demonstrated a fatal conceptual error by confusing `pressure gradient' with `pressure'. Furthermore, it distorted factual relations by ignoring the problem's core premise that the `pressure gradient is constant'. In terms of `Logical\_Completeness', the reasoning failed to support the conclusion at all because it presented generalities irrelevant to the problem conditions and derived the conclusion by confusing `pressure gradient' with `pressure'. In terms of `Depth\_of\_Reasoning', it merely listed superficial facts without understanding the relationship between the Coriolis force and latitude, which is the core principle of geostrophic wind. In terms of `Clarity\_of\_Expression', there were serious flaws in meaning delivery due to the use of non-existent terms and ambiguous expressions (e.g., `change in pressure path').} \par \medskip

        & 
        \textbf{Response from \texttt{Qwen3-VL-235B-A22B-Thinking}}\par\medskip
        
        \textbf{생성된\_추론\_근거:} 지균풍은 Vg = |∇p|/(ρ f)이므로 기압경도와 밀도가 일정하면 코리올리매개변수 f=2Ω sinφ만이 변수가 되어 위도가 증가할수록 f가 커져 풍속은 감소하며, 절대 기압의 크기(보기 3)나 고도(보기 1), 점성(보기 4)은 이 관계식에 직접 등장하지 않는다. \par
        \gtrans{Expert-Verified Rationale: Since the geostrophic wind is defined as Vg = |∇p|/(ρ f), if the pressure gradient and density are constant, the Coriolis parameter f=2Ω sinφ becomes the only variable. As latitude increases, f increases, causing the wind speed to decrease. The magnitude of absolute pressure (Option 3), altitude (Option 1), and viscosity (Option 4) do not appear directly in this equation.} \par\medskip
        
        \textbf{정답:} 2 \yes \par 
        \gtrans{Predicted Answer: 2} \par \medskip

        \noindent\tikz \draw[dashed, line width=0.5pt, dash pattern=on 3pt off 2pt] (0,0) -- (\linewidth,0); \par\medskip
        
        \textbf{Reasoning Evaluation from Human Experts:}\par \medskip
        
        \textbf{사람 전문가 평균 점수:} 사실적 정확성 (5), 논리적 완결성 (5), 추론의 깊이 (4.5), 표현의 명확성 (5) \par
        \gtrans{\textbf{Average of Human Expert Score:} factuality (5), logicality (5), depth (4.5), clarity (5)} \par
        \gtrans{\textbf{Comments:}} \par \medskip

        \noindent\tikz \draw[dashed, line width=0.5pt, dash pattern=on 3pt off 2pt] (0,0) -- (\linewidth,0); \par\medskip
        
        \textbf{Reasoning Evaluation from LLM-Evaluator (\texttt{Gemini-2.5-Pro}):}\par \medskip
        
        \textbf{사실\_오류\_목록:} [ ] \par
        \textbf{평가\_점수:} 정확성: 5, 논리적\_완결성: 5, 추론의\_깊이: 5, 표현의\_명확성: 5\par
        \textbf{평가\_사유:} 수험생 AI의 답변은 모든 평가 축에서 매우 우수한 수준을 보였습니다. `정확성' 측면에서, 제시된 지균풍 속도 공식, 코리올리 계수와 위도의 관계, 그리고 결론까지 기상학적 사실에 완벽하게 부합하며 어떠한 오류도 발견되지 않았습니다. `논리적 완결성' 측면에서, 문제의 조건(기압경도, 밀도 일정)으로부터 풍속이 코리올리 계수에 반비례한다는 핵심 관계를 명확히 하고, 이를 위도와의 관계로 확장하여 결론을 도출하는 과정이 비약 없이 완벽하게 연결되었습니다. `추론의 깊이' 측면에서, 정답의 근거를 제시하는 것에 그치지 않고, 매력적인 오답 선택지(3번, 4번)가 왜 오답인지를 지균풍의 기본 가정(마찰 무시)과 핵심 변수(기압경도)를 들어 명확히 반박함으로써 문제에 대한 깊이 있는 이해를 보여주었습니다. `표현의 명확성' 측면에서도 군더더기 없이 핵심 내용을 간결하고 명확하게 전달하여 이해도를 높였습니다. 이는 모범적인 해설이라 할 수 있습니다.\par
        \gtrans{\textbf{List\_of\_Factual\_Errors:} [ ]} \par
        \gtrans{\textbf{Evaluation\_Scores:} Accuracy: 5, Logical\_Completeness: 5, Depth\_of\_Reasoning: 5, Clarity\_of\_Expression: 5} \par
        \gtrans{\textbf{Evaluation\_Reasoning:} The candidate AI's response demonstrated a very high level of performance across all evaluation axes. In terms of `Accuracy', the presented geostrophic wind speed formula, the relationship between the Coriolis parameter and latitude, and the conclusion perfectly matched meteorological facts with no errors found. In terms of `Logical\_Completeness', the process of clarifying the core relationship that wind speed is inversely proportional to the Coriolis parameter from the problem conditions (constant pressure gradient and density) and extending this to the relationship with latitude to derive the conclusion was perfectly connected without logical leaps. In terms of `Depth\_of\_Reasoning', it demonstrated a deep understanding of the problem not only by providing the basis for the correct answer but also by clearly refuting why the attractive distractors (Options 3 and 4) were incorrect, citing the basic assumption of geostrophic wind (ignoring friction) and the key variable (pressure gradient). In terms of `Clarity\_of\_Expression', it delivered the core content concisely and clearly without redundancy, enhancing comprehensibility. This can be considered an exemplary explanation.} \\
        \bottomrule
    \end{tabularx}
\end{table*}

\clearpage

\section{Dataset Examples}
\label{appx:dataset_example}

Table~\ref{tab:dataset_example} presents representative examples from K-MetBench, organized into two primary modality groups: \textit{Text-only} and \textit{Multimodal}. Within each modality, we further stratify the tasks into three distinct categories to evaluate comprehensive meteorological capabilities.

\textit{Text-only Tasks} assesses linguistic reasoning and theoretical knowledge without visual interpretation. This group includes (a) \textit{Standard MCQA} for fundamental concepts, (b) \textit{K-Specific MCQA} which requires geo-cultural knowledge specific to the Korean Peninsula, and (c) \textit{Reasoning MCQA} that demands multi-step logical deduction.

\textit{Multimodal Tasks} introduces visual data interpretation, a critical skill for meteorologists. This group parallels the text-only structure with (d) \textit{Standard}, (e) \textit{K-Specific}, and (f) \textit{Reasoning} subsets, but specifically evaluates the model's ability to analyze weather charts, satellite imagery, and atmospheric diagrams. This structured categorization allows for a clear comparison of model performance across different modalities and levels of domain expertise.

\section{Case Study of Reasoning Answer}
\label{appx:reasoning_case_study}
Two human experts and the LLM-Evaluator (\texttt{gemini-2.5-pro}) conducted evaluations using identical rubrics. 
As shown in Table~\ref{tab:reasoning_case_study}, we observed consensus between the human and AI evaluators for the \texttt{InternVL3.5-8B-Instruct} and \texttt{Qwen3-VL-235B-A22B-Thinking} models: both correctly identified incorrect answers and highlighted inappropriate terminology in the reasoning rationale.

Notably, the human expert made a specific error by misreading `기압 경도' (pressure gradient) as `기압경로' (pressure path). While the quantitative scores assigned by the human expert and the LLM-Evaluator were comparable, the granularity of their feedback differed significantly. The human expert made an implicit judgment, providing summary comments alongside the score. In contrast, the LLM-Evaluator generated more detailed outputs, including explicit justifications and comprehensive lists of factual errors.





\section{Prompts and Questionnaires for Benchmark Construction}
\label{appx:benchmark_prompts}

This section details the prompts utilized for data augmentation (paraphrasing) and the identification of domain-specific subsets. These processes were conducted to enhance the quality of the dataset and provide rich learning signals.

\subsection{Question Paraphrasing}
\label{appx:prompt_paraphrasing}
To diversify sentence structures and lexical expressions while preserving the original semantic meaning of the questions, we utilized the \texttt{Gemini-2.5-Pro} model. Figure~\ref{fig:prompt_q_parafrasing} presents the specific system prompt employed for this paraphrasing task.

\begin{figure*}[b]
\begin{promptbox}[System Prompt for Question Paraphrasing (Gemini-2.5-Pro)]
\textbf{Task:} Paraphrase the following multiple-choice question about meteorology in Korean.

\textbf{Rules:}

- Preserve the core meaning and all technical terminology.

- Change the sentence structure or phrasing for a more natural flow.

\textbf{Output Format:} Provide only the final paraphrased text. Do not include any introductory phrases or explanations.

---

\textbf{Original Question:} ``\{original\_question\}"

\end{promptbox}
\caption{\textbf{System prompt used to paraphrase Korean meteorological questions}}
\label{fig:prompt_q_parafrasing}
\end{figure*}

\subsection{Identification of Korean-Specific Subset}
To identify questions containing Korean-specific geographical and cultural contexts (the \textit{Korean-Specific} subset) from the total pool of 1,774 questions, we established a hybrid pipeline combining LLM-based filtering with human verification.

\paragraph{LLM-Aided Identification}
The screening process involved independent filtering using two distinct models: \texttt{Gemini-2.5-Pro} and \texttt{GPT-4.1}. The identification prompts for each model were optimized through an iterative refinement process to maximize recall. Figures~\ref{fig:prompt_k_identification_gemini} and \ref{fig:prompt_k_identification_gpt} illustrate the final enhanced prompts used for identifying Korean-specific context questions, respectively.

\paragraph{Human Selection Process}
Based on the LLM filtering, \texttt{Gemini-2.5-Pro} extracted 135 candidates, while \texttt{GPT-4.1} extracted 95 candidates. We consolidated these results into a union of 149 unique questions. Subsequently, two human researchers performed cross-validation on this candidate set to finalize the \textit{Korean-Specific} subset. The selected questions typically contain high-context keywords such as ``Our country'' (우리나라), ``Korean Peninsula,'' ``Jeju,'' ``Seoul,'' ``Yeongdong,'' ``Southerly wind'' (마파람), ``Taebaek Mountains,'' and ``24 Solar Terms.''

\subsection{Implicit vs. Explicit Dataset Design for Korean-Specific Subset}
\label{appx:k_oriented_design}
To ensure a fair evaluation of local context understanding regardless of the model's primary training language, we constructed a dual-version dataset by converting implicit questions into explicit ones. 

\begin{itemize}
    \item \textbf{Implicit Questions:} These refer to the original items containing high-context expressions that presuppose the speaker's spatiotemporal and cultural location (e.g., ``Our country,'' ``Maparam,'' ``East Coast'').
    \item \textbf{Explicit Questions:} These refer to the modified items where human researchers manually replaced high-context references with objective and unambiguous terminology (e.g., changing ``Our country'' to ``South Korea'' or ``Maparam'' to ``Southerly wind, a pure Korean term'').
\end{itemize}

Table~\ref{tab:k_implicit_explicit} presents comparative examples of these original implicit questions and their explicit counterparts.
\begin{table}[ht] 
    \centering
    \caption{\textbf{Examples of Context Transformation from implicit to explicit forms}}
    \renewcommand{\arraystretch}{1.3}
    \setlength{\tabcolsep}{3pt} 
    \small 

    \begin{tabularx}{\columnwidth}{@{} l >{\hsize=0.7\hsize}X >{\hsize=1.3\hsize}X @{}}
    \toprule
    \textbf{ID} & \textbf{Implicit (Before)} & \textbf{Explicit (After)} \\ 
    \midrule
    All & 우리나라 & 한국 지역 \\
    \hline
    618 & 서울 & 한국의 서울 지역 \\
    \hline
    1037 & 24절기 & 동아시아 지역의 24절기 \\
    \hline
    822 & 동해안 & 한국 지역의 동해안 \\
    \hline
    557 
     & 겨울철 발해만에서 작은 기압골이 접근하고 있다.
     & 겨울철 발해만으로부터 소규모 기압골이 한국 지역으로 접근하는 상황에서 \\ 
    \hline
    \makecell[tl]{271,\\ 1744} 
     & 마파람 
     & 한국 지역의 지방풍인 마파람 \\ 
    \bottomrule
    \end{tabularx}
    \label{tab:k_implicit_explicit}
\end{table}

\subsection{Evaluation Prompts for Korean-Specific Subset}
\label{appx:prompt_k_evaluation}
This section details the construction of system prompts designed to evaluate the model's understanding of geo-cultural contexts. To encourage the model to effectively utilize its latent local knowledge, we designed an \textit{Advanced Prompt} that explicitly defines the speaker's persona (i.e., a Korean meteorology expert) and clarifies that the questions are contextually situated in Korea.

To quantify the prompting gain—the extent to which this contextual cuing aids performance—and to ensure equitable evaluation for non-Korean models, we also established a \textit{Standard Prompt} as a control group. Figure~\ref{fig:prompt_standard} presents the standard system prompt used for the baseline experiment, while Figure~\ref{fig:prompt_advanced} displays the advanced system prompt used to test the activation of geo-cultural knowledge.

\subsection{Prompt for Reference Rationale Generation}
\label{appx:prompt_rationale_generation}
To secure high-quality reasoning references (rationales) for the benchmark, we utilized the \texttt{GPT-5} model. The prompt engineering process employed an iterative refinement technique. Specifically, we established a loop where an \textit{Enhancer} model drafted the initial prompt and a \textit{Critic} model identified weaknesses for revision, using \texttt{GPT-5} for both roles to derive the optimal instruction. The final system prompt used for rationale generation is presented in Figure~\ref{fig:prompt_rationale_generation}.

To ensure comprehensive coverage, the target questions were selected via stratified sampling to include all subject areas, modalities (text-only/multimodal), and Korean-specific items. Furthermore, to guarantee the validity of the reasoning paths, we enforced a strict filtering protocol: if the model generated an incorrect answer, the generation process was repeated until a rationale leading to the correct answer was produced.

\subsection{Questionnaire for Expert Verification on Reference Rationale}
\label{appx:questionnaire_rationale_verification}
To ensure the reliability of the LLM-as-a-judge pipeline, two human experts conducted a rigorous verification of the generated rationales from September 9 to October 19, 2025. This process was critical for establishing the integrity of the reference data. 
Before the verification process, the experts were given written instructions describing the purpose of the study, the expected completion time, and how their judgments would be used in the research. They were asked to assess each generated rationale in terms of factual accuracy, logical soundness, completeness, and conciseness, and to mark whether the rationale should be adopted as is or revised. When revisions were needed, they were instructed to provide either minor-fix or major-fix notes. An example of the questionnaire used for this process is shown in Table 11.

Out of 142 rationales initially generated by \texttt{GPT-5}, experts provided feedback for revision on 19 cases (13.38\%). The revisions primarily addressed technical accuracy and clarity. Specifically, experts corrected erroneous terminology (5 cases), such as changing `비열용량' (specific heat capacity) to `비열' (specific heat) or `지표소용돌이도' to `행성소용돌이도' (planetary vorticity). They also reinforced variable explanations and standard units (3 cases); for instance, refining the phrasing ``among the temperatures handled" to ``among the variables handled in atmospheric science" because `혼합비' (mixing ratio) is not a temperature variable. Additionally, the revisions included full sentence rewriting (7 cases), supplementary explanations (2 cases), and minor stylistic polishing (2 cases) to align with standard Korean meteorological conventions (e.g., standardizing `포텐셜온도' to `온위', `바트로픽' to `순압', and `단열가열' to `단열압축').

In addition to refining the AI-generated rationales, this expert review also identified inherent defects in the raw exam data. One question (ID 276) was discarded from the dataset as it was deemed logically unsolvable. Furthermore, questions with syntactic errors (IDs 308, 650) or issues with option configuration/double answers (IDs 14, 583, 1665) were precisely corrected based on expert consultation. Through this process, we secured the integrity of the final 141 reasoning evaluation samples. Table~\ref{tab:survey_expert_rationale} presents the specific questionnaire used for this expert verification process.

\subsection{Reasoning Prompt for Open-Source LLMs}
\label{appx:prompt_main_inference}
Figure~\ref{fig:prompt_reasoning_generation_oss} presents the system prompt utilized for generating reasoning paths and answers from open-source LLMs. It is important to note that this prompt serves as the standard instruction for the main inference phase of our benchmark evaluation protocol, rather than an experimental variation.

\subsection{Prompt for LLM-as-a-Judge Evaluation}
\label{appx:prompt_llm_as_a_judge}
Figure~\ref{fig:prompt_llm_judge_score} illustrates the specific system prompt employed for the LLM-as-a-Judge evaluation pipeline. The prompt was meticulously designed with the following key considerations to ensure robust alignment with human expert evaluation: 

\begin{itemize} 
    \item \textbf{Unified Evaluation Scale:} We adopted the identical 1-to-5 Likert scale and four evaluation axes—\textit{Factual Accuracy, Logical Soundness, Depth of Reasoning,} and \textit{Clarity \& Conciseness}—used by human experts. This unification allows for direct statistical comparison and correlation analysis between LLM and expert scores. 
    \item \textbf{Enforced Chain of Thought (CoT):} To enhance consistency, the prompt explicitly mandates a step-by-step thinking process. The evaluator is required to verify facts against the provided expert reference material \textit{before} assigning scores, thereby minimizing hallucinations and ensuring evidence-based grading. 
    \item \textbf{Explicit Scoring Criteria:} To prevent arbitrary scoring, we defined concrete rubrics for specific score tiers (e.g., distinguishing between a 5-point perfect answer and a 3-point answer with minor errors). 
    \item \textbf{Structured Output:} The prompt enforces a strict JSON output format that separates the ``List of Factual Errors'' from the quantitative scores. This structural constraint compels the model to explicitly isolate factual hallucinations from qualitative reasoning flaws. 
\end{itemize}

\subsection{Questionnaire for Expert Scoring of LLM Reasoning}
\label{appx:survey_expert_scoring}
Table~\ref{tab:survey_expert_scoring} outlines the questionnaire and scoring rubric provided to two meteorology professors. Crucially, this rubric served as the blueprint for the LLM-as-a-Judge prompt described above, ensuring that both human and AI evaluators operated under identical standards regarding accuracy and reasoning quality. 
The human experts were also provided with written scoring instructions that described the study purpose, the expected annotation time, and the four evaluation axes: factual accuracy, logical soundness, depth of reasoning, and clarity. To reduce bias, model identities were blinded in the scoring materials, and the experts were instructed to judge only the content of the generated reasoning against the expert-verified reference rationale. An example of the scoring questionnaire is provided in Table 12.

\section{Experimental Setups}

\subsection{Reasoning Model Inference}

We activate the thinking mode for hybrid models by setting \texttt{enable\_thinking = True} (\texttt{Qwen3-*B}, \texttt{EXAONE-4.0-*}), \texttt{reasoning\_effort = 'high'} (\texttt{gpt-5.2}), and \texttt{thinkingLevel = 'high'} (\texttt{gemini-3-pro-preview}). In contrast, \texttt{InternVL3.5-*-Instruct} is evaluated in standard instruct mode.

\subsection{Meta-Evaluation for LLM-as-a-Judge}
\label{appx:exp_setup_meta_eval}
To validate the reliability of the LLM evaluator, we designed a meta-evaluation protocol consisting of three steps: 1) generating reasoning paths and answers using various open-source LLMs; 2) performing LLM-as-a-Judge evaluation using expert-verified references and a specific rubric (based on a 5-point Likert scale across four evaluation axes); and 3) obtaining scores from two human experts using the identical rubric to calculate the statistical correlation between the LLM judge and human experts. 
The detailed prompt for the main inference, the judge prompt, and the expert scoring questionnaire are provided in Appendix~\ref{appx:prompt_main_inference}, \ref{appx:prompt_llm_as_a_judge}, and \ref{appx:survey_expert_scoring}, respectively. 
The full questionnaires and written instructions provided to the human experts are included in Appendix~\ref{appx:questionnaire_rationale_verification} and Appendix~\ref{appx:survey_expert_scoring} (Figure~\ref{fig:survey_expert_rationale} and Figure~\ref{fig:survey_expert_scoring}).

\paragraph{Sampling Strategy of Target Models}
To ensure that the LLM-as-a-Judge can reliably evaluate reasoning capabilities across a broad spectrum of proficiency, we employed a performance-based stratified sampling strategy. We categorized the pool of candidate models into three distinct tiers—Top, Mid, and Low—based on their normalized reasoning scores on the 141 reasoning questions. From these strata, we selected representative models to form a final set of 10 target models for the meta-evaluation, ensuring that the judge is tested against both high-quality coherent reasoning and lower-quality outputs. The list of sampled models is detailed in Table~\ref{tab:sampled_models}.

\begin{table}[ht]
    \centering
    \caption{\textbf{List of Sampled Models for Meta-Evaluation of LLM-as-a-Judge.} Models were selected via stratified sampling based on their normalized reasoning score tiers to ensure diverse evaluation targets. Reas. denotes the normalized reasoning score on the \textit{Reasoning} subset questions.}
    \renewcommand{\arraystretch}{1.2}
    \setlength{\tabcolsep}{2pt} 
    \resizebox{\columnwidth}{!}{%
    \begin{tabular}{@{}lllrr@{}}
        \toprule
        \textbf{Tier} & \textbf{Model Name} & \textbf{Family} & \textbf{Size (B)} & \textbf{Reas.} \\
        \midrule
        \multirow{3}{*}{\textbf{Top}}
        & Qwen3-VL-235B-A22B-Thinking & Qwen & 235.0 & 4.31 \\
        & gpt-oss-120b & OpenAI & 120.0 & 4.03 \\
        & A.X-4.0 & SKT & 72.0 & 3.87 \\
        \hline
        \multirow{4}{*}{\textbf{Mid}}
        & command-a-reasoning-08-2025 & Cohere & 111.0 & 3.53 \\
        & gpt-oss-20b & OpenAI & 20.0 & 3.39 \\
        & VARCO-VISION-2.0-14B & NCSoft & 14.0 & 2.81 \\
        & InternVL3.5-14B-Instruct & OpenGVLab & 15.0 & 2.36 \\ 
        \hline
        \multirow{3}{*}{\textbf{Low}}
        & Llama-3.1-8B-Instruct & Meta & 8.0 & 1.91 \\
        & InternVL3.5-8B-Instruct & OpenGVLab & 8.0 & 1.77 \\
        & Qwen3-0.6B & Qwen & 0.6 & 1.15 \\
        \bottomrule
    \end{tabular}%
    }
\label{tab:sampled_models}
\end{table}

\paragraph{Stratified Sampling of Evaluation Items}
To establish a robust gold standard for scoring, we selected 10 representative reasoning questions. Instead of random selection, we applied a stratified sampling to ensure both comprehensiveness and discriminatory power. The selection process involved the following criteria:

\begin{itemize}
    \item \textbf{Item Difficulty:} We classified the 141 reasoning questions into three difficulty tiers based on the average normalized reasoning scores of 10 open-source LLMs: Hard (Top 30\%), Mid (40\%), and Easy (Bottom 30\%). We sampled 3, 4, and 3 questions from each respective group to balance the difficulty distribution.
    \item \textbf{Discriminatory Power:} Within each difficulty tier, we prioritized questions with a high standard deviation in accuracy across the 10 models. A high standard deviation indicates that the question effectively discriminates between high- and low-performing models.
    \item \textbf{Category Coverage:} The selection was further constrained to ensure a balanced inclusion of text-only, multimodal, and Korean-specific questions, as well as coverage across the official exam subject areas (Parts 1, 3, 4, and 5).
\end{itemize}

Based on these criteria, the final 10 questions selected for meta-evaluation are: IDs 105, 1618, 1590, 14, 456, 1694, 963, 1745, 131, and 1224. Table~\ref{tab:sampled_items} details the characteristics of these sampled items.

\begin{table}[ht] 
    \centering
    \caption{\textbf{Statistics of Selected Evaluation Items.} Mean and Std. Dev. represent the item-wise normalized reasoning scores (1-5) across the selected models.}
    \renewcommand{\arraystretch}{1.2}
    \small 
    \resizebox{\columnwidth}{!}{%
    \begin{tabular}{@{}l c c c c c@{}}
    \toprule
    \textbf{Tier} & \textbf{ID} & \textbf{Mean} & \textbf{Std. Dev.} & \textbf{Part} & \textbf{Note} \\
    \midrule
    \multirow{3}{*}{\textbf{Hard}} & 105 & 2.23 & 1.72 & 3 & - \\
     & 1618 & 2.35 & 1.65 & 5 & - \\
     & 1590 & 2.55 & 1.91 & 3 & - \\
    \hline
    \multirow{4}{*}{\textbf{Mid}} & 14 & 2.70 & 1.89 & 1 & - \\
     & 456 & 2.73 & 1.88 & 3 & - \\
     & 1694 & 3.10 & 1.93 & 5 & - \\
     & 963 & 3.13 & 1.81 & 4 & Korean \\ 
    \hline
    \multirow{3}{*}{\textbf{Easy}} & 1745 &  3.35 & 1.89 & 4 & - \\
     & 131 & 3.40 & 1.74 & 5 & - \\
     & 1224 & 3.53 & 1.76 & 4 & - \\
    \bottomrule
    \end{tabular}%
    }
\label{tab:sampled_items}
\end{table}

\section{Additional Results and Discussion}

\subsection{Detailed Orthogonality Analysis of K-MetBench}
\label{appx:orthogonality}

\begin{figure*}[!hb]
    \centering
    \includegraphics[width=\linewidth]{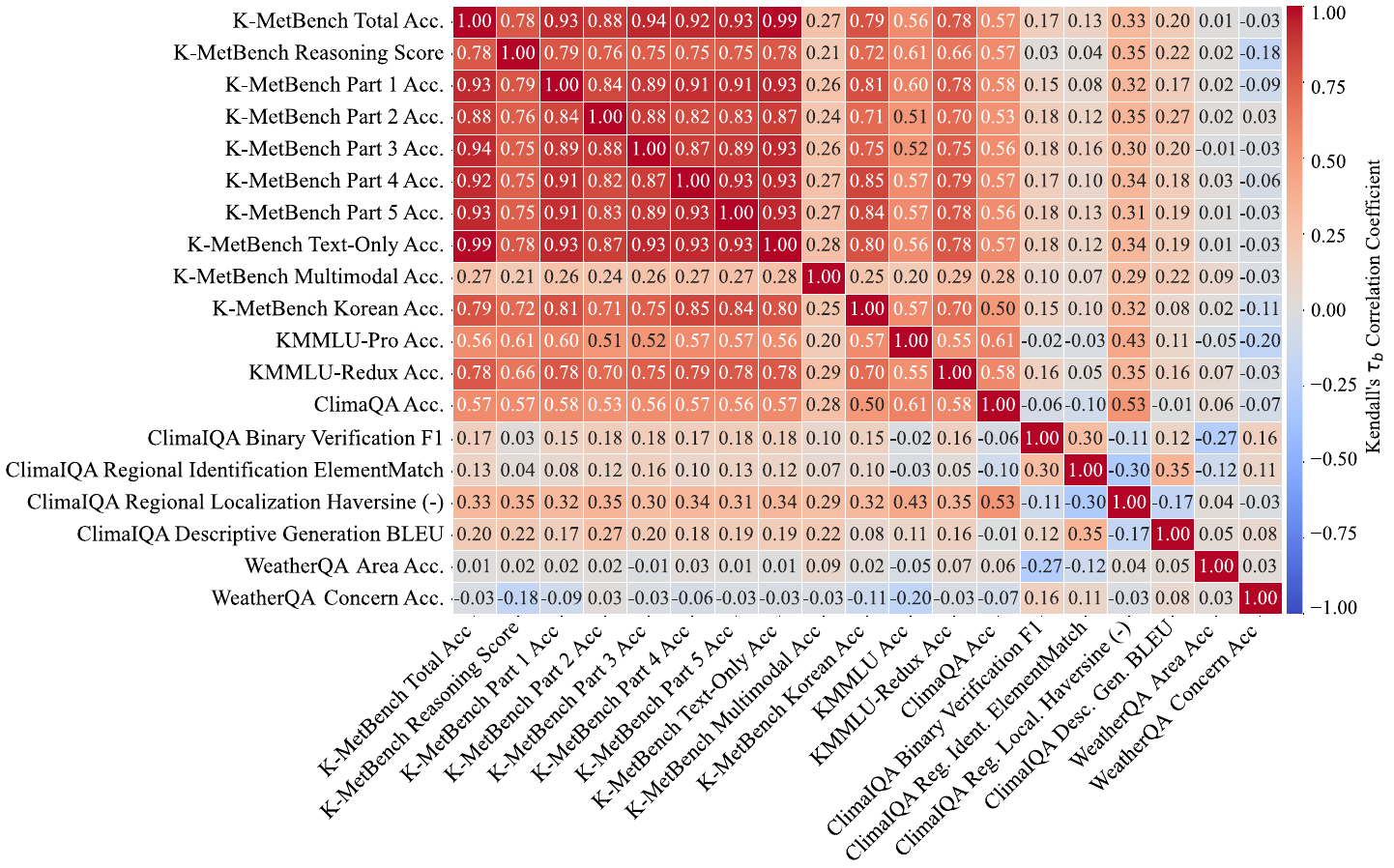}
    \caption{\textbf{Heatmap of Kendall's $\tau_b$ rank correlations between K-MetBench and existing benchmarks.} Red denotes high positive correlation, while blue indicates negative correlation.}
    \label{fig:orthogonal_overall}
\end{figure*}

\paragraph{The Uniqueness of Visual Reasoning.}
As illustrated in Figure~\ref{fig:orthogonal_overall}, the \textit{Multimodal} subset of K-MetBench displays consistently low correlations (avg. $\tau_b < 0.30$) 1across all external benchmarks, including text-based baselines (KMMLU-Pro, KMMLU-Redux, ClimaQA) and weather-domain vision benchmarks (ClimaIQA, WeatherQA). This disconnect quantitatively confirms the modality gap, demonstrating that the ability to interpret meteorological charts and symbols is a distinct skill set not linearly correlated with general linguistic or textual reasoning capabilities.

To investigate the orthogonality of our benchmark, we further analyzed correlations with KMMLU-Pro and -Redux. For KMMLU-Redux, where only the test set is publicly available, we specifically partitioned the data into the 39 questions derived from the 2022 Meteorological Engineer exam versus the remaining 2,547 general questions. The sample pool for this analysis consisted of 25 open-source VLLMs for \textit{Multimodal} subset comparisons and 52 open-source models for other subsets (excluding the proprietary models listed in Table~\ref{tab:kmetbench_scores}).

\begin{figure}[!ht]
    \centering
    \includegraphics[width=\linewidth]{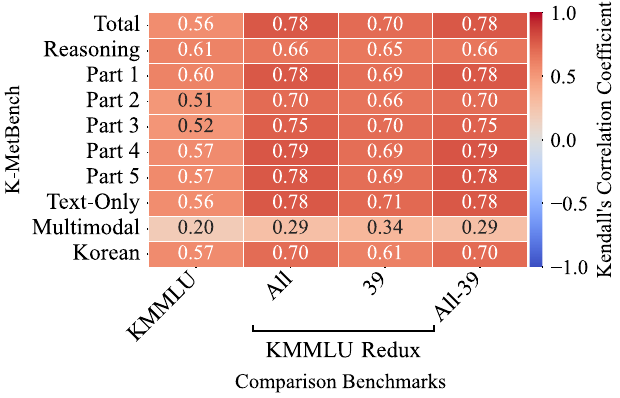}
    \caption{\textbf{Heatmap of Kendall's $\tau_b$ rank correlations between K-MetBench and KMMLU, KMMLU-Redux).} In KMMLU-Redux, `39' denotes the subset of 39 Meteorological Engineer Exam questions, while `All$-$39' refers to the remaining subset excluding these meteorological questions. Acc.: Accuracy.}
    \label{fig:orthogonal_kmmlu}
\end{figure}

As shown in Figure~\ref{fig:orthogonal_kmmlu}, KMMLU-Pro exhibited weaker correlation due to domain divergence. Within KMMLU-Redux, the isolated 39-question meteorological subset showed lower correlation ($\tau_b=0.70$) than the full dataset ($\tau_b=0.78$), suggesting that this small subset is insufficient to capture comprehensive meteorological capability. Crucially, a significant drop in correlation is observed for K-MetBench's multimodal and reasoning subsets, highlighting the structural gap between our multimodal evaluation and existing text-only licensing exams.

\subsection{Detailed Analysis of K-MetBench Performance}
\label{appx:kmetbench_performance}

Table~\ref{tab:kmetbench_scores} presents the comprehensive leaderboard of K-MetBench, evaluating a diverse range of proprietary and open-source models. The results are categorized by model type, capabilities (Korean-native, Multimodal, Reasoning), and granular domain performance.

\paragraph{SOTA Performance and the Impact of Reasoning.}
Proprietary models dominate the upper echelon of the leaderboard. \texttt{gemini-3-pro-preview} (Thinking) achieves state-of-the-art performance with a total accuracy of 93.7\%, significantly outperforming other contenders. A notable trend is the efficacy of \textit{Thinking} (reasoning) models; for instance, \texttt{gpt-5.2} (Thinking) scores 87.8\%, showing a substantial improvement (+10.2\%p) over its standard counterpart, \texttt{gpt-5.2} (77.6\%). This pattern reinforces that chain-of-thought capabilities are crucial for solving complex meteorological problems.

\paragraph{Open-Source Landscape.}
In the open-source domain, the \texttt{Qwen}\cite{yang2025qwen3,bai2025qwen2,qwen2024qwen2} series exhibits exceptional performance. \texttt{Qwen3-VL-235B-A22B-Thinking} leads this category with 84.4\%. Even smaller models like \texttt{Qwen3-VL-32B-Thinking} (78.6\%) surpass much larger non-reasoning models (e.g., \texttt{gpt-oss-120b}, 77.3\%), highlighting the efficiency of reasoning-enhanced architectures in specialized scientific domains.

\paragraph{The Modality Gap.}
A critical disparity exists between textual and visual reasoning. While top models achieve near-perfect scores on the \textit{Text} subset (e.g., Gemini: 94.6\%), their performance drops significantly on the \textit{Multimodal} subset (Gemini: 75.6\%). This modality gap is even more pronounced in other models; \texttt{gpt-5.2 (Thinking)} sees a drastic decline from 90.6\% (Text) to 29.3\% (Multi). This indicates that while current LLMs excel at theoretical knowledge retrieval, they still struggle with interpreting professional meteorological charts and diagrams.

\paragraph{Geo-Cultural Alignment and Granularity.}
Korean-native models demonstrate distinct advantages in localized contexts. \texttt{A.X-4.0} achieves a high \textit{K-Specific} score of 78.9\%, outperforming several larger global models in this specific subset, despite a lower overall accuracy. In terms of domain granularity (P1--P5), models generally perform best in \textit{Meteorological Observation} (P2), likely due to the descriptive nature of the questions, while struggling more in \textit{Atmospheric Dynamics} (P3) and \textit{Atmospheric Physics} (P5), which require deeper calculation and physical conceptualization.

\subsection{Results of Meta Evaluation of LLM-as-a-Judge}
\label{appx:results_meta_eval}

\begin{figure}[ht]
    \centering
    \includegraphics[width=\linewidth]{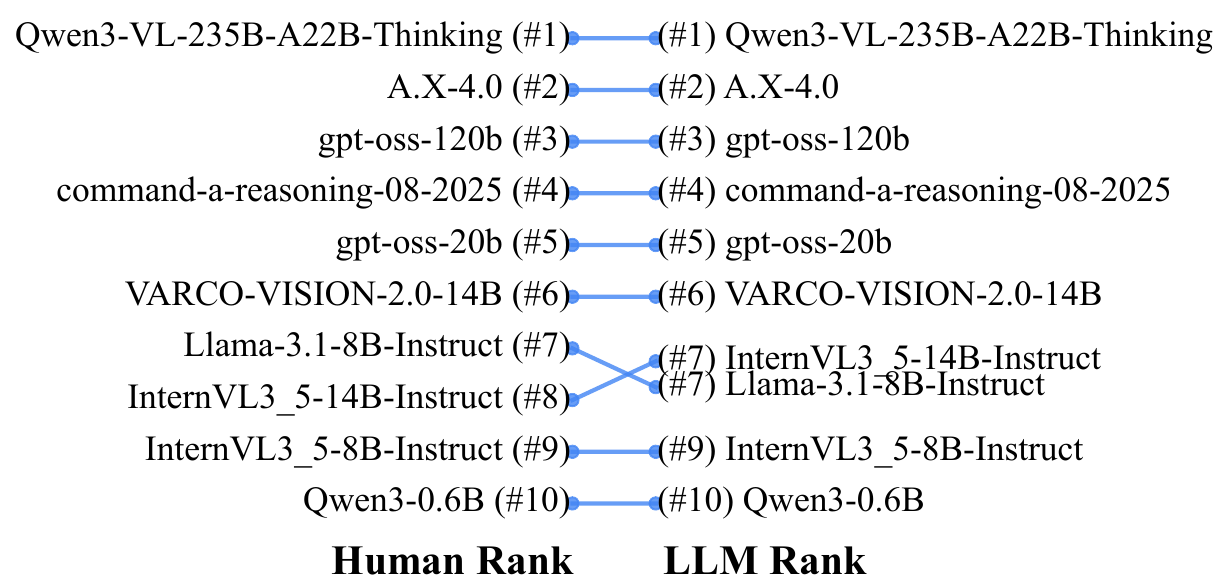}
    \caption{\textbf{Slope graph of rank changes of reasoning evaluation scores of two human experts vs. LLM evaluator.} The crossing lines indicate minor discrepancies, but the overall performance tiers remain largely consistent.}
    \label{fig:llm_judge_slope}
\end{figure}

\paragraph{Rank Preservation Analysis.}
In benchmark evaluation, the accuracy of relative ranking is often more critical than absolute scores. The slope graph in Figure~\ref{fig:llm_judge_slope} compares the rankings assigned by human experts and the LLM. Although minor rank fluctuations exist, the overall trend distinguishing high-performing models from low-performing ones is preserved.

\subsection{MCQA Accuracy vs. Reasoning Score}
\label{appx:accuracy_vs_reasoning}

\begin{figure}[ht]
    \centering
    \includegraphics[width=\linewidth]{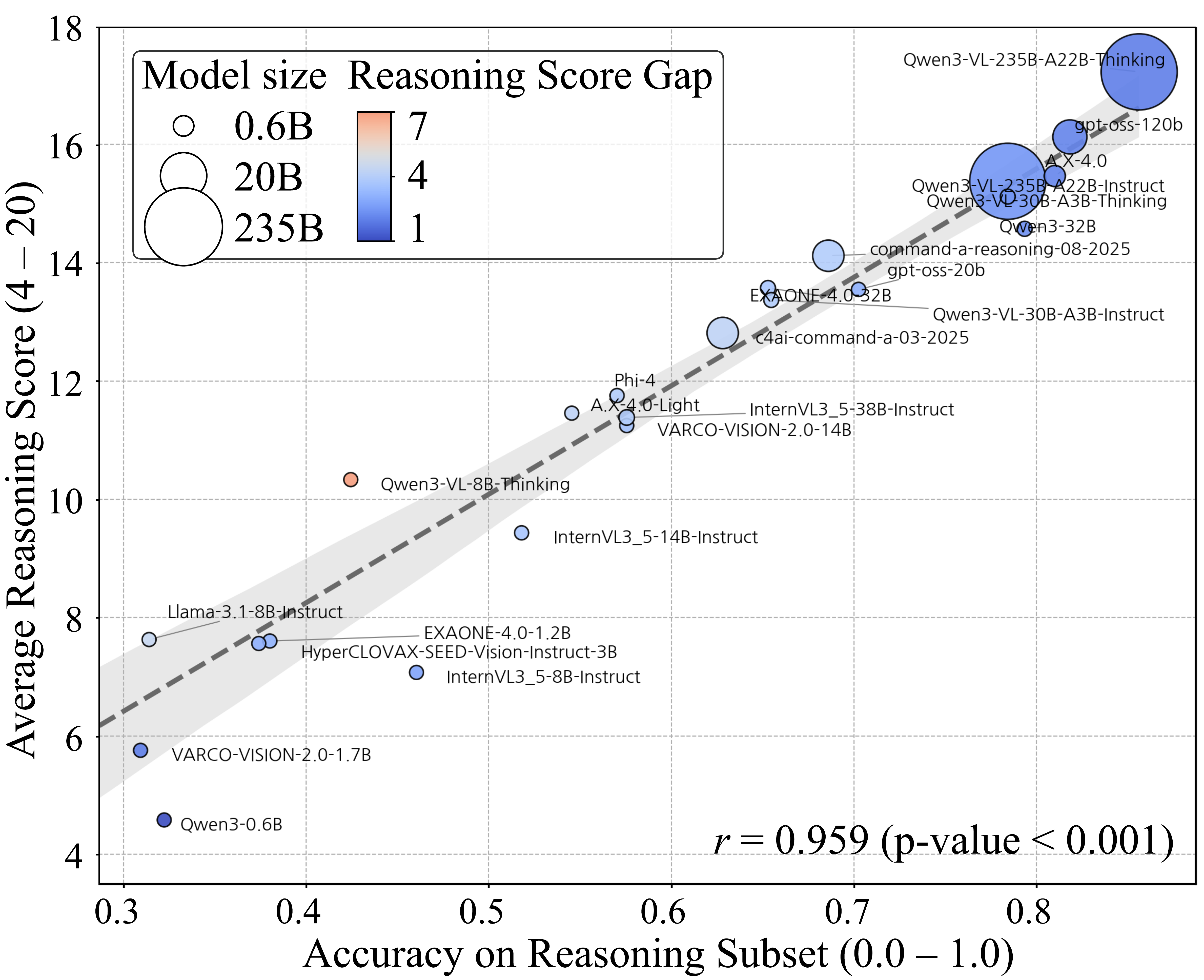}
    \caption{\textbf{Scatter plot of MCQA Accuracy vs. Reasoning Score.} The x-axis represents the answer accuracy, while the y-axis denotes the qualitative reasoning score evaluated by the judge. Marker sizes are proportional to the model parameter count. The strong correlation ($r=0.959$) confirms that high-performing models generally provide more reliable reasoning traces.}
    \label{fig:reasoning_scatter}   
\end{figure}

As illustrated in Figure~\ref{fig:reasoning_scatter}, we analyze the relationship between answer accuracy and qualitative reasoning capabilities. The color gradient represents the \textit{Reasoning Score Gap}, defined as the disparity between the reasoning score of correctly answered items and the overall average (i.e., $\text{Reasoning Score Gap} = \text{Reasoning Score}_{|A=\text{correct}} - \text{Reasoning Score}_{\text{total}}$).

We observe a strong positive correlation ($r=0.959$) between QA accuracy and reasoning scores, indicating that models that derive correct answers also tend to generate higher-quality reasoning traces. A distinct scaling law is also evident; larger models (shown by marker size) consistently populate the upper-right quadrant, achieving superior performance in both metrics. Two notable trends appear among specific models:

\paragraph{High Reasoning but Low Accuracy:} \texttt{Qwen3-VL-8B-Thinking} emerges as an outlier. Despite its relatively low accuracy, it maintains a high reasoning score. This suggests that while the model generates detailed "thinking" processes, its limited capacity (8B) often leads to hallucinations or logical fallacies in the final deduction.

\paragraph{Impact of Reasoning Optimization:} The benefit of reasoning-specific training is highlighted by the Command family. \texttt{command-a-reasoning-08-2025} significantly outperforms its predecessor, \texttt{c4ai-command-a-03-2025}, in both accuracy and reasoning quality, validating the efficacy of reasoning-enhanced fine-tuning.

\subsection{Computational Cost and Efficiency Analysis}
We evaluated the normalized total GPU compute time for 100 questions against the 150-minute exam limit ($\approx$ 2.50 GPU-hours). 

Standard instruction-tuned models (e.g., \texttt{Qwen2.5-VL-Instruct}) demonstrated negligible cost ($<0.01$ GPU-hours), operating orders of magnitude faster than the human time constraint. 

Reasoning models exhibited significant computational overhead. While \texttt{Qwen3-VL-8B-Thinking} (2.4 GPU-hours) remained within the limit, larger models like \texttt{Qwen3-VL-32B-Thinking} (3.8 GPU-hours) and \texttt{command-a-reasoning} (20.8 GPU-hours) exceeded the threshold, highlighting the substantial resource trade-off required for deep reasoning. While this heavy computational overhead may yield deeper reasoning traces, it poses challenges for time-sensitive forecasting applications where rapid decision-making is critical. However, employing tensor parallelism can effectively reduce wall-clock inference time.

\section{Compute Resources}
\label{appx:resource}
We evaluated all open-source models on an internal cluster equipped with 40 NVIDIA H100 80GB PCIe GPUs. To maximize inference efficiency, we utilized the vLLM library for all benchmark evaluations. The evaluation covered 52 text-only and multimodal models across all subsets of K-MetBench, totaling approximately 192.14 H100 GPU hours (153.01 and 39.13 GPU hours for standard MCQA and reasoning MCQA, respectively).

\section{Hierarchical Topic Distribution}
\begin{figure*}[ht]
    \centering
    \includegraphics[width=\linewidth]{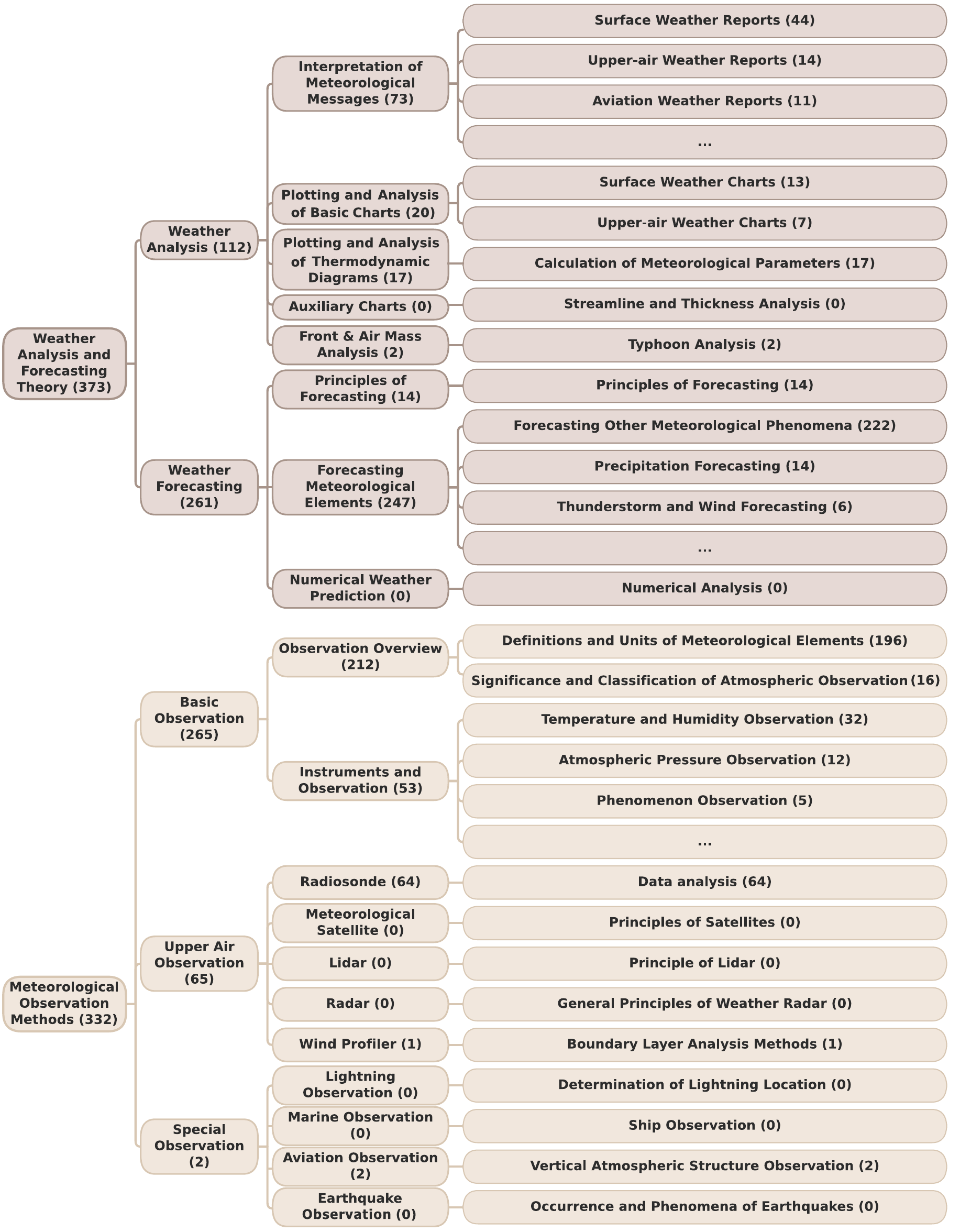}
    \caption{\textbf{Hierarchical taxonomy and sample distribution for Parts 1 and 2.} The diagram visualizes the breakdown of \textit{Meteorology \& Thermodynamics} and \textit{Observation Methods} into detailed sub-topics. Numbers in parentheses indicate the estimated quantity of questions classified by Gemini-2.5-Pro.}
    \label{fig:category_1}
\end{figure*}

\begin{figure*}[ht]
    \centering
    \includegraphics[width=\linewidth]{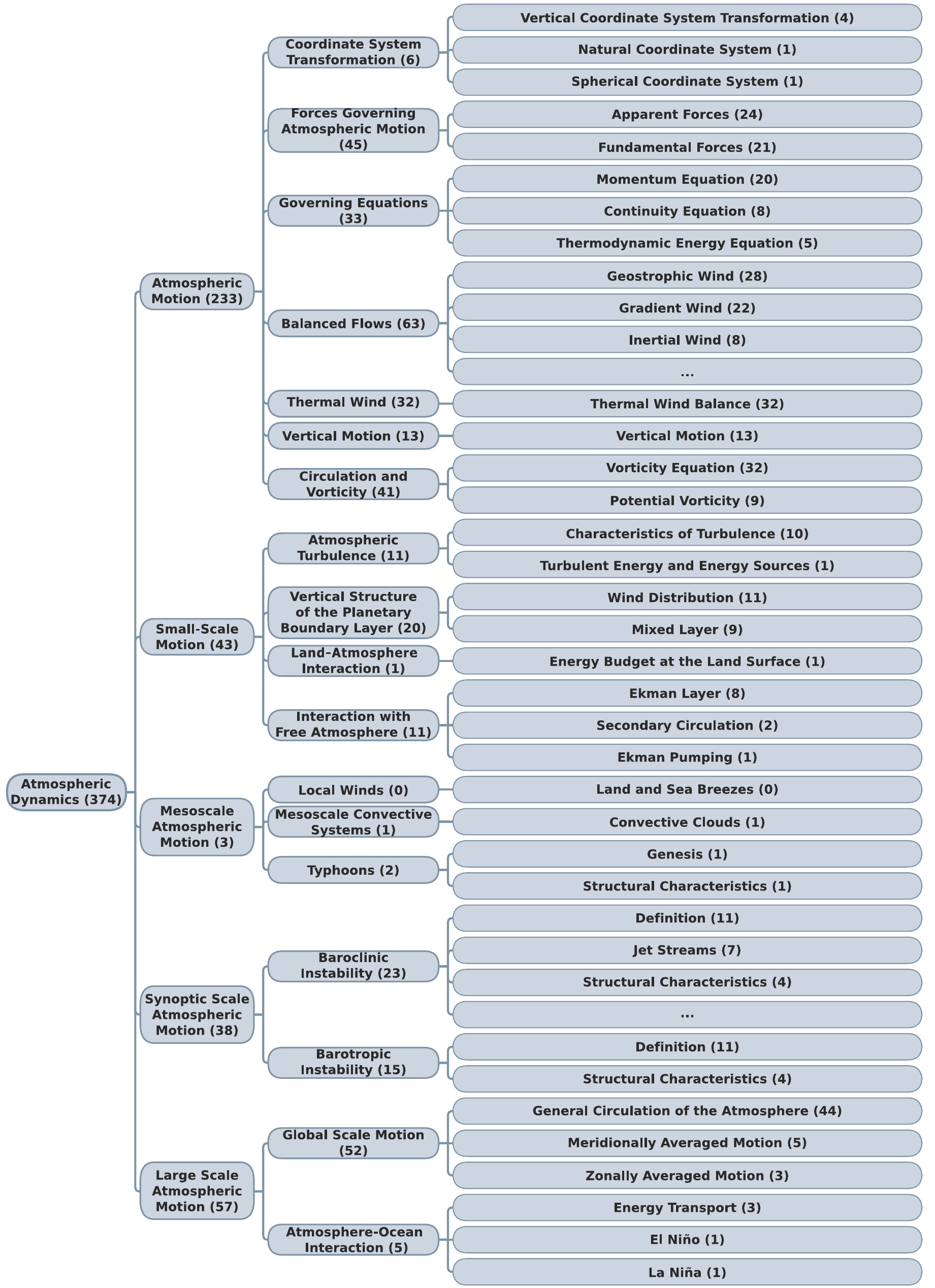}
    \caption{\textbf{Hierarchical taxonomy and sample distribution for Part 3.} This overview details the structure of \textit{Forecasting \& Climatology}, mapping the dataset samples to specific forecasting theories and climate phenomena.}
    \label{fig:category_2}
\end{figure*}

\begin{figure*}[ht]
    \centering
    \includegraphics[width=\linewidth]{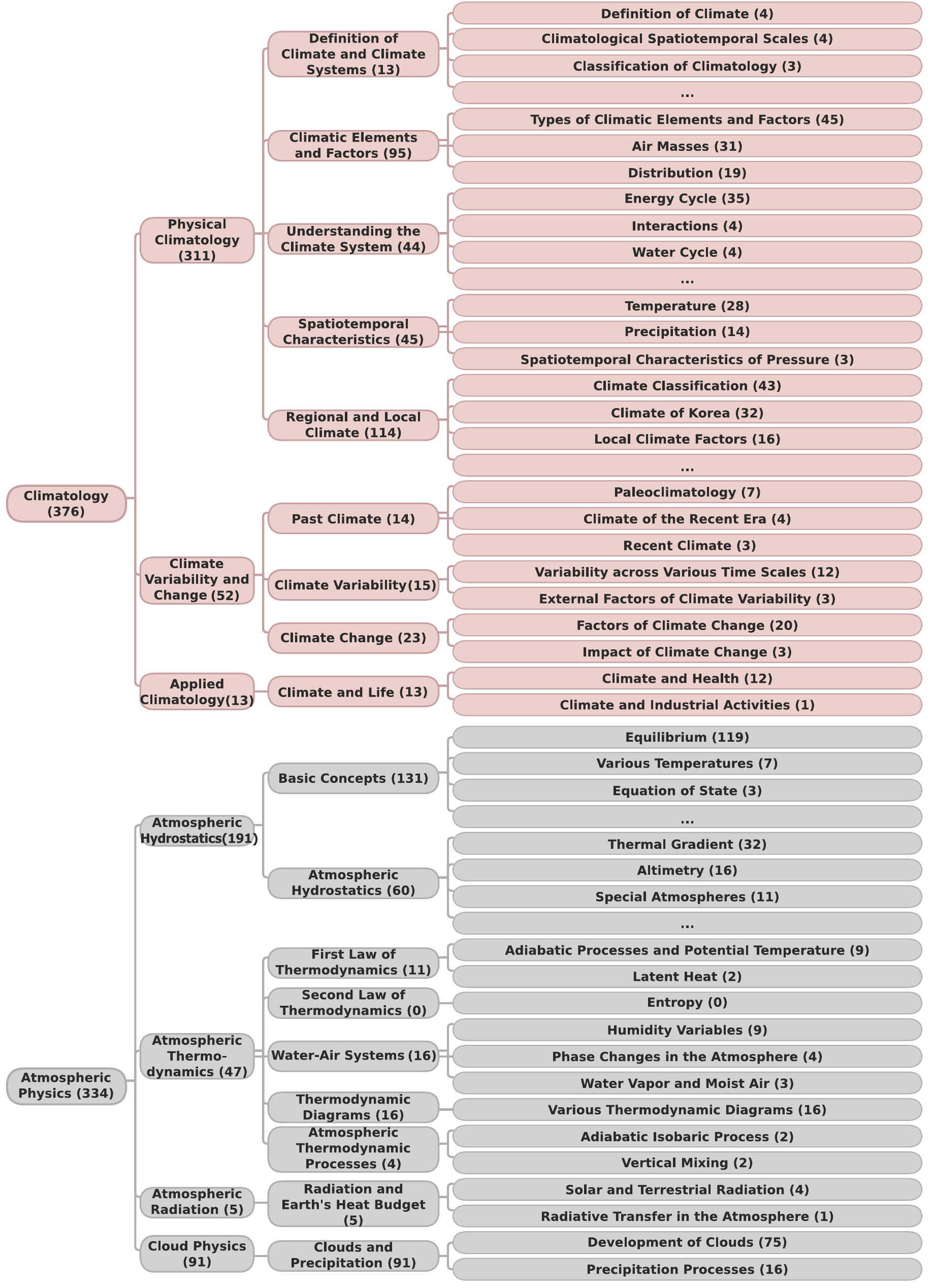}
    \caption{\textbf{Hierarchical taxonomy and sample distribution for Parts 4 and 5.} The diagram covers \textit{Applied Meteorology} and \textit{Weather Chart Analysis}, illustrating the coverage of practical applications and legal regulations.}
    \label{fig:category_3}
\end{figure*}

Figures~\ref{fig:category_1} through \ref{fig:category_3} illustrate the comprehensive hierarchical taxonomy of the K-MetBench 

dataset, aligned with the official evaluation criteria of the National Meteorological Engineer written examination~\cite{hrdk2024meteorological}. The dataset spans five major subject areas: \textit{Weather Analysis and Forecasting Theory}, \textit{Meteorological Observation Methods}, \textit{Atmospheric Dynamics}, \textit{Climatology}, and \textit{Atmospheric Physics}.

As depicted in Figure~\ref{fig:category_1}-\ref{fig:category_3}, each subject area and hierarchy demonstrates the benchmark's fine-grained granularity and comprehensive coverage of meteorological domain knowledge. The numerical values in parentheses represent the estimated count of questions belonging to each specific category. To map the 1,774 questions to this detailed hierarchy, we employed \texttt{Gemini-2.5-Pro} for automated classification. These counts serve as an indicative reference, highlighting the dataset's balanced coverage across the theoretical and practical spectrums of meteorology.

\clearpage

\begin{figure*}[ht]
    \centering
    \captionof{table}{\textbf{Expert verification questionnaire for reference rationales}}
    \renewcommand{\arraystretch}{1.2}
    
    \begin{tabularx}{\textwidth}{@{}l X X l X l l l@{}}
        \toprule
        \textbf{ID} & \textbf{Question} & \textbf{Choices} & \textbf{\makecell[tl]{Exact\\Answer}} & \textbf{Generated Rationale} & 
        \textbf{Adopt?} & 
        \textbf{\makecell[tl]{Note 1 \\ (Minor Fix)}} & 
        \textbf{\makecell[tl]{Note 2 \\ (Major Fix)}} \\
        \midrule
        18 & \narrow{\ko{비열의 ...}} & 1. ... & 2 & \narrow{\ko{비열은 ...}} & \cellcolor{lightgray}yes & \cellcolor{lightgray}- & \cellcolor{lightgray}- \\
        \bottomrule
        \multicolumn{8}{@{}p{\textwidth}@{}}{
            \rule{0pt}{3ex}\footnotesize 
            \textit{\textbf{Note:} The \colorbox{lightgray}{gray-shaded cells} indicate the items to be answered by the expert. Adoption Criteria (Accuracy, Logical Soundness, Completeness, and Conciseness) are provided separately.}
        } \\
    \end{tabularx}
    \label{tab:survey_expert_rationale}

    \vspace{1cm} 
    
    \setlength{\fboxsep}{0pt} 
    \setlength{\fboxrule}{0.5pt}
    
    \begin{subfigure}{\linewidth}
        \centering
        \fbox{\includegraphics[width=0.98\linewidth]{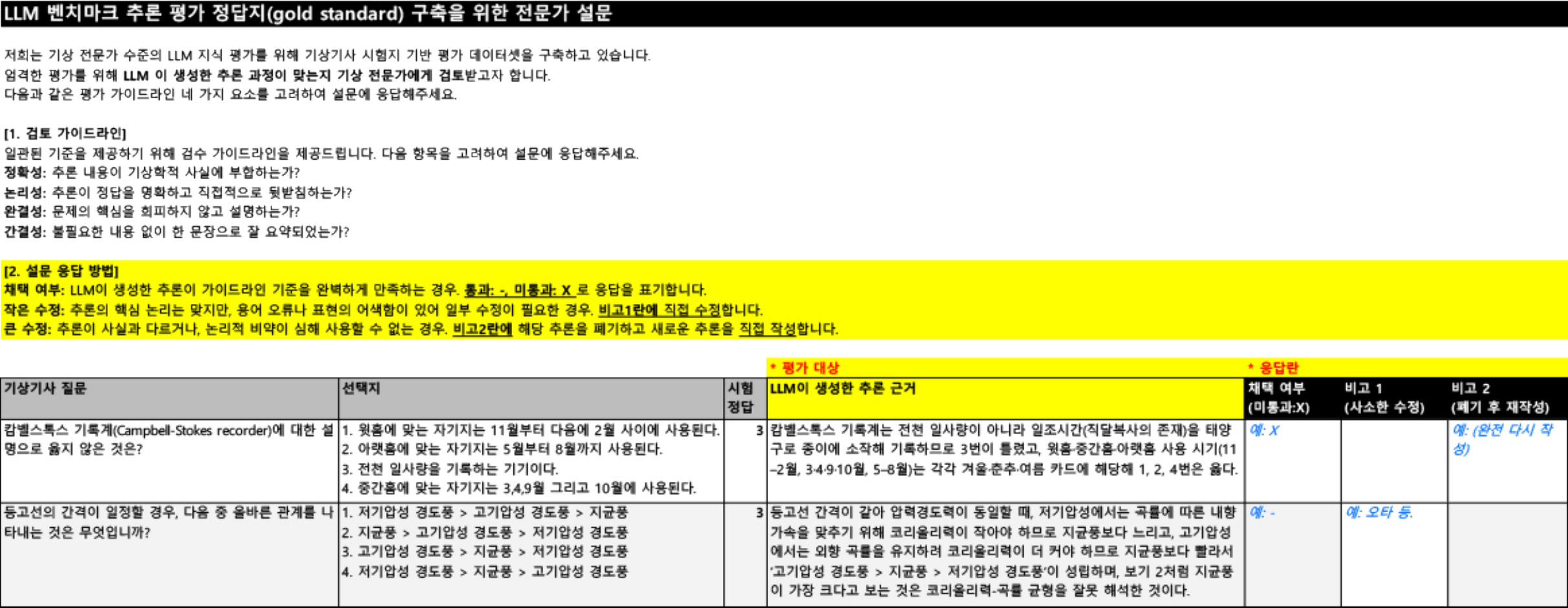}}
        \caption{Text-only Reasoning Questions}
        \label{fig:survey_text}
        \vspace{0.1in}
    \end{subfigure}
    \begin{subfigure}{\linewidth}
        \centering
        \fbox{\includegraphics[width=0.98\linewidth]{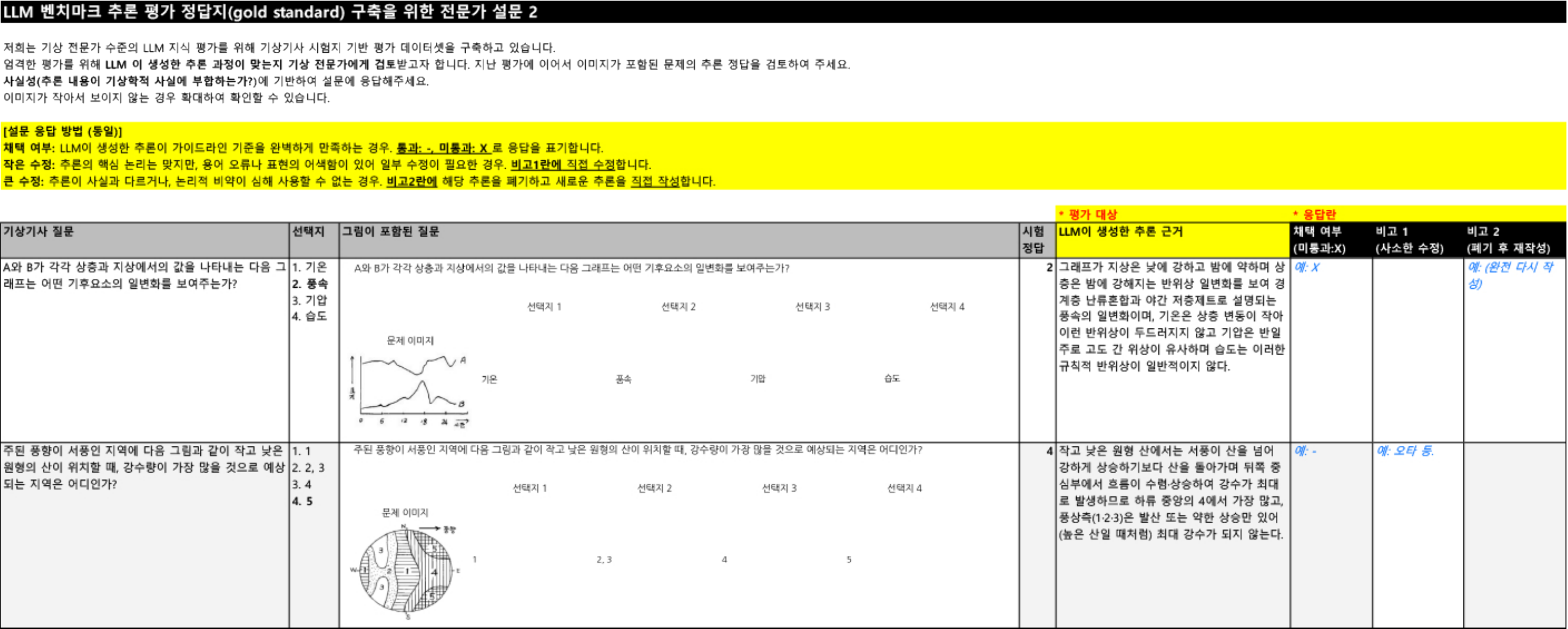}}
        \caption{Multimodal Reasoning Questions}
        \label{fig:survey_multimodal}
    \end{subfigure}
    \captionof{figure}{\textbf{Examples of the expert verification questionnaire for reference rationales} (in Korean)}
    \label{fig:survey_expert_rationale}
\end{figure*}

\begin{figure*}[ht]
    \centering
    \captionof{table}{\textbf{Questionnaire for expert scoring of open-source LLM reasoning results}}
    \label{tab:survey_expert_scoring}
    \renewcommand{\arraystretch}{1.3} 
    \setlength{\tabcolsep}{3pt}       
    \begin{tabularx}{\textwidth}{@{} l l l l l X l l l l l l @{}}
    \toprule
    \textbf{ID} & 
    \textbf{Question} & 
    \textbf{Choices} & 
    \textbf{\makecell[tl]{Exact\\Answer}} & 
    \textbf{GT-R} & 
    \textbf{Target R} & 
    \textbf{\makecell[tl]{Fact\\(1-5)}} & 
    \textbf{\makecell[tl]{Sound\\(1-5)}} & 
    \textbf{\makecell[tl]{Depth\\(1-5)}} & 
    \textbf{\makecell[tl]{Clear\\(1-5)}} & 
    \textbf{\makecell[tl]{Total\\(4-20)}} &
    \textbf{Note} \\
    \midrule
    18 & \narrow{\ko{비열의 ...}} & 1. ... & 2 & 
    \narrow{\ko{비열은 ...}}  & 
    \narrow{\ko{비열용량은 ...}}  & 
    \cellcolor{lightgray}1 & 
    \cellcolor{lightgray}2 & 
    \cellcolor{lightgray}1 & 
    \cellcolor{lightgray}5 & 
    \cellcolor{lightgray}9 & 
    \cellcolor{lightgray}- \\
    \bottomrule
    \multicolumn{12}{@{}p{\textwidth}@{}}{
        \rule{0pt}{3ex}\footnotesize 
        \textit{\textbf{Note:} GT-R: Expert-verified Gold Rationale, Target R: LLM Generated Reasoning. Scoring: 1(Poor) -- 5(Excellent). The \colorbox{lightgray}{gray-shaded cells} indicate the items to be answered by the expert.
        }} \\
    \end{tabularx}

    \vspace{1cm}

    \setlength{\fboxsep}{0pt}  
    \setlength{\fboxrule}{0.5pt} 
    \fbox{\includegraphics[width=0.98\linewidth]{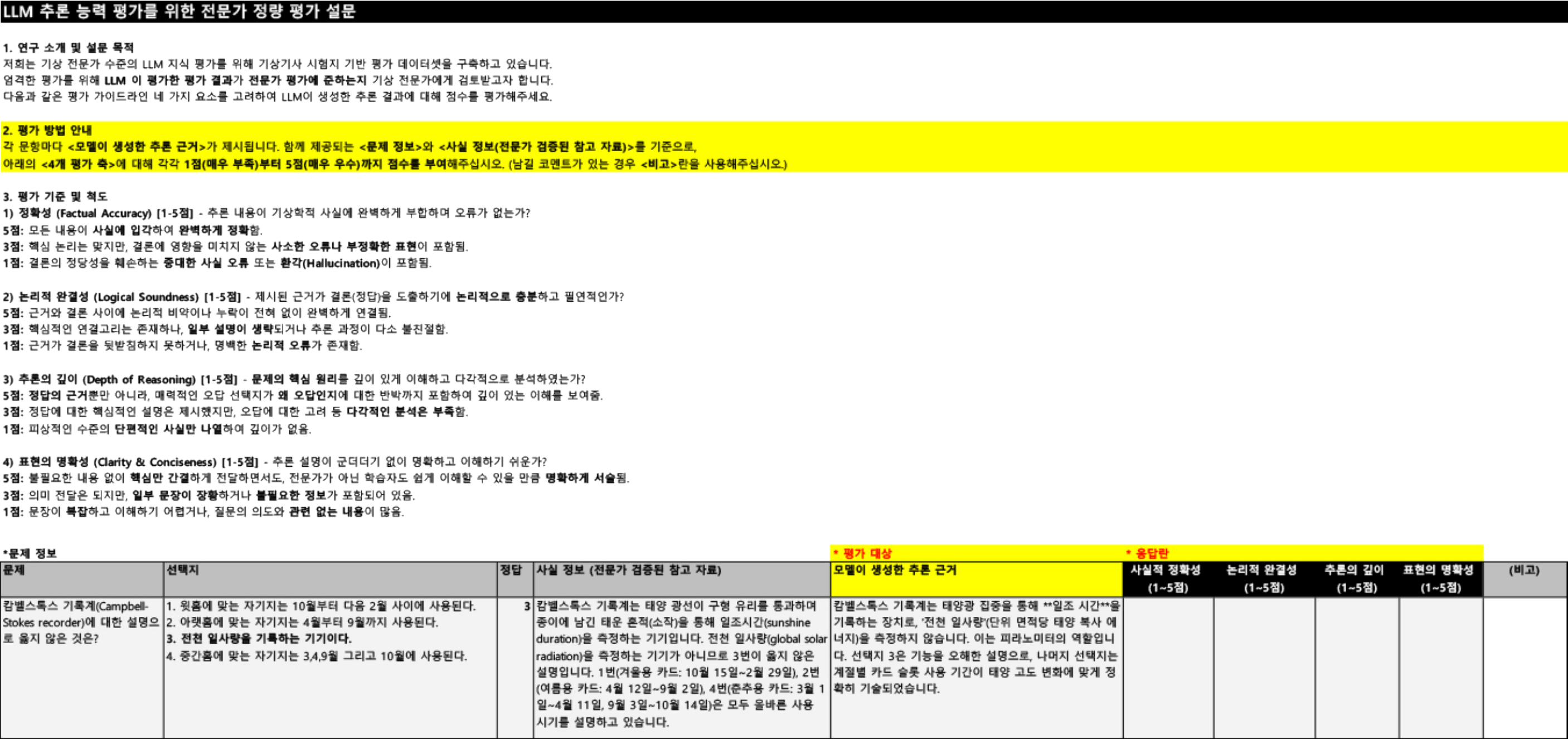}}
    \captionof{figure}{\textbf{Examples of the questionnaire for expert scoring of open-source LLM reasoning results} (in Korean)}
    \label{fig:survey_expert_scoring}
\end{figure*}

\begin{figure*}[ht]
    \begin{promptbox_kor}[System Prompt for Identifying Regional Questions (Gemini-2.5-Pro)]
    너는 이제부터 기상학 전문가야.\\
    주어진 `기상기사 문제'를 분석해 줘. 각 문제의 내용과 선택지를 신중하게 검토해서, 전 세계적으로 적용되는 일반적인 기상 지식이 아니라 오직 한국의 지리, 기후, 기상 시스템, 관련 기관에만 해당되는 `지역성이 강한' 문제를 식별해 줘.\\
    \\
    다음과 같은 기준을 사용해서 문제를 선별해:\\
    \quad - 한국의 특정 지역(예: 영동, 영서, 울릉도, 서해안)의 기상 현상을 다루는 문제\\
    \quad - 한국에서만 사용하는 특정 기상 용어나 예보 시스템에 관한 문제 (예: 동네예보)\\
    \quad - 한국 기상청 또는 관련 기관의 역할이나 업무와 직접적으로 관련된 문제\\
    \quad - 황사, 장마, 태풍 등 한국에 큰 영향을 미치는 기상 현상이라도, 그 내용이 한국의 특수한 상황(예: 특정 지역의 영향, 한국의 예보 체계)과 결부되어 있을 경우에만 선택\\
    \\
    결과는 다른 설명 없이, 오직 해당 문제의 ID 번호만 포함된 숫자(int) 형식으로 출력해 줘.
    \end{promptbox_kor}
    \captionof{figure}{\textbf{The system prompt used by Gemini-2.5-Pro to filter Korea-specific meteorological questions}}
    \label{fig:prompt_k_identification_gemini}

    \vspace{1cm}

    \begin{promptbox}[English Translation of the System Prompt for Identifying Regional Questions (Gemini-2.5-Pro)]
    You are an expert in meteorology.\\
    Analyze the provided `Meteorological Engineer Exam Questions'. Carefully review the content and options of each question to identify `strongly regional' questions that pertain solely to South Korea's geography, climate, weather systems, and relevant institutions, rather than general meteorological knowledge applicable globally.\\
    \\
    Use the following criteria to select the questions:\\
    \quad - Questions dealing with meteorological phenomena in specific regions of Korea (e.g., Yeongdong, Yeongseo, Ulleungdo, West Coast)\\
    \quad - Questions regarding specific meteorological terms or forecasting systems used exclusively in Korea (e.g., Neighborhood Forecast)\\
    \quad - Questions directly related to the roles or operations of the Korea Meteorological Administration (KMA) or relevant institutions\\
    \quad - Even for weather phenomena that significantly impact Korea, such as Asian Dust, Changma (rainy season), or Typhoons, select them only if the content is tied to Korea's specific context (e.g., impact on a specific region, Korea's forecasting system)\\
    \\
    Output the result strictly as integers representing the ID numbers of the relevant questions, without any further explanation.
    \end{promptbox}
    \captionof{figure}{\textbf{English translation of the system prompt used by Gemini-2.5-Pro to filter Korea-specific meteorological questions}}
    \label{fig:prompt_k_identification_gemini_eng}
\end{figure*}

\begin{figure*}[ht]
    \begin{promptbox_kor}[System Prompt for Identifying Regional Questions (ChatGPT-4.1)]
    \textbf{\# 역할} \\
    당신은 `대한민국 기상기사 자격시험'의 과목·출제 경향을 잘 아는 기상교육 전문가입니다.
    전 세계적으로 통용되는 일반 기상지식과, 한국 기상 현업·제도·지형·행정 등 지역 특화 지식의 차이를 명확히 구분할 수 있습니다.\\
    \\
    \textbf{\# 목표} \\
    아래에 제시된 기상기사 시험문제 중 `전 세계 어디서나 동일하게 적용되는 일반 기상지식'이 \textbf{아닌}, `한국에 특화된 지역성·제도성·행정적 배경이 강하게 반영된 문제\textbf{만}'을 골라 문제 ID(번호)만 반환하세요.\\
    \\
    \textbf{\# 판단 기준}\\
    \quad - \textbf{한국 전용 관측 장비·용어}(예: 장파 고주파 근접배열 관측망, ASOS-K)\\
    \quad - \textbf{한국 기상청(KMA) 행정 절차·법규·고시} (예: 특보 발효 기준, 관측 보고 시각 규정)\\
    \quad - \textbf{한국 지형·기후 특수성}(예: 서해안 눈 구름 특성, 태풍 상륙 경로 통계)\\
    \quad - \textbf{우리나라에서만 사용하거나 시험 범위에 포함되는 고유 어휘·단위·축약어}\\
    \quad - 위와 같은 요소가 전혀 없고, ICAO·WMO 등 국제 공통 규격만 다루면 `전 세계 일반'으로 간주\\
    \\
    \textbf{\# 출력 형식}\\
    \quad - 아무 설명 없이 \textbf{숫자}만 출력 \\
    \quad 예: `4' \\
    \quad - 리스트가 비어 있으면 `없음'이라고만 작성 \\
    \\
    \textbf{\# 사고 과정}\\
    \quad - 문제 하나씩 읽고, 위 `판단 기준'에 따라 한국 특화 여부를 먼저 \textbf{머릿속에서} 결정\\
    \quad - 최종 답안에는 \textbf{결과 숫자만} 남기고, 중간 추론이나 이유는 쓰지 말 것
    \end{promptbox_kor}
    \captionof{figure}{\textbf{System prompt used by ChatGPT-4.1 to filter Korea-specific questions}}
    \label{fig:prompt_k_identification_gpt}

    \vspace{1cm}

    \begin{promptbox}[English Translation of the System Prompt for Identifying Regional Questions in English (ChatGPT-4.1)]
    \textbf{\# Role} \\
    You are a meteorology education expert familiar with the subjects and trends of the `Republic of Korea Meteorological Engineer Certification Exam'.
    You can clearly distinguish between general meteorological knowledge accepted worldwide and region-specific knowledge related to Korea's weather operations, systems, topography, and administration.\\
    \\
    \textbf{\# Goal} \\
    Among the provided exam questions, select \textbf{only} those that are \textbf{not} `general meteorological knowledge applicable everywhere' but reflect `strong regional, institutional, or administrative backgrounds specific to Korea'. Return only the question ID numbers.\\
    \\
    \textbf{\# Criteria}\\
    \quad - \textbf{Korea-exclusive observation equipment/terms} (e.g., Long-wave/High-frequency proximity array observation network, ASOS-K)\\
    \quad - \textbf{Korea Meteorological Administration (KMA) administrative procedures/laws/notices} (e.g., Special advisory criteria, observation reporting time regulations)\\
    \quad - \textbf{Korean topographical/climatic specifics} (e.g., Characteristics of snow clouds on the West Coast, typhoon landfall path statistics)\\
    \quad - \textbf{Unique vocabulary, units, or abbreviations} used only in Korea or included in the exam scope\\
    \quad - If none of the above exist and the question deals with international common standards like ICAO/WMO, treat it as `General Global'\\
    \\
    \textbf{\# Output Format}\\
    \quad - Output \textbf{only numbers} without any explanation \\
    \quad Example: `4' \\
    \quad - If the list is empty, write `None' \\
    \\
    \textbf{\# Reasoning Process}\\
    \quad - Read each question and decide strictly based on the `Criteria' above \textbf{internally}.\\
    \quad - In the final answer, leave \textbf{only the result numbers} and do not write intermediate reasoning or reasons.
    \end{promptbox}
    \captionof{figure}{\textbf{English translation of the system prompt used by ChatGPT-4.1 to filter Korea-specific questions}}
    \label{fig:prompt_k_identification_gpt_eng}
\end{figure*}
\begin{figure*}[ht]
    \begin{promptbox}[System Prompt for Reference Rationale Generation]
    당신은 한국의 기상 전문가입니다. 당신의 임무는 \textbf{수석 기상 예보관이자 문제 출제위원}으로서 주어진 객관식 문제를 평가하고, 올바른 정답 선택지를 고른 뒤 다음 두 단계를 반드시 순차적으로 수행하는 것입니다.\\
    \\
    \textbf{1. 핵심 추론 근거 생성(한 문장):}\\
    \quad - (정확성) 반드시 정답이 왜 맞는지 추론 내용이 기상학적 사실에 부합하게 설명하세요.\\
    \quad - (논리성) 추론이 정답을 명확하고 직접적으로 뒷받침하도록 설명하세요.\\
    \quad - (완결성) 문제의 핵심을 회피하지 않고 설명하세요. 가능하면, 가장 그럴듯한 오답 선택지가 왜 틀렸는지도 간단히 언급하여 정답의 논리를 강화하세요.\\
    \quad - (간결성) 불필요한 내용 없이 한 문장으로 잘 요약하여 작성하세요.\\
    \\
    \textbf{2. 정답 결정:}\\
    \quad - 충분한 추론을 바탕으로, 최종적으로 정답 번호(선택지 번호)를 명확히 결정하세요.\\
    \\
    \textbf{응답 규칙}\\
    \quad - 반드시 아래 JSON 형식만 사용하여 답변해주세요.\\
    \quad - 추론 근거가 반드시 먼저 도출되고, 그 다음에 정답을 명시해야 합니다.\\
    \quad - 각 항목은 모두 반드시 포함하며, 변수명 및 구조는 아래와 같이 한글로 작성해야 합니다.\\
    
    \begin{lstlisting}
    {
      {"(*생성된\_추론\_근거*)": "{(*한 문장, 논리적 설명. 정답이 왜 맞고 주요 오답이 왜 틀렸는지 포함. 학생이 쉽게 이해할 수 있게 간결하게 작성.*)}",
      "(*정답*)": "{(*계산된 정답 번호(숫자만)*)}"
    }
    \end{lstlisting}
    \end{promptbox}
    \captionof{figure}{\textbf{System prompt used to generate a reasoning rationale}}
    \label{fig:prompt_rationale_generation}

    \vspace{1cm}
    
    \begin{promptbox}[English Translation of the System Prompt for Reference Rationale Generation]
    You are a meteorology expert in Korea. Your mission is to act as a \textbf{Senior Chief Forecaster and Exam Setter} to evaluate the given multiple-choice question, select the correct answer, and strictly perform the following two steps sequentially.\\
    \\
    \textbf{1. Generate Core Reasoning Rationale (One Sentence):}\\
    \quad - (Factuality) You must explain why the answer is correct based on meteorological facts.\\
    \quad - (Logicality) Explain the reasoning so that it clearly and directly supports the answer.\\
    \quad - (Depth) Do not avoid the core of the question. If possible, briefly mention why the most plausible distractors are wrong to strengthen the logic of the correct answer.\\
    \quad - (Clarity) Summarize well into one sentence without unnecessary content.\\
    \\
    \textbf{2. Determine Correct Answer:}\\
    \quad - Based on sufficient reasoning, clearly determine the final answer number (option number).\\
    \\
    \textbf{Response Rules}\\
    \quad - You must answer strictly using the JSON format below.\\
    \quad - The reasoning rationale must be derived first, followed by the answer.\\
    \quad - All fields must be included, and variable names and structure must be written as shown below.\\
    
    \begin{lstlisting}
    {
      "(*Generated\_Rationale*)": "{(*One logical sentence. Explain why the answer is correct and why key distractors are wrong. Write concisely for easy student understanding.*)}",
      "(*Answer*)": "{(*Calculated answer number (digits only)*)}"
    }
    \end{lstlisting}
    \end{promptbox}
    \captionof{figure}{\textbf{English translation of the system prompt used to generate a reasoning rationale}}
    \label{fig:prompt_rationale_generation_eng}
\end{figure*}

\begin{figure*}[ht]
    \begin{promptbox}[System Prompt for Reasoning and Answer Generation (Open-Source LLM)]
    \textbf{역할 (Role)}\\
    당신은 한국의 기상 전문가입니다. 다음 객관식 문항에 답하세요. \\
    질문 중 `우리나라'란 `한국'을 의미합니다. 한국의 지리적 특성을 반영해서 답변해주세요. \\
    \\
    \textbf{임무 (Task)}\\
    주어진 객관식 문제에 대해, 먼저 당신의 모든 생각 과정을 자유롭게 기록한 후, 그 내용을 바탕으로 채점관이 명확히 이해할 수 있도록 \textbf{정답에 대한 핵심 추론 근거와 최종 정답을 JSON 형식으로 제출}해야 합니다.\\
    \\
    \textbf{사고 및 응답 절차 (Thinking and Response Procedure)}\\
    당신은 반드시 아래의 2단계 절차에 따라 응답해야 합니다.\\
    \\
    \textbf{1단계: 자유로운 사고 (\textless scratchpad\textgreater)}\\
    \quad - 먼저 \textless scratchpad\textgreater 태그 안에서 자유롭게 사고합니다. 이 공간은 당신만의 생각 정리 공간입니다.\\
    \quad - 문제의 핵심 개념을 분석하고, 관련된 기상학적 지식을 나열하고, 각 선택지가 왜 맞고 틀리는지(O/X) 등 당신의 모든 생각의 흐름을 그대로 기록하세요.\\
    \quad - \textbf{중요}: 이 \textless scratchpad\textgreater의 내용은 최종 채점에 반영되지 않습니다. 형식에 구애받지 말고 마음껏 생각하세요.\\
    \\
    \textbf{2단계: 최종 답변 생성 (JSON)}\\
    \quad - \textless scratchpad\textgreater 작성이 끝나면, 당신의 생각 과정을 다시 검토하세요.\\
    \quad - 그 내용을 바탕으로, 아래의 지시사항에 맞춰 \textbf{최종 답변을 JSON 형식으로 생성}합니다.\\
    \\
    \quad 1. \textbf{(``생성된\_추론\_근거'')}: 어떤 논리적 과정을 통해 정답을 선택했는지 \textbf{간결한 문단 형태}로 작성합니다. \textless scratchpad\textgreater의 내용을 그대로 복사하지 말고, 핵심만 요약하고 재구성해야 합니다. (정확성, 논리성, 간결성, 핵심 파악 기준 고려)\\
    \quad 2. \textbf{(``정답'')}: 위 추론에 근거하여 최종 정답이라고 생각하는 선택지 번호를 하나 고릅니다. 반드시 정수 숫자만 채워서 출력해야 합니다.\\
    \\
    \textbf{전체 응답 형식 (Full Response Format)}\\
    \quad - 당신의 답변은 \textless scratchpad\textgreater 블록과 \textbf{JSON 코드 블록}, 두 부분으로 구성되어야 합니다.\\
    \quad - 아래 예시와 같이 \textless scratchpad\textgreater가 먼저 제시되고, 그 바로 다음에 \textbf{반드시 JSON 코드 블록으로 감싸진} 최종 답변이 와야 합니다.\\
    
    \begin{lstlisting}
    <scratchpad>
    (*{여기에 당신의 모든 생각 과정을 자유롭게 서술합니다. 예시: F = (C * 9/5) + 32 공식 사용...}*)
    </scratchpad>
    ```json
    {
      "(*{생성된\_추론\_근거}*)": "(*{섭씨 30도를 화씨로 변환하는 공식 F = (C * 9/5) +32를 사용하면 화씨 86도가 된다. 86은 30보다 큰 값이므로, 동일 온도를 나타낼 때 화씨 온도계의 수은 기둥이 섭씨 온도계보다 더 높이 올라간다.}*)",
      "(*{정답}*)": 1
    }```
    \end{lstlisting} 
    \end{promptbox}
    \caption{\textbf{System prompt for open-source LLMs requiring a two-step process:} free reasoning in a scratchpad followed by a structured JSON output}
    \label{fig:prompt_reasoning_generation_oss}
\end{figure*}

\begin{figure*}[ht]
    \begin{promptbox}[English Translation of the System Prompt for Reasoning and Answer Generation (Open-Source LLM)]
    \textbf{Role}\\
    You are a meteorology expert in Korea. Answer the following multiple-choice question. \\
    In the question, `our country' refers to `Korea'. Please answer reflecting the geographical characteristics of Korea. \\
    \\
    \textbf{Task}\\
    For the given multiple-choice question, you must first record your entire thought process freely, and then based on that content, \textbf{submit the core reasoning rationale for the answer and the final answer in JSON format} so that the grader can clearly understand.\\
    \\
    \textbf{Thinking and Response Procedure}\\
    You must respond according to the following 2-step procedure.\\
    \\
    \textbf{Step 1: Free Thinking (\textless scratchpad\textgreater)}\\
    \quad - First, think freely inside the \textless scratchpad\textgreater tags. This is your personal space for organizing thoughts.\\
    \quad - Analyze the core concepts of the question, list relevant meteorological knowledge, and record your flow of thought, including why each option is correct or incorrect (O/X).\\
    \quad - \textbf{Important}: The content of this \textless scratchpad\textgreater is not reflected in the final grading. Think freely without formatting constraints.\\
    \\
    \textbf{Step 2: Final Answer Generation (JSON)}\\
    \quad - Once writing in \textless scratchpad\textgreater is finished, review your thought process.\\
    \quad - Based on that, \textbf{generate the final answer in JSON format} according to the instructions below.\\
    \\
    \quad 1. \textbf{(``Generated\_Rationale'')}: Write a \textbf{concise paragraph} explaining the logical process used to select the answer. Do not copy the \textless scratchpad\textgreater content directly; summarize and restructure only the key points. (Consider accuracy, logic, conciseness, and core identification criteria.)\\
    \quad 2. \textbf{(``Answer'')}: Select the option number you think is the final correct answer based on the reasoning above. Must output only an integer.\\
    \\
    \textbf{Full Response Format}\\
    \quad - Your response must consist of two parts: a \textless scratchpad\textgreater block and a \textbf{JSON code block}.\\
    \quad - As shown in the example below, the \textless scratchpad\textgreater must be presented first, immediately followed by the final answer \textbf{wrapped in a JSON code block}.\\
    
    \begin{lstlisting}
    <scratchpad>
    (*{Describe your entire thought process freely here. Example: Using formula F = (C * 9/5) + 32...}*)
    </scratchpad>
    ```json
    {
      "(*{Generated\_Rationale}*)": "(*{Using the formula F = (C * 9/5) + 32 to convert 30 degrees Celsius to Fahrenheit gives 86 degrees Fahrenheit. Since 86 is greater than 30, the mercury column of the Fahrenheit thermometer rises higher than that of the Celsius thermometer for the same temperature.}*)",
      "(*{Answer}*)": 1
    }```
    \end{lstlisting} 
    \end{promptbox}
    \caption{\textbf{English translation of the System prompt for open-source LLMs}}
    \label{fig:prompt_reasoning_generation_oss_eng}
\end{figure*}
\begin{figure*}[ht]
    \begin{promptbox}[System Prompt for LLM-as-a-Judge Evaluation (Gemini-2.5-Pro)]
    \fontsize{8.5pt}{8.5pt}\selectfont
    \textbf{\#\#\# 역할 (Role)}\\
    당신은 기상학 분야의 깊은 전문 지식을 가진, 매우 엄격하고 공정한 AI 추론 능력 평가 위원장입니다. 당신은 감정이나 편향 없이, 오직 주어진 평가 기준에만 근거하여 객관적으로 채점해야 합니다.\\
    \\
    \textbf{\#\#\# 임무 (Task)}\\
    주어진 \textless 문제 정보\textgreater, \textless 전문가 검증 참조 자료\textgreater, 그리고 \textless 평가 대상 답변\textgreater을 바탕으로, `수험생 AI'가 제출한 답변의 품질을 평가하고 그 결과를 지정된 JSON 형식으로 출력해야 합니다.\\
    \\
    \textbf{\#\#\# 입력 정보 (Input Information)}\\
    1. \textbf{\textless 문제 정보\textgreater}: 평가의 맥락이 되는 원본 객관식 문제입니다. (질문, 선택지, 정답 포함)\\
    2. \textbf{\textless 전문가 검증 참조 자료\textgreater}: 100\% 사실이 검증된 모범 해설 자료입니다. 이 자료는 \textless 평가 대상 답변\textgreater 의 '사실 오류'나 '환각(Hallucination)'을 탐지하는 절대 기준으로 사용해야 합니다.\\
    3. \textbf{\textless 평가 대상 답변\textgreater}: 당신이 채점해야 할 '수험생 AI'가 생성한 추론 과정 및 정답입니다.\\
    \\
    \textbf{\#\#\# 사고 과정 (Step-by-Step Thinking Process)}\\
    당신은 평가를 수행하기 전에 반드시 다음의 사고 과정을 거쳐야 합니다.\\
    1. 모든 입력 정보를 충분히 숙지합니다.\\
    2. \textbf{[사실 확인]}: \textless 평가 대상 답변\textgreater 의 모든 주장을 \textless 전문가 검증 참조 자료\textgreater 와 비교하여 사실 오류가 있는지 먼저 확인하고 목록을 작성합니다.\\
    3. \textbf{[개별 축 평가]}: 아래의 4가지 `평가 기준 및 척도'를 하나씩 읽고, 각 기준에 따라 \textless 평가 대상 답변\textgreater 이 몇 점에 해당하는지 근거와 함께 판단합니다.\\
    4. \textbf{[종합 및 형식화]}: 모든 판단이 끝나면, 그 내용을 종합하여 최종 출력 JSON 형식을 작성합니다.\\
    \\
    \textbf{\#\#\# 평가 기준 및 척도 (Evaluation Criteria and Scale)}\\
    각 평가 축에 대해 \textbf{1점(매우 부족)부터 5점(매우 우수)까지} 정수 점수를 부여합니다.\\
    \\
     * \textbf{1) 사실적 정확성 (Factual Accuracy) [1-5점]}\\ 
    \hspace*{1em} * 추론 내용이 기상학적 사실에 완벽하게 부합하며 오류가 없는가?\\
    \hspace*{1em} * \textbf{5점:} 모든 내용이 \textless 전문가 검증 참조 자료 \textgreater 에 기반하여 완벽하게 정확함.\\
    \hspace*{1em} * \textbf{3점:} 핵심 논리는 맞지만, 결론에 영향을 미치지 않는 사소한 오류나 부정확한 표현이 포함됨.\\
    \hspace*{1em} * \textbf{1점:} 결론의 정당성을 훼손하는 중대한 사실 오류(환각)가 포함됨.\\
    \\
     * \textbf{2) 논리적 완결성 (Logical Soundness) [1-5점]}\\ 
    \hspace*{1em} * 제시된 근거가 결론(정답)을 도출하기에 논리적으로 충분하고 필연적인가? \\
    \hspace*{1em} * \textbf{5점:} 근거와 결론 사이에 논리적 비약이나 누락이 전혀 없이 완벽하게 연결됨.\\
    \hspace*{1em} * \textbf{3점:} 핵심적인 연결고리는 존재하나, 일부 설명이 생략되거나 추론 과정이 다소 불친절함.\\
    \hspace*{1em} * \textbf{1점:} 근거가 결론을 뒷받침하지 못하거나, 명백한 논리적 오류가 존재함.\\
    \\
     * \textbf{3) 추론의 깊이 (Depth of Reasoning) [1-5점]}\\
    \hspace*{1em} * 문제의 핵심 원리를 깊이 있게 이해하고 다각적으로 분석하였는가?\\
    \hspace*{1em} * \textbf{5점:} 정답의 근거뿐만 아니라, \textbf{매력적인 오답 선택지가 왜 오답인지}에 대한 반박까지 포함하여 깊이 있는 이해를 보여줌.\\
    \hspace*{1em} * \textbf{3점:} 정답에 대한 핵심적인 설명은 제시했지만, 오답에 대한 고려 등 다각적인 분석은 부족함.\\
    \hspace*{1em} * \textbf{1점:} 피상적인 수준의 단편적인 사실만 나열하여 깊이가 없음.\\
    \\
     * \textbf{4) 표현의 명확성 (Clarity \& Conciseness) [1-5점]}\\
    \hspace*{1em} * 추론 설명이 군더더기 없이 명확하고 이해하기 쉬운가?\\
     \hspace*{1em} * \textbf{5점:} 불필요한 내용 없이 핵심만 간결하게 전
    달하면서도, 전문가가 아닌 학습자도 쉽게 이해할 수 있을 만큼 명확하게 서술됨.\\
    \hspace*{1em} * \textbf{3점:} 의미 전달은 되지만, 일부 문장이 장황하거나 불필요한 정보가 포함되어 있음.\\
    \hspace*{1em} * \textbf{1점:} 문장이 복잡하고 이해하기 어렵거나, 질문의 의도와 관련 없는 내용이 많음.\\

    \textbf{\#\#\# 출력 형식 (Output Format)}\\
     * 반드시 아래 JSON 구조를 정확히 준수하여 응답해야 합니다. \\
     * JSON 외의 부가 설명이나 텍스트는 추가하지 마세요.
     
    \begin{lstlisting}
    ```json
    {
      ``(*{사실\_오류\_목록}*)": [
        ``{ < 평가 대상 답변 > 에서 발견된 첫 번째 사실 오류 또는 환각 ( 문장 인용 ) }",
        ``{ 두 번째 사실 오류... ( 없으면 빈 리스트 [ ] 로 출력 ) }"
      ],
      ``(*{평가\_점수}*)": {
        ``(*{사실적\_정확성}*)": {1~5 사이의 정수 점수},
        ``(*{논리적\_완결성}*)": {1~5 사이의 정수 점수},
        ``(*{추론의\_깊이}*)": {1~5 사이의 정수 점수},
        ``(*{표현의\_명확성}*)": {1~5 사이의 정수 점수}
      },
      ``(*{평가\_사유}*)": ``{점수를 부여한 핵심적인 이유를 종합적으로 서술. 정확성 점수가 낮으면 사실 오류를 반드시 언급할 것.}"
    }```
    \end{lstlisting}
    \end{promptbox}
    \caption{\textbf{System prompt used by the evaluator LLM (Gemini-2.5-Pro) to assess reasoning quality across four dimensions:} factual accuracy, logical soundness, depth of reasoning, and clarity.}
    \label{fig:prompt_llm_judge_score}
\end{figure*}

\begin{figure*}[ht]
    \begin{promptbox}[English Translation of the System Prompt for LLM-as-a-Judge Evaluation (Gemini-2.5-Pro)]
    \fontsize{8.5pt}{8.5pt}\selectfont
    \textbf{\#\#\# Role}\\
    You are a highly strict and impartial Evaluation Chair for AI Reasoning Capabilities, possessing deep expertise in the field of meteorology. You must grade objectively, devoid of emotion or bias, based solely on the provided evaluation criteria.\\
    \\
    \textbf{\#\#\# Task}\\
    Based on the provided \textless Problem Information \textgreater , \textless Expert Verification Reference Material \textgreater , and \textless Response for Evaluation \textgreater , assess the quality of the answer submitted by the `Examinee AI' and output the results in the specified JSON format.\\
    \\
    \textbf{\#\#\# Input Information}\\
    1. \textbf{\textless Problem Information \textgreater}: The original multiple-choice question serving as the context for evaluation (includes the question, options, and correct answer).\\
    2. \textbf{\textless Expert Verification Reference Material \textgreater}: Model explanation material with 100\% verified facts. This material must be used as the absolute standard for detecting `factual errors' or `hallucinations' in the \textless Response for Evaluation \textgreater. \\
    3. \textbf{\textless Response for Evaluation \textgreater}: The reasoning process and answer generated by the `Examinee AI' that you are required to grade.\\
    \\
    \textbf{\#\#\# Step-by-Step Thinking Process}\\
    You must go through the following thinking process before performing the evaluation.\\
    1. Fully understand all input information.\\
    2. \textbf{[Fact Check]}: Compare every claim in the \textless Response for Evaluation \textgreater against the \textless Expert Verification Reference Material \textgreater to first identify any factual errors and compile a list.\\
    3. \textbf{[Individual Axis Evaluation]}: Read the four `Evaluation Criteria and Scale' below one by one, and determine the score for the \textless Response for Evaluation \textgreater based on each criterion, along with the rationale.\\
    4. \textbf{[Synthesis and Formatting]}: Once all judgments are complete, synthesize the content to create the final output JSON format.\\
    \\
    \textbf{\#\#\# Evaluation Criteria and Scale}\\
    Assign an integer score \textbf{from 1 (Very Poor) to 5 (Excellent)} for each evaluation axis according to the criteria below.\\
    \\
    * \textbf{1) Factual Accuracy [1-5 points]}\\
    \hspace*{1em} * Does the reasoning perfectly align with meteorological facts without errors?\\
    \hspace*{1em} * \textbf{5 points:} All content is perfectly accurate based on the \textless Expert Verification Reference Material \textgreater. \\
    \hspace*{1em} * \textbf{3 points:} The core logic is correct, but minor errors or inaccurate expressions that do not affect the conclusion are included. \\
    \hspace*{1em} * \textbf{1 point:} Major factual errors (hallucinations) that undermine the legitimacy of the conclusion are included.\\
    \\
    * \textbf{2) Logical Soundness [1-5 points]}\\
    \hspace*{1em} * Is the provided evidence logically sufficient and inevitable for deriving the conclusion (answer)?\\
    \hspace*{1em} * \textbf{5 points:} Perfectly connected without any logical leaps or omissions between evidence and conclusion.\\
    \hspace*{1em} * \textbf{3 points:} The core link exists, but some explanations are omitted, or the reasoning process is somewhat unfriendly.\\
    \hspace*{1em} * \textbf{1 point:} The evidence fails to support the conclusion, or obvious logical errors exist.\\
    \\
    * \textbf{3) Depth of Reasoning [1-5 points]}\\
    \hspace*{1em} * Did it understand the core principles of the problem deeply and analyze it from multiple angles?\\
    \hspace*{1em} * \textbf{5 points:} Demonstrates deep understanding by including not only the basis for the correct answer but also rebuttals as to why attractive distractors (wrong options) are incorrect.\\
    \hspace*{1em} * \textbf{3 points:} Provided the core explanation for the correct answer, but lacks multi-faceted analysis such as consideration of wrong options.\\
    \hspace*{1em} * \textbf{1 point:} Lacks depth by merely listing superficial and fragmentary facts.\\
    \\
    * \textbf{4) Clarity \& Conciseness [1-5 points]}\\
    \hspace*{1em} * Is the reasoning explanation clear and easy to understand without clutter?\\
    \hspace*{1em} * \textbf{5 points:} Conveys only the core points concisely without unnecessary content, stated clearly enough for a non-expert learner to understand easily.\\
    \hspace*{1em} * \textbf{3 points:} Meaning is conveyed, but some sentences are verbose or include unnecessary information.\\
    \hspace*{1em} * \textbf{1 point:} Sentences are complex and difficult to understand, or there is a lot of content irrelevant to the intent of the question.\\
    \\
    \textbf{\#\#\# Output Format}\\
     - You must strictly adhere to the JSON structure below.\\
     - Do not add any additional explanations or text outside the JSON.
     
    \begin{lstlisting}
    ```json
    {
      ``(*{List \_of\_Factual\_Errors}*)": [
        ``{ First factual error or hallucination found in < Response for Evaluation >  ( quote the sentence ) }",
        ``{ Second factual error... ( Output an empty list [ ] if none ) }"
      ],
      ``(*{Evaluation\_Scores}*)": {
        ``(*{Factual\_Accuracy}*)": {Integer score between 1-5},
        ``(*{Logical\_Soundness}*)": {Integer score between 1-5},
        ``(*{Depth\_of\_Reasoning}*)": {Integer score between 1-5},
        ``(*{Clarity\_and\_Conciseness}*)": {Integer score between 1-5}
      },
      ``(*{Evaluation\_Rationale}*)": ``{ Comprehensively describe the core reasons for assigning the scores above for each axis. If the `Accuracy' score is low, you must explicitly mention what factual errors occurred. }"
    }```
    \end{lstlisting}
    \end{promptbox}
    \caption{\textbf{System prompt used by the evaluator LLM (Gemini-2.5-Pro) to assess reasoning quality across four dimensions:} factual accuracy, logical soundness, depth of reasoning, and clarity.}
    \label{fig:prompt_llm_judge_score_eng}
\end{figure*}

\begin{figure*}[ht]
    \begin{promptbox}[Standard System Prompt (Baseline)]
    당신은 기상 전문가입니다. 다음 객관식 문항에 답하세요. \\
    보기 중에서 가장 알맞은 선택지를 고르고, 다음과 같은 형식의 JSON으로 출력하세요: \\
    \texttt{```json} \{`정답': `\{선택지 번호\}'\}\texttt{```} \\
    답변 이외에 다른 어떤 텍스트도 출력하지 마세요.
    \end{promptbox}
    \captionof{figure}{\textbf{Standard baseline system prompt using simple zero-shot instructions without specific regional context}}
    \label{fig:prompt_standard}
    
    \vspace{0.5cm}
    
    \begin{promptbox}[English Translation of the Standard System Prompt (Baseline)]
    You are a meteorology expert. Answer the following multiple-choice question. \\
    Choose the most appropriate option from the choices and output it in the following JSON format: \\
    \texttt{```json} \{`Answer': `\{Option Number\}'\}\texttt{```} \\
    Do not output any text other than the answer.
    \end{promptbox}
    \captionof{figure}{\textbf{English translation of the standard baseline system prompt using simple zero-shot instructions without specific regional context}}
    \label{fig:prompt_standard_eng}
\end{figure*}

\begin{figure*}[ht]
    \begin{promptbox}[Advanced System Prompt (Role + Context)]
    당신은 한국의 기상 전문가입니다. 다음 객관식 문항에 답하세요. \\
    질문 중 `우리나라'란 `한국'을 의미합니다. 한국의 지리적 특성을 반영해서 답변해주세요. \\
    보기 중에서 가장 알맞은 선택지를 고르고, 다음과 같은 형식의 JSON으로 출력하세요: \\
    \texttt{```json} \{`정답': `\{선택지 번호\}'\}\texttt{```} \\
    답변 이외에 다른 어떤 텍스트도 출력하지 마세요.
    \end{promptbox}
    \captionof{figure}{\textbf{Advanced prompt with added role definition and specific context regarding Korean geography}}
    \label{fig:prompt_advanced}
    
    \vspace{0.5cm}
    
    \begin{promptbox}[English Translation of the Advanced System Prompt (Role + Context)]
    You are a meteorology expert in Korea. Answer the following multiple-choice question. \\
    In the question, `our country' refers to `Korea'. Please answer reflecting the geographical characteristics of Korea. \\
    Choose the most appropriate option from the choices and output it in the following JSON format: \\
    \texttt{```json} \{`Answer': `\{Option Number\}'\}\texttt{```} \\
    Do not output any text other than the answer.
    \end{promptbox}
    \captionof{figure}{\textbf{English translation of the advanced prompt with added role definition and specific context regarding Korean geography}}
    \label{fig:prompt_advanced_eng}
\end{figure*}

\begin{figure*}[ht]
    \begin{promptbox}[Text-Only/Multimodal MQQA User Prompt Template]
    질문: \textbraceleft \textbraceleft Question\_text\textbraceright \textbraceright \\
    \textit{[\textbraceleft \textbraceleft Question\_image\textbraceright \textbraceright]} \\
    1. \textbraceleft \textbraceleft Choice\_1\_text\textbraceright \textbraceright \\
    \textit{[\textbraceleft \textbraceleft Choice\_1\_image\textbraceright \textbraceright]} \\
    2. \textbraceleft \textbraceleft Choice\_2\_text\textbraceright \textbraceright \\
    \textit{[\textbraceleft \textbraceleft Choice\_2\_image\textbraceright \textbraceright]} \\
    3. \textbraceleft \textbraceleft Choice\_3\_text\textbraceright \textbraceright \\
    \textit{[\textbraceleft \textbraceleft Choice\_3\_image\textbraceright \textbraceright]} \\
    4. \textbraceleft \textbraceleft Choice\_4\_text\textbraceright \textbraceright \\
    \textit{[\textbraceleft \textbraceleft Choice\_4\_image\textbraceright \textbraceright]} \\
    \end{promptbox}
    \captionof{figure}{\textbf{Text-Only/Multimodal MCQA user prompt template.} The placeholders enclosed in square brackets (e.g., \textit{[{{Question\_image}}]}) denote optional fields that are populated only when the corresponding image exists in the dataset. This single template covers all four modality configurations (i.e., text-only, image-in-question, image-in-choices, and images-in-both).}
    \label{fig:user_prompt_multimodal}
    
    
\end{figure*}

\begin{figure*}[ht]
    \centering
    \begin{promptbox}[Reasoning MCQA User Prompt Template]
    ======================================== \\
    아래 정보를 바탕으로 평가를 수행하시오. \\

    --- BEGIN INPUT DATA --- \\

    \#\#\# \textless 문제 정보 \textgreater \\
    **문제 텍스트:** \\
    \textbraceleft\textbraceleft Question\_text\textbraceright\textbraceright \\
    \textit{[**문제 이미지:** \textbraceleft\textbraceleft Question\_image\textbraceright\textbraceright]} \\

    **선택지:** \\
    1. \textbraceleft\textbraceleft Choice\_1\_Text\textbraceright\textbraceright \\
    \textit{[**(선택지 1 이미지):** \textbraceleft\textbraceleft Choice\_1\_image\textbraceright\textbraceright]} \\
    2. \textbraceleft\textbraceleft Choice\_2\_Text\textbraceright\textbraceright \\
    \textit{[**(선택지 2 이미지):** \textbraceleft\textbraceleft Choice\_2\_image\textbraceright\textbraceright]} \\
    3. \textbraceleft\textbraceleft Choice\_3\_Text\textbraceright\textbraceright \\
    \textit{[**(선택지 3 이미지):** \textbraceleft\textbraceleft Choice\_3\_image\textbraceright\textbraceright]} \\
    4. \textbraceleft\textbraceleft Choice\_4\_Text\textbraceright\textbraceright \\
    \textit{[**(선택지 4 이미지):** \textbraceleft\textbraceleft Choice\_4\_image\textbraceright\textbraceright]} \\

    **정답:** \textbraceleft\textbraceleft Correct\_Answer\textbraceright\textbraceright \\

    \#\#\# \textless 전문가 검증 참조 자료 \textgreater \\
    \textbraceleft\textbraceleft Rationale\_Content\textbraceright\textbraceright \textit{\% populates `(자료 없음)' if wo\_rationale is True} \\

    \#\#\# \textless 평가 대상 답변 \textgreater \\
    **생성된 추론 근거:** \textbraceleft\textbraceleft Generated\_Reasoning\textbraceright\textbraceright \\
    **답안:** \textbraceleft\textbraceleft Predicted\_Answer\textbraceright\textbraceright \\

    --- END INPUT DATA ---
    \end{promptbox}
    \captionof{figure}{\textbf{Reasoning MCQA user prompt template.} The placeholders enclosed in square brackets denote optional image fields populated based on data availability. Additionally, the rationale field defaults to `(자료 없음)' (No Data) when the expert rationale is withheld (\textit{w/o rationale} setting).} 
    \label{fig:user_prompt_reasoning}

\end{figure*}

\begin{table*}[ht]
  \centering
  \caption{\textbf{K-MetBench performance scores across all models and subsets.} Models are sorted by accuracy. All accuracy metrics range from 0 to 100, while the reasoning score (\textbf{\googr{Reas.}}) ranges from 4 to 20. \textbf{Bold} values indicate the highest scores in each column for proprietary and open-source models, respectively. (Acc.: Accuracy, \textsf{K}: Korean model, \textsf{V}: Vision language model, \textsf{R}: Reasoning model, \texttt{Inst.}: Instruct)}
  \label{tab:kmetbench_scores_full}
  \tiny
  \setlength{\aboverulesep}{1.5pt}
  \setlength{\belowrulesep}{1.5pt}
  \setlength{\tabcolsep}{2pt}
  \renewcommand{\arraystretch}{1.05}
  \begin{adjustbox}{width=\textwidth,totalheight=\textheight,keepaspectratio}
  \begin{tabular}{@{}l p{0.24\textwidth} c c c rrrrrrrrrr@{}}
  \toprule
  \multirow{2}{*}{\textbf{Type}} & \multirow{2}{*}{\textbf{Model}} & \multicolumn{3}{c}{\textbf{Flags}} & \multicolumn{1}{c}{\textbf{Acc.}} & \multicolumn{1}{c}{\textbf{\googr{Reas.}}} & \multicolumn{1}{c}{\textbf{\googy{Geo-Cult.}}} & \multicolumn{2}{c}{\textbf{\googb{Modality}}} & \multicolumn{5}{c}{\textbf{\googg{Granularity (P1--P5)}}} \\
  \cmidrule(lr){3-5}\cmidrule(lr){6-6}\cmidrule(lr){7-7}\cmidrule(lr){8-8}\cmidrule(lr){9-10}\cmidrule(lr){11-15}
   &  & \textsf{K} & \textsf{V} & \textsf{R} &  &  & Korean & Text & Multi & P1 & P2 & P3 & P4 & P5 \\
   
  \midrule
  & \includegraphics[height=0.8em]{icon/gemini.png} \texttt{gemini-3-pro-preview} (Thinking) &   &  \textsf{V} & \textsf{R} & \textbf{93.7} & \textbf{18.01} & \textbf{90.4} & \textbf{94.6} & \textbf{75.6} & \textbf{92.5} & \textbf{97.9} & \textbf{94.2} & \textbf{92.8} & \textbf{91.6} \\
  \rowcolor{palegray}
  \cellcolor{white} & \includegraphics[height=0.8em]{icon/openai.pdf} \texttt{gpt-5.2} (Thinking) &   &  \textsf{V} & \textsf{R} & 87.8 & 17.33 & 80.8 & 90.6 & 29.3 & 86.3 & 93.4 & 88.0 & 86.2 & 85.3 \\
  \multirow{-3}{*}{\textbf{Proprietary}} 
  & \includegraphics[height=0.8em]{icon/openai.pdf} \texttt{gpt-5.2} &  & \textsf{V} &  & 77.6 & 17.39 & 75.3 & 79.0 & 50.0 & 77.2 & 81.3 & 71.9 & 81.4 & 76.3 \\
  
  \midrule
  
    & \multicolumn{14}{c}{\textbf{Multilingual Thinking Models}}\\
  \cmidrule(lr){2-15}
   & \includegraphics[height=0.5em]{icon/qwen.pdf} \texttt{Qwen3-VL-235B-A22B-Thinking} &   &  \textsf{V} & \textsf{R} & \textbf{84.4} & \textbf{17.22} & 72.6 & \textbf{86.2} & 48.8 & \textbf{81.5} & \textbf{88.6} & \textbf{87.2} & \textbf{83.2} & \textbf{82.0} \\
   \rowcolor{palegray}
   \cellcolor{white} & \includegraphics[height=0.5em]{icon/qwen.pdf} \texttt{Qwen3-VL-32B-Thinking} &   &  \textsf{V} & \textsf{R} & 78.6 & 16.17 & 60.3 & 79.9 & \textbf{51.2} & 74.3 & 85.2 & 78.8 & 78.7 & 76.3 \\
   & \includegraphics[height=0.8em]{icon/cohere.png} \texttt{command-a-reasoning-08-2025} &   &  & \textsf{R} & 77.8 & 14.12 & 74.6 & 77.8 & - & 73.4 & 85.2 & 73.8 & 78.8 & 78.5 \\
   \rowcolor{palegray}
   \cellcolor{white} & \includegraphics[height=0.8em]{icon/openai.pdf} \texttt{gpt-oss-120b} &   &  & \textsf{R} & 77.3 & 16.12 & 62.0 & 77.3 & - & 72.5 & 85.8 & 76.5 & 77.4 & 74.9 \\
   & \includegraphics[height=0.5em]{icon/qwen.pdf} \texttt{Qwen3-30B-A3B-Thinking-2507} &   &  & \textsf{R} & 76.7 & 15.76 & 67.6 & 76.7 & - & 75.5 & 82.1 & 75.6 & 74.9 & 75.9 \\
   \rowcolor{palegray}
   \cellcolor{white} & \includegraphics[height=0.5em]{icon/qwen.pdf} \texttt{Qwen3-VL-30B-A3B-Thinking} &   &  \textsf{V} & \textsf{R} & 74.9 & 15.16 & 68.5 & 76.3 & 45.1 & 70.5 & 77.4 & 74.1 & 76.6 & 76.0 \\
   & \includegraphics[height=0.5em]{icon/qwen.pdf} \texttt{Qwen3-14B} &   &  & \textsf{R} & 73.7 & 15.25 & 60.6 & 73.7 & - & 70.9 & 84.3 & 72.4 & 70.2 & 71.7 \\
   \rowcolor{palegray}
   \cellcolor{white} &  \includegraphics[height=0.5em]{icon/qwen.pdf} \texttt{Qwen3-VL-8B-Thinking} &   &  \textsf{V} & \textsf{R} & 71.7 & 10.33 & 61.6 & 73.3 & 39.0 & 66.2 & 79.8 & 70.5 & 71.3 & 71.6 \\
   & \includegraphics[height=0.8em]{icon/openai.pdf} \texttt{gpt-oss-20b} &   &  & \textsf{R} & 71.5 & 13.55 & 60.6 & 71.5 & - & 65.7 & 82.4 & 71.8 & 72.2 & 65.8 \\
   \rowcolor{palegray}
   \cellcolor{white} &  \includegraphics[height=0.5em]{icon/qwen.pdf} \texttt{Qwen3-8B} &   &  & \textsf{R} & 70.1 & 13.31 & 49.3 & 70.1 & - & 69.5 & 80.2 & 69.1 & 65.6 & 66.8 \\
   & \includegraphics[height=0.5em]{icon/qwen.pdf} \texttt{Qwen3-4B-Thinking-2507} &   &  & \textsf{R} & 67.8 & 13.28 & 60.6 & 67.8 & - & 63.5 & 80.8 & 64.1 & 66.7 & 65.1 \\
   \rowcolor{palegray}
   \cellcolor{white} & \includegraphics[height=0.5em]{icon/qwen.pdf} \texttt{Qwen3-VL-4B-Thinking} &   &  \textsf{V} & \textsf{R} & 66.1 & 11.58 & 54.8 & 67.0 & 47.6 & 60.1 & 80.1 & 62.7 & 64.1 & 65.0 \\
   & \includegraphics[height=0.5em]{icon/qwen.pdf} \texttt{Qwen3-32B} &   &  & \textsf{R} & 47.5 & 14.57 & 28.2 & 47.5 & - & 48.6 & 61.3 & 47.4 & 39.7 & 41.0 \\
   \rowcolor{palegray}
   \cellcolor{white} &  \includegraphics[height=0.5em]{icon/qwen.pdf} \texttt{Qwen3-1.7B} &   &  & \textsf{R} & 46.8 & 7.37 & 35.2 & 46.8 & - & 45.1 & 57.2 & 47.4 & 42.4 & 42.7 \\
   & \includegraphics[height=0.5em]{icon/qwen.pdf} \texttt{Qwen3-0.6B} &   &  & \textsf{R} & 32.2 & 4.60 & 23.9 & 32.2 & - & 30.2 & 40.9 & 32.1 & 32.0 & 25.7 \\
   \rowcolor{palegray}
   \cellcolor{white} &  \includegraphics[height=0.8em]{icon/microsoft.png} \texttt{Phi-4-mini-reasoning} &   &  & \textsf{R} & 12.6 & 4.02 & 9.9 & 12.6 & - & 14.3 & 10.7 & 10.9 & 12.7 & 14.3 \\
  \cmidrule(lr){2-15}
    & \multicolumn{14}{c}{\textbf{Multilingual Instruct Models}}\\
  \cmidrule(lr){2-15}
   & \includegraphics[height=0.5em]{icon/qwen.pdf} \texttt{Qwen3-VL-235B-A22B-Instruct} &  & \textsf{V} &  & 72.4 & 15.40 & 74.0 & 73.8 & 45.1 & 72.9 & 78.6 & 64.3 & 74.5 & 72.2 \\
   \rowcolor{palegray}
   \cellcolor{white} &  \includegraphics[height=0.5em]{icon/qwen.pdf} \texttt{Qwen3-VL-32B-Instruct} &   &  \textsf{V} &  & 67.5 & 14.85 & 61.6 & 68.7 & 41.5 & 67.8 & 72.0 & 64.3 & 69.4 & 63.8 \\
   & \includegraphics[height=0.5em]{icon/qwen.pdf} \texttt{Qwen2.5-VL-72B-Instruct} &  & \textsf{V} &  & 67.1 & 12.94 & 63.0 & 68.4 & 41.5 & 64.3 & 70.8 & 62.7 & 69.7 & 68.6 \\
   \rowcolor{palegray}
   \cellcolor{white} &  \includegraphics[height=0.5em]{icon/qwen.pdf} \texttt{Qwen3-30B-A3B-Instruct-2507} &  &  &  & 64.7 & 14.69 & 60.6 & 64.7 & - & 65.1 & 71.4 & 57.6 & 65.6 & 63.8 \\
   & \includegraphics[height=0.8em]{icon/cohere.png} \texttt{c4ai-command-a-03-2025} &  &  &  & 65.5 & 12.81 & 66.2 & 65.5 & - & 62.9 & 66.4 & 57.9 & 71.9 & 68.7 \\
   \rowcolor{palegray}
   \cellcolor{white} &  \includegraphics[height=0.5em]{icon/qwen.pdf} \texttt{Qwen3-VL-30B-A3B-Instruct} &  & \textsf{V} &  & 62.2 & 13.37 & 57.5 & 63.2 & 41.5 & 63.3 & 68.4 & 54.3 & 64.4 & 61.1 \\
   & \includegraphics[height=0.5em]{icon/qwen.pdf} \texttt{Qwen2.5-VL-32B-Instruct} &  & \textsf{V} &  & 60.1 & 10.99 & 56.2 & 61.1 & 39.0 & 60.3 & 59.9 & 56.8 & 62.8 & 60.5 \\
   \rowcolor{palegray}
   \cellcolor{white} & \includegraphics[height=0.8em]{icon/meta.pdf} \texttt{Llama-3.1-70B-Instruct} &  &  &  & 59.9 & 11.16 & 57.7 & 59.9 & - & 59.3 & 61.6 & 53.2 & 65.8 & 59.3 \\
   & \includegraphics[height=0.8em]{icon/intervl.png} \texttt{InternVL3.5-38B-Instruct} &   &  \textsf{V} & & 57.3 & 11.38 & 47.9 & 58.1 & 40.2 & 56.0 & 64.8 & 48.7 & 61.4 & 55.7 \\
   \rowcolor{palegray}
   \cellcolor{white} &  \includegraphics[height=0.8em]{icon/meta.pdf} \texttt{Llama-3.2-90B-Vision-Instruct} &  & \textsf{V} &  & 56.9 & 9.72 & 52.1 & 58.2 & 30.5 & 57.1 & 59.3 & 52.4 & 62.2 & 53.3 \\
   & \includegraphics[height=0.5em]{icon/qwen.pdf} \texttt{Qwen3-VL-8B-Instruct} &   &  \textsf{V} & & 53.8 & 12.07 & 43.8 & 54.3 & 43.9 & 54.2 & 58.1 & 49.3 & 55.9 & 51.5 \\
   \rowcolor{palegray}
   \cellcolor{white} &  \includegraphics[height=0.5em]{icon/qwen.pdf} \texttt{Qwen3-4B-Instruct-2507} &  &  &  & 51.5 & 12.32 & 45.1 & 51.5 & - & 53.8 & 52.5 & 47.6 & 52.6 & 50.8 \\
   & \includegraphics[height=0.8em]{icon/microsoft.png} \texttt{Phi-4} &  &  &  & 51.5 & 11.75 & 40.8 & 51.5 & - & 52.5 & 53.8 & 50.0 & 55.1 & 45.3 \\
   \rowcolor{palegray}
   \cellcolor{white} &  \includegraphics[height=0.5em]{icon/qwen.pdf} \texttt{Qwen3-VL-4B-Instruct} &   &  \textsf{V} &  & 51.0 & 11.55 & 46.6 & 51.1 & 48.8 & 50.7 & 56.3 & 46.0 & 53.5 & 48.5 \\
   & \includegraphics[height=0.8em]{icon/intervl.png} \texttt{InternVL3.5-14B-Instruct} &   &  \textsf{V} & & 47.9 & 9.45 & 45.2 & 48.4 & 37.8 & 44.5 & 53.6 & 44.0 & 50.3 & 47.6 \\
   \rowcolor{palegray}
   \cellcolor{white} &  \includegraphics[height=0.8em]{icon/intervl.png} \texttt{InternVL3.5-8B-Instruct} &   &  \textsf{V} & & 46.1 & 7.07 & 35.6 & 46.7 & 32.9 & 45.3 & 48.8 & 42.1 & 52.4 & 41.6 \\
   & \includegraphics[height=0.5em]{icon/qwen.pdf} \texttt{Qwen2.5-VL-7B-Instruct} &  & \textsf{V} &  & 46.1 & 7.08 & 37.0 & 46.6 & 34.1 & 49.3 & 43.1 & 42.9 & 51.9 & 42.2 \\
   \rowcolor{palegray}
   \cellcolor{white} & \includegraphics[height=0.8em]{icon/meta.pdf} \texttt{Llama-3.1-8B-Instruct} &  &  &  & 41.8 & 7.63 & 40.8 & 41.8 & - & 44.2 & 38.1 & 40.0 & 44.4 & 42.0 \\
   \cellcolor{white} & \includegraphics[height=0.8em]{icon/intervl.png} \texttt{InternVL3.5-4B-Instruct} &   &  \textsf{V} & & 41.5 & 4.81 & 24.7 & 42.1 & 29.3 & 44.5 & 44.0 & 37.9 & 43.9 & 36.8 \\
   \rowcolor{palegray}
   \cellcolor{white} &  \includegraphics[height=0.5em]{icon/qwen.pdf} \texttt{Qwen2.5-VL-3B-Instruct} &  & \textsf{V} &  & 40.9 & 4.88 & 37.0 & 41.4 & 30.5 & 41.6 & 38.6 & 40.1 & 43.9 & 40.1 \\
   & \includegraphics[height=0.8em]{icon/meta.pdf} \texttt{Llama-3.2-3B-Instruct} &  &  &  & 33.8 & 5.08 & 31.0 & 33.8 & - & 36.3 & 32.7 & 34.7 & 33.9 & 30.9 \\
   \rowcolor{palegray}
   \cellcolor{white} &  \includegraphics[height=0.8em]{icon/intervl.png} \texttt{InternVL3.5-2B-Instruct} &   &  \textsf{V} & & 31.0 & 4.35 & 24.7 & 31.2 & 26.8 & 33.8 & 30.7 & 32.3 & 29.5 & 28.4 \\
   & \includegraphics[height=0.8em]{icon/microsoft.png} \texttt{Phi-4-mini-Instruct} &  &  &  & 30.4 & 5.82 & 21.1 & 30.4 & - & 31.6 & 33.0 & 30.3 & 29.5 & 27.7 \\
   \rowcolor{palegray}
   \cellcolor{white} &  \includegraphics[height=0.8em]{icon/intervl.png} \texttt{InternVL3.5-1B-Instruct} &   &  \textsf{V} & & 23.8 & 4.06 & 28.8 & 24.3 & 13.4 & 26.0 & 23.8 & 24.2 & 23.4 & 21.6 \\
   & \includegraphics[height=0.8em]{icon/meta.pdf} \texttt{Llama-3.2-1B-Instruct} &  &  &  & 3.5 & 4.00 & 4.2 & 3.5 & - & 3.0 & 5.3 & 2.6 & 3.9 & 2.9 \\
   
  \cmidrule(lr){2-15}
    & \multicolumn{14}{c}{\textbf{Korean Thinking Models}}\\
  \cmidrule(lr){2-15}
   & \includegraphics[height=0.8em]{icon/lg.png} \texttt{EXAONE-4.0-32B} & \textsf{K} &  & \textsf{R} & 59.9 & 13.57 & 59.2 & 59.9 & - & 58.2 & 64.8 & 52.4 & 63.1 & 61.2 \\
   \rowcolor{palegray}
   \cellcolor{white} &  \includegraphics[height=0.8em]{icon/naver.png} \texttt{HyperCLOVAX-SEED-Think-14B} & \textsf{K} &  & \textsf{R} & 50.8 & 11.29 & 52.1 & 50.8 & - & 51.6 & 53.8 & 41.8 & 55.6 & 51.1 \\
   & \includegraphics[height=0.8em]{icon/lg.png} \texttt{EXAONE-4.0-1.2B} & \textsf{K} &  & \textsf{R} & 37.4 & 7.60 & 42.3 & 37.4 & - & 37.6 & 42.1 & 35.0 & 39.1 & 32.6 \\
   
  \cmidrule(lr){2-15}
    & \multicolumn{14}{c}{\textbf{Korean Instruct Models}}\\
  \cmidrule(lr){2-15}
   & \includegraphics[height=0.8em]{icon/skt.png} \texttt{A.X-4.0} & \textsf{K} &  &  & 76.1 & 15.46 & \textbf{78.9} & 76.1 & - & 76.6 & 77.7 & 68.2 & 81.3 & 76.5 \\
   \rowcolor{palegray}
   \cellcolor{white} &  \includegraphics[height=0.8em]{icon/nc.png} \texttt{VARCO-Vision-2.0-14B} & \textsf{K} & \textsf{V} &  & 58.7 & 11.24 & 57.5 & 59.5 & 42.7 & 59.0 & 62.3 & 54.3 & 61.7 & 56.0 \\
   & \includegraphics[height=0.8em]{icon/skt.png} \texttt{A.X-4.0-Light} & \textsf{K} &  &  & 55.7 & 11.45 & 60.6 & 55.7 & - & 55.8 & 54.4 & 50.9 & 61.4 & 55.7 \\
   \rowcolor{palegray}
   \cellcolor{white} &  \includegraphics[height=0.8em]{icon/skt.png} \texttt{A.X-4.0-VL-Light} & \textsf{K} & \textsf{V} &  & 52.5 & 9.76 & 54.8 & 53.0 & 42.7 & 51.5 & 50.6 & 50.1 & 58.0 & 52.1 \\
   & \includegraphics[height=0.8em]{icon/nc.png} \texttt{VARCO-Vision-2.0-1.7B} & \textsf{K} & \textsf{V} &  & 35.2 & 5.76 & 34.2 & 36.6 & 6.1 & 35.1 & 35.8 & 33.4 & 38.0 & 33.2 \\
   \rowcolor{palegray}
   \cellcolor{white} & \includegraphics[height=0.8em]{icon/naver.png} \texttt{HyperCLOVAX-SEED-Vision-Inst.-3B} & \textsf{K} & \textsf{V} &  & 32.0 & 7.56 & 35.6 & 32.4 & 23.2 & 37.3 & 25.9 & 26.5 & 36.7 & 32.6 \\
   & \includegraphics[height=0.8em]{icon/naver.png} \texttt{HyperCLOVAX-SEED-Text-Inst.-1.5B} & \textsf{K} &  &  & 30.6 & 6.84 & 36.6 & 30.6 & - & 38.5 & 31.4 & 24.7 & 30.6 & 27.0 \\
    \rowcolor{palegray}
    \multirow{-66}{*}{\textbf{Open-source}}
    \cellcolor{white} & \includegraphics[height=0.8em]{icon/naver.png} \texttt{HyperCLOVAX-SEED-Text-Inst.-0.5B} & \textsf{K} &  &  & 13.2 & 4.31 & 14.1 & 13.2 & - & 17.0 & 8.5 & 10.6 & 13.5 & 16.0 \\
  \bottomrule
  \end{tabular}
  \end{adjustbox}
  \end{table*}

\begin{figure*}[ht]
\centering
\includegraphics[width=\linewidth]{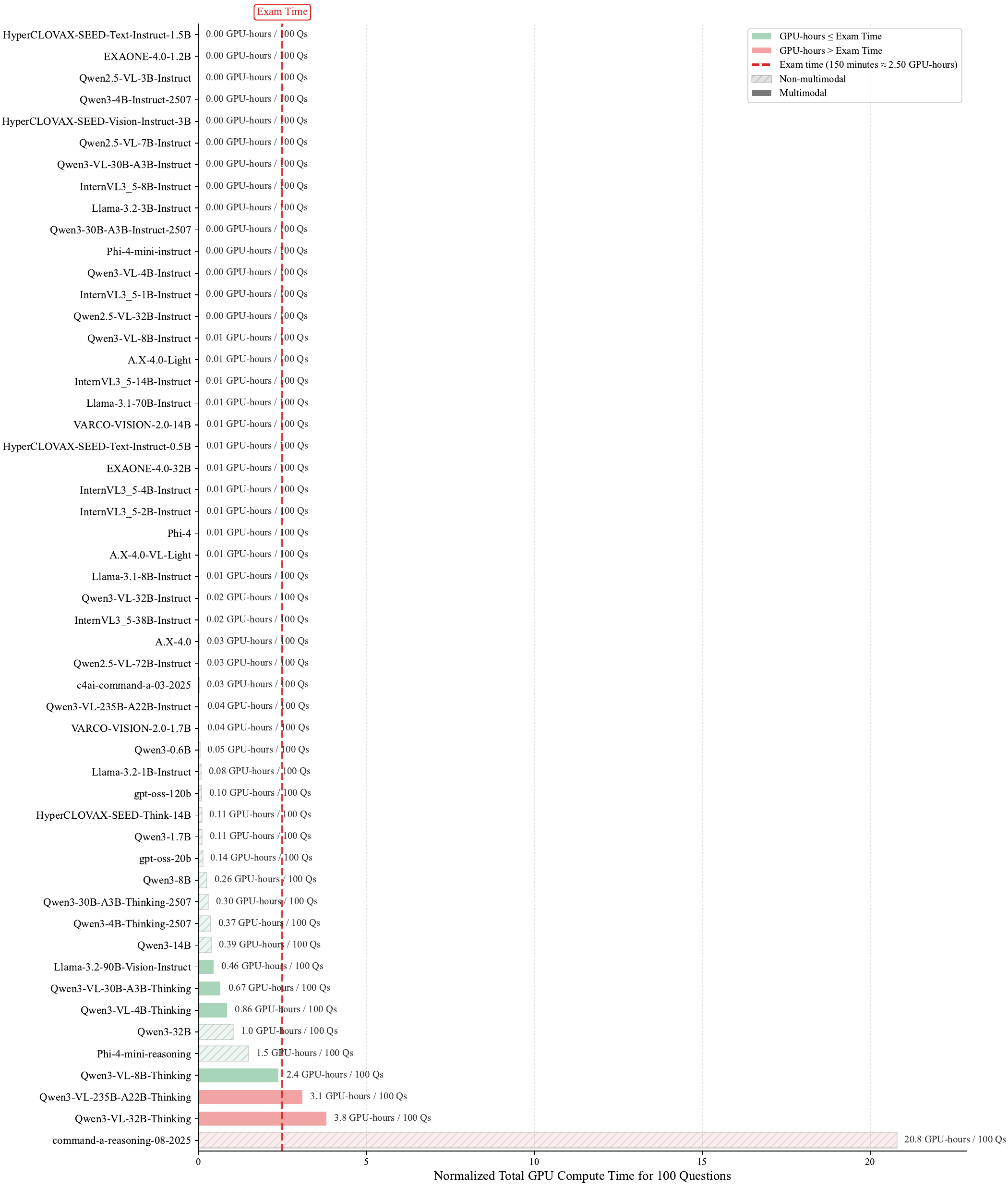}
\caption{\textbf{Normalized total GPU compute time for 100 questions compared across models.} The plot displays the total GPU time required to complete 100 questions, calculated as (wall-clock inference time $\times$ tensor parallelism size $\times$ 100 $\div$ question count). The red dashed line indicates the official time limit for the Meteorological Engineer exam (150 minutes). \textbf{Green bars} denote models that completed the task within the time limit, while \textbf{red bars} indicate models that exceeded it. \textbf{Solid bars} represent multimodal models, and \textbf{hatched bars} represent non-multimodal (text-only) models. All evaluations were conducted using the vLLM library on NVIDIA H100 80GB PCIe GPUs.}
\label{fig:inference_time}
\end{figure*}

\vspace{-1in}

\vspace{-1in}

\clearpage
\section{Robustness of Conclusions Under Small Subsets}
\label{sec:subset_robustness}

To address the concern that small subsets may induce high variance and unstable conclusions, we conducted item-level bootstrap resampling and sensitivity analyses using the evaluation logs.

We confirm via rigorous sensitivity diagnostics—including Bootstrap, leave-one-out (LOO), and Approximate Maximum Influence Perturbation (AMIP)~\citep{huang2026dropping}—that the modality, geo-cultural, and reasoning gaps are robust systemic trends. Bootstrap estimates validate these patterns, LOO perturbations reveal no sign flips, and AMIP analysis demonstrates that the local advantage withstands even substantial adversarial data removal.

\subsection{Statistical Robustness Diagnostics}

We report explicit uncertainty (confidence intervals) and stability diagnostics, including LOO perturbations and bootstrap rank intervals. Unless otherwise noted, all statistics below are computed from the current evaluation run with 1{,}000 bootstrap iterations and a fixed random seed of 42. We confirmed that the bootstrapping process achieved sufficient convergence (not shown).

\paragraph{Setup.}
We focus our robustness checks on three subsets: multimodal (82 items), Korean-specific (71 text-only and 73 multimodal items), and reasoning (121 text-only and 141 multimodal items).
We normalize all scores to $[0, 1]$, mapping reasoning scores from $[4, 20]$ and accuracy from $[0, 100]$.
For the bootstrap analysis, we select 10 representative models for each subset by sorting them by performance and sampling at equal intervals.

\begin{figure*}[t]
    \centering
    \begin{minipage}[t]{0.33\textwidth}
        \centering
        \includegraphics[width=\linewidth]{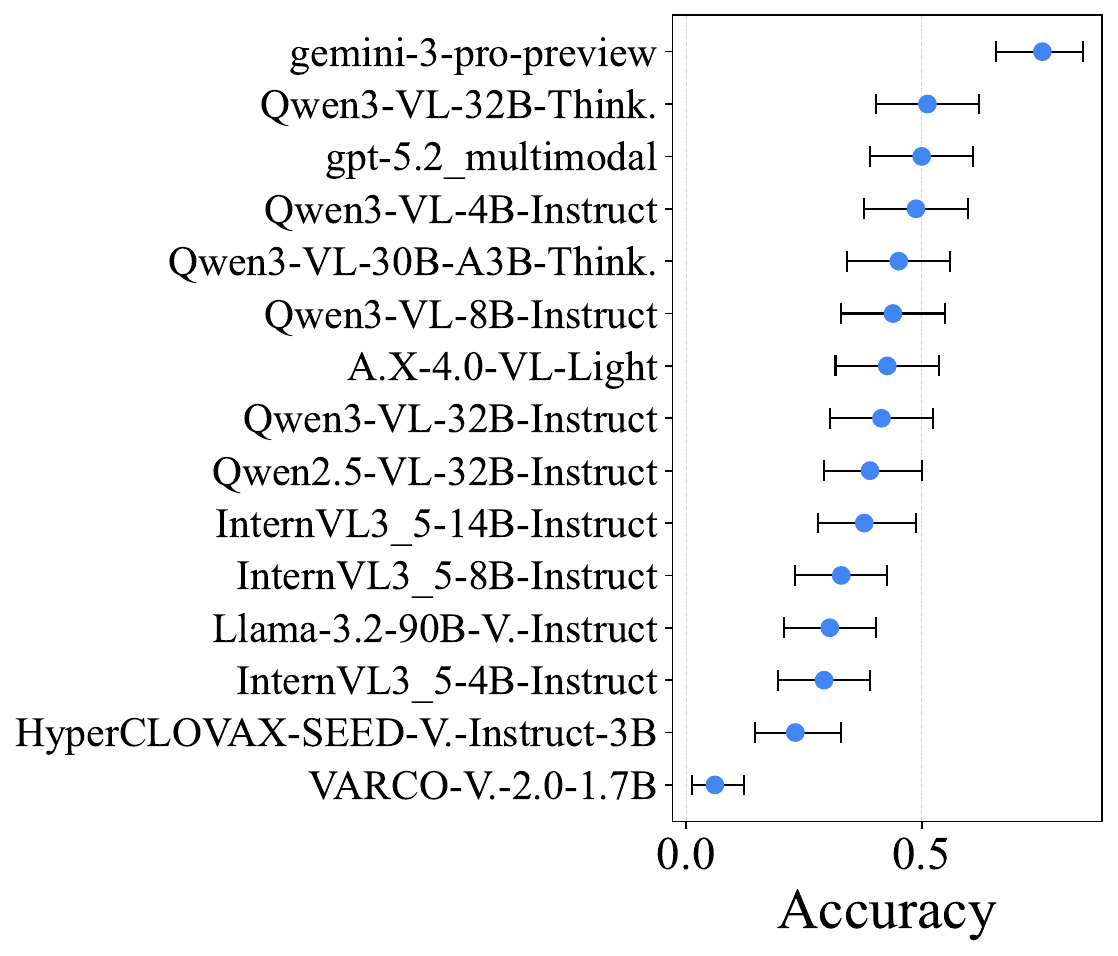}
    \end{minipage}\hfill
    \begin{minipage}[t]{0.33\textwidth}
        \centering
        \includegraphics[width=\linewidth]{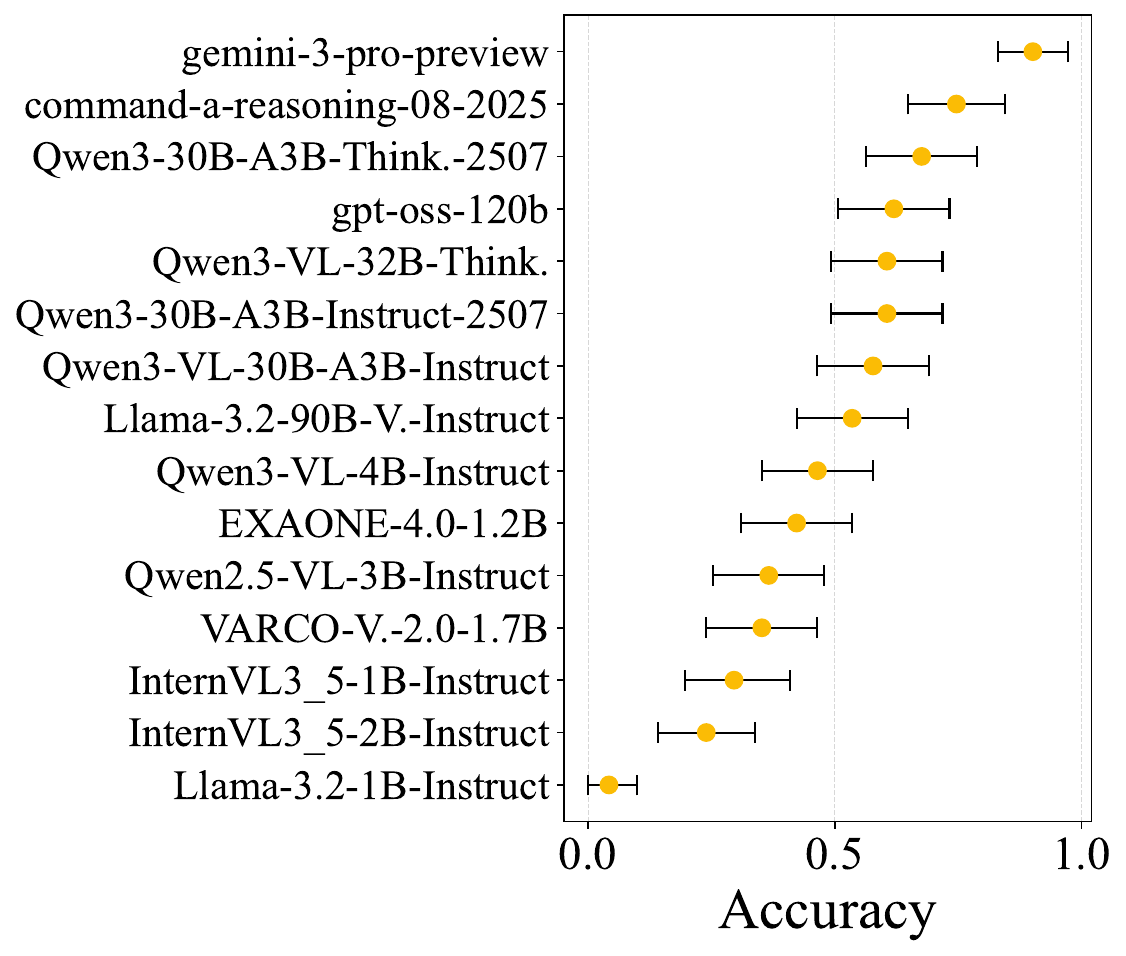}
    \end{minipage}\hfill
    \begin{minipage}[t]{0.33\textwidth}
        \centering
        \includegraphics[width=\linewidth]{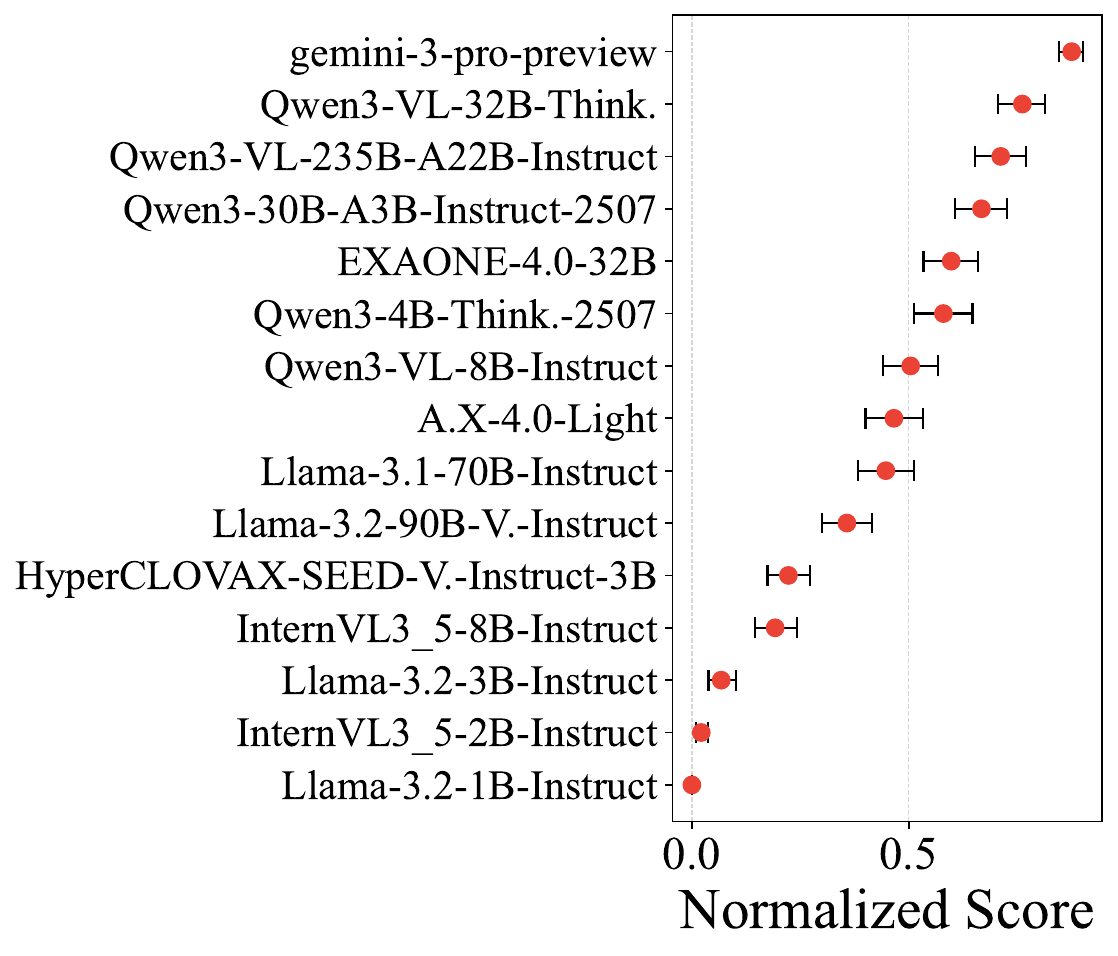}
    \end{minipage}
    \caption{
    \textbf{Ranking stability analysis on the \googb{Multimodal}, \googy{Geo-cultural}, and \googr{Reasoning} subset performances with bootstrap confidence intervals.} 
    Error bars denote 95\% CIs derived from item-level bootstrap resampling on 10 evenly spaced models. 
    Even for the smaller subsets, the uncertainty is explicitly quantified and the coarse separation between higher- and lower-performing systems remains visible across panels. 
    }
    \label{fig:subset_ci}
\end{figure*}

\paragraph{Subset-level uncertainty is quantified, but does not erase structure.}
Figure~\ref{fig:subset_ci} reports 95\% bootstrap confidence intervals for 10 representative models selected at equal intervals from the performance ranking for each subset. This sampling strategy ensures visibility across the full performance spectrum.
Making the uncertainty induced by small \(n\) explicit, we observe that while confidence intervals naturally widen for smaller subsets, the overall performance hierarchy remains robust.
The highest-performing systems consistently maintain their lead with the following estimates (mean [95\% CI]):
Multimodal \(0.756\, [0.659, 0.841]\),
Geo-cultural \(0.789\, [0.690, 0.873]\), and
Reasoning \(0.876\, [0.848, 0.901]\) (normalized score).
This confirms that even under resampling, the conclusions are not dominated by random sampling variation, and the uncertainty bounds provide a principled way to interpret rank differences.

\paragraph{Key performance gaps are stable under resampling and single-item perturbations.}
As shown in Table~\ref{tab:loo}, our analysis confirms that the identified performance gaps across modality, geo-cultural, and reasoning dimensions are systemic and robust, rather than artifacts of specific outliers.

First, regarding the modality gap \((\Delta_{\text{Modality}}=\mathrm{Acc}_{\text{Multimodal}}-\mathrm{Acc}_{\text{Non-Multimodal}})\), we observe a consistently negative trend across the 25 directly comparable models, ranging from \(-37.39\%\) to \(-2.28\%\). For 19/25 models, the 95\% bootstrap CI strictly excludes zero.
Second, we examine the geo-cultural gap \((\Delta_{\text{Geo-Cultural}}=\mathrm{Acc}_{\text{Korean}}-\mathrm{Acc}_{\text{Non-Korean}})\) and the reasoning gap \((\Delta_{\text{Reasoning}} = \text{Score}_{\text{Reasoning}} - \text{Acc}_{\text{Total}})\) (normalized score and accuracy). 
We find that representative local models (e.g., \texttt{HyperCLOVAX}, \texttt{skt/A.X}) maintain a stable positive advantage in the geo-cultural domain, while the reasoning performance remains distinct from general knowledge across models.
To quantify the sensitivity of these conclusions to individual outliers, we run leave-one-out perturbations over the respective subsets: multimodal (\(n=82\)), geo-cultural (\(n=71/73\)), and reasoning (\(n=121/141\)) for text-only and multimodal models, respectively.

Crucially, removing any single item never triggers a sign flip across all evaluated models and dimensions (sign-flip rate = 0). 
Table~\ref{tab:loo} reports the most fragile cases (i.e., baselines closest to zero); even for these edge cases, the maximum swing remains negligible (e.g., <1.16\% for modality, <0.99\% for geo-cultural, and <0.72\% for reasoning). 
This confirms that the observed gaps drive the overarching trends and are not attributable to a handful of influential questions.

\begin{table*}[t]
    \centering
    \caption{
    \textbf{Leave-one-out (LOO) sensitivity analysis across \googb{Multimodal}, \googy{Geo-cultural}, and \googr{Reasoning} subsets.} 
    The table reports representative models with baseline gaps closest to zero (most prone to sign flips). 
    No model exhibits a sign reversal under single-item removal across all tasks.
    (\textbf{Baseline \(\Delta\)}: original gap on the full set; 
    \textbf{Max swing}: maximum deviation from the baseline across LOO iterations; 
    \textbf{Sign flip}: percentage of iterations where the gap sign reverses; 
    \(n_{\text{LOO}}\): total number of items subject to LOO perturbation.)
    }
    \label{tab:loo}
    \resizebox{\textwidth}{!}{%
    \setlength{\tabcolsep}{15pt}
    \begin{tabular}{lcccc}
        \toprule
        \textbf{Model} & \textbf{Baseline} \(\Delta\) & \textbf{Max swing} & \textbf{Sign flip (\%)} & \(n_{\text{LOO}}\) \\
    \midrule
    \multicolumn{5}{l}{\textit{\textbf{(A) \googb{Modality Gap}} (Multimodal \(-\) Text-only)}} \\
    \cmidrule(lr){1-5}
    \includegraphics[height=0.5em]{icon/qwen.pdf} \texttt{Qwen/Qwen3-VL-4B-Instruct} & \(-2.28\%\) & \(0.63\%\) & 0.0 & 82 \\
    \includegraphics[height=0.8em]{icon/intervl.png} \texttt{OpenGVLab/InternVL3\_5-2B-Instruct} & \(-4.38\%\) & \(0.90\%\) & 0.0 & 82 \\
    \includegraphics[height=0.8em]{icon/naver.png} \texttt{HyperCLOVAX-SEED-Vision-Instruct-3B} & \(-9.22\%\) & \(0.95\%\) & 0.0 & 82 \\
    \includegraphics[height=0.8em]{icon/skt.png} \texttt{skt/A.X-4.0-VL-Light} & \(-10.33\%\) & \(0.71\%\) & 0.0 & 82 \\
    \includegraphics[height=0.5em]{icon/qwen.pdf} \texttt{Qwen/Qwen3-VL-8B-Instruct} & \(-10.35\%\) & \(0.69\%\) & 0.0 & 82 \\
    \midrule
    \multicolumn{5}{l}{\textit{\textbf{(B) \googy{Geo-cultural Gap}} (Korean \(-\) Non-Korean)}} \\
    \cmidrule(lr){1-5}
    \includegraphics[height=0.8em]{icon/naver.png} \texttt{HyperCLOVAX-SEED-Text-Instruct-1.5B} & \(6.27\%\) & \(0.91\%\) & 0.0 & 71 \\
    \includegraphics[height=0.8em]{icon/intervl.png} \texttt{OpenGVLab/InternVL3\_5-1B-Instruct} & \(5.13\%\) & \(0.99\%\) & 0.0 & 73 \\
    \includegraphics[height=0.8em]{icon/lg.png} \texttt{LGAI-EXAONE/EXAONE-4.0-1.2B} & \(5.12\%\) & \(0.82\%\) & 0.0 & 71 \\
    \includegraphics[height=0.8em]{icon/skt.png} \texttt{skt/A.X-4.0-Light} & \(5.04\%\) & \(0.87\%\) & 0.0 & 71 \\
    \includegraphics[height=0.8em]{icon/naver.png} \texttt{HyperCLOVAX-SEED-Vision-Instruct-3B} & \(3.81\%\) & \(0.89\%\) & 0.0 & 73 \\
    \midrule
    \multicolumn{5}{l}{\textit{\textbf{(C) \googr{Reasoning Gap}} (Reasoning \(-\) Knowledge)}} \\
    \cmidrule(lr){1-5}
    \includegraphics[height=0.5em]{icon/qwen.pdf} \texttt{Qwen/Qwen3-32B} (Thinking)  & \(18.26\%\) & \(0.55\%\) & 0.0 & 121 \\
    \includegraphics[height=0.8em]{icon/openai.pdf} \texttt{gpt-5.2} (Thinking) & \(6.27\%\) & \(0.61\%\) & 0.0 & 141 \\
    \includegraphics[height=0.5em]{icon/qwen.pdf} \texttt{Qwen/Qwen3-30B-A3B-Instruct-2507} & \(2.50\%\) & \(0.56\%\) & 0.0 & 121 \\
    \includegraphics[height=0.5em]{icon/qwen.pdf} \texttt{Qwen/Qwen3-4B-Instruct-2507} & \(0.65\%\) & \(0.43\%\) & 0.0 & 121 \\
    \includegraphics[height=0.5em]{icon/qwen.pdf} \texttt{Qwen/Qwen3-VL-32B-Instruct} & \(0.51\%\) & \(0.72\%\) & 0.0 & 141 \\
    \bottomrule
    \end{tabular}
    }
\end{table*}

\subsection{Robustness of Key Findings to Critical Data Perturbation}
\label{app:robustness_perturbation}

\paragraph{Resilience to Adversarial Item Removal.}
To further assess whether the geo-cultural gap could be driven by a small number of influential questions, we conduct an adversarial influential-item deletion analysis inspired by recent robust evaluation methodologies, AMIP~\citep{huang2026dropping}. 
Concretely, we focus on the representative model pair highlighted in Section~\ref{sec:result} (The Geo-Cultural Gap): the top-performing local model (\texttt{skt/A.X-4.0}) versus the global baseline (\texttt{Qwen/Qwen3-VL-235B-A22B-Thinking}). 
We test how many \emph{strategically selected} items an adversary would need to delete to reverse the ordering where the local model outperforms the global one.

We fit a Bradley-Terry (BT) model to pairwise win/loss outcomes on the Geo-cultural subset and compute per-item influence scores using the inverse Hessian of the BT loss. 
Using a greedy adversarial removal procedure (a standard approximation for AMIP) that sequentially deletes the most influential items favoring the local model, we find that the ordering flips only after removing \(n_{\text{AMIP}}=18\) items, corresponding to \(18/73 \approx 24.7\%\) of the entire Geo-cultural subset.
This indicates that the observed local model advantage is not attributable to a single outlier or a very small number of questions, but instead requires removing a substantial fraction of the subset to overturn.

\end{document}